\pdfoutput=1
\documentclass[12pt]{article}
\usepackage{amsmath}
\usepackage{times}
\usepackage{graphicx}
\usepackage{color}
\usepackage{multirow}
\usepackage{import} 
\usepackage[style=apa,backend=biber,language=american]{biblatex}
\DeclareLanguageMapping{american}{american-apa}
\usepackage{hyperref}
\usepackage{rotating}
\usepackage{bbm}
\usepackage{latexsym}

\newcommand{\captionfonts}{\normalsize}

\makeatletter  
\long\def\@makecaption#1#2{%
  \vskip\abovecaptionskip
  \sbox\@tempboxa{{\captionfonts #1: #2}}%
  \ifdim \wd\@tempboxa >\hsize
    {\captionfonts #1: #2\par}
  \else
    \hbox to\hsize{\hfil\box\@tempboxa\hfil}%
  \fi
  \vskip\belowcaptionskip}
\makeatother   

\usepackage{amsmath,amsfonts,bm}

\def\eqref#1{equation~\ref{#1}}

\def\1{\bm{1}}

\DeclareMathAlphabet{\mathsfit}{\encodingdefault}{\sfdefault}{m}{sl}
\SetMathAlphabet{\mathsfit}{bold}{\encodingdefault}{\sfdefault}{bx}{n}

\newcommand{\citep}{\parencite}
\newcommand{\citet}{\cite}

\newcommand*{\eg}{\emph{e.g.~}}
\newcommand*{\ie}{\emph{ie.~}}
\newcommand*{\etal}{\emph{et al.~}}
\newcommand*{\etc}{\emph{etc.}}

\begin{document}
\hspace{13.9cm}

\ \vspace{20mm}\\

{\LARGE \noindent Do Neural Networks for Segmentation \\ Understand Insideness?}

\ \\
{\bf \large Kimberly Villalobos$^{*,1}$, Vilim \v{S}tih$^{*,1,2}$, Amineh Ahmadinejad$^{*,1}$, }\\
{\bf \large Shobhita Sundaram$^1$,   Jamell Dozier$^1$,  {Andrew Francl$^1$},\\  {Frederico Azevedo$^1$},} {\bf \large {Tomotake Sasaki}$^{\dagger,1,3}$, {Xavier Boix}$^{\dagger,1,\bullet}$} \\
$^*$ and $^\dagger$ indicate equal contribution\\
$^\bullet$ Correspondence to \texttt{xboix@mit.edu}\\ \\
$^1$ Center for Brains, Minds and Machines, Massachusetts Institute of Technology (USA) \\
$^2$ Max Planck Institute of Neurobiology (Germany)\\
$^3$ Fujitsu Laboratories Ltd. (Japan) \\ \\

\noindent {\bf Published at:} Neural Computation 33, 2511–2549 (2021) \\ \url{https://doi.org/10.1162/neco_a_01413} \\
\ \\[-2mm]

\thispagestyle{empty}
\markboth{}{Do Neural Networks for Segmentation Understand Insideness?}
\ \vspace{-0mm}\\
%
\begin{center} {\bf Abstract} \end{center}


The insideness problem is an aspect of image segmentation that consists of determining which pixels are inside and outside a region. Deep Neural Networks (DNNs) excel in segmentation benchmarks, but it is unclear if they have the ability to solve the insideness problem as it requires evaluating long-range spatial dependencies. In this paper, the insideness problem is analysed in isolation, without texture or semantic cues, such that other aspects of segmentation do not interfere in the analysis. We demonstrate that DNNs for segmentation with few units have sufficient complexity to solve insideness for any curve. Yet, such DNNs have severe problems with learning general solutions. Only recurrent networks trained with small images learn solutions that generalize well to almost any curve. Recurrent networks can decompose the evaluation of long-range dependencies into a sequence of local operations, and learning with small images alleviates the common difficulties of training recurrent networks with a large number of unrolling steps. 

\section{Introduction}

A key component of image segmentation is to determine whether a pixel is inside or outside a region,~\ie the ``insideness'' problem~\citep{Ullman84,Ullman96}. This problem involves evaluating long-range spatial dependencies. Capturing such long-range dependencies may be challenging for artificial neural networks as pointed out in Minsky \& Papert's historic book~\emph{Perceptrons}~\citep{MP69} and recent works on capturing other spatial relationships such as containment~\citep{kim2018not} and connectedness~\citep{linsley2018learning}. 

Deep Neural Networks (DNNs) have been tremendously successful in image segmentation benchmarks, but it is not well understood whether DNNs represent insideness or how.
Insideness has been overlooked in DNNs for segmentation since they have been mainly applied to the modality of ``semantic segmentation'',~\ie labelling each pixel with its object category~\citep{RFB15,YuKoltun2016,visin2016reseg,BKC17,chen2018deeplab,long2015fully,LATEEF2019321}. The same could be said for DNNs for more advanced segmentation modalities and applications that have been recently introduced,~\eg segmentation of individual object instances rather than  object categories~\citep{li2016iterative,li2017fully,song2018seednet,chen2018masklab,hu2018learning,maninis2018deep,liu2018path,he2017mask} and generating realistic images~\citep{zhu2017unpaired}. In such cases, insideness has not been considered because solutions can rely  on texture, shape and other visual cues. Yet, investigating whether DNNs understand insideness could reveal new insights about their ability to capture long-range spatial relationships, which is key for a full image understanding.



In this paper, we investigate analytically-derived and learned representations in DNNs for insideness. We take the reductionist approach by isolating insideness such that other components of image segmentation do not provide additional cues and ensure that our analysis focuses on the long-range spatial dependencies involved  in insideness. Thus, we analyze the segmentation of Jordan curves,~\ie 
closed curves synthetically generated without texture nor object category. 
We analytically demonstrate that state-of-the-art network architectures,~\ie DNNs with dilated convolutions~\citep{YuKoltun2016,chen2018deeplab} and convolutional LSTMs (ConvLSTMs)~\citep{xingjian2015convolutional}, among other networks, can exactly solve the insideness problem for any  curve with network sizes that are easily implemented in practice. The proofs draw on algorithmic ideas from classical work on visual routines~\citep{Ullman84,Ullman96}, namely, the ray-intersection method and the coloring method, to derive equivalent neural networks that implement these algorithms.  
However, our experiments show that in practice, most DNNs  for segmentation do not learn general solutions for insideness, even though the architectures are complex enough to capture the long-range relationships. These DNNs learn to recognize specific features of the family of curves of the training set that do not generalize to new families of curves lacking those features. Only recurrent networks (such as the {ConvLSTM})  when trained on small images, generalize to almost any given curve of any size. This is because training on small images alleviates the well-known difficulties of training recurrent networks with a large number of unrolling steps~\citep{bengio1994learning,HS97,pascanu2013difficulty,gruslys2016memory}. It also facilitates learning a strategy that deals with long-range dependencies by breaking them into local operations that are reusable for any curve, even curves in larger images that were not seen during training. Our investigations are summarized in three questions and this paper provides the answers to them: 
\begin{enumerate}
    \item \label{contrib_1} \emph{Are state-of-the-art DNNs for segmentation sufficiently complex to solve insideness?} Yes, they are sufficiently complex to implement a general solution for insideness for any given curve.
    \item \emph{Do state-of-the-art DNNs for segmentation learn general solutions for insideness with the standard training paradigm?} No, they do not. Even though they are sufficiently complex to solve insideness, the learnt solutions consist of recognizing patterns of the curves in the training set that are not general.
    \item \emph{Is there a training procedure that leads to learning general solutions for insideness?}  Yes, this is the case for convolutional recurrent networks trained on small images. Recurrent networks can decompose the evaluation of long-range dependencies into a sequence of local operations, and learning with small images alleviates the common difficulties of training recurrent networks with a large number of unrolling steps.  
\end{enumerate}

These results  add  to the growing body of work that demonstrates that DNNs have problems 
in learning to solve 
some elemental visual tasks~\citep{linsley2018learning,liu2018intriguing,wu2018cognitive,SOS17}. \cite{SOS17} introduced several tasks that DNNs can in theory solve, as demonstrated mathematically, but were unable to solve in practice, even for the training dataset, due to difficulties in the optimization with gradient descent. In contrast, the challenges we report for insideness are related to poor generalization rather than optimization, as 
our experiments show  the networks 
succeed in solving  
insideness for the family of curves seen during training. \cite{linsley2018learning} and \cite{KimDisentangling} introduced new architectures that better capture the long-range dependencies in images. 
Here, we show that the training strategy has a significant impact on capturing the long-range dependencies, as even DNNs with the capacity to capture such dependencies do not learn a general solution with standard training strategies. Our results highlight the need to decompose the evaluation of long-range dependencies in a sequence of local operations, that can be learned with recurrent networks by controlling the number of unrolling steps with the image size.

\section{The Reductionist Approach to Insideness}

We now introduce a paradigm for analyzing the ability of DNNs to solve insideness.
Rather than natural images, we use synthetic stimuli that consist only of one closed curve. In this way, we do not mix the insideness problem with other components of image segmentation found in natural images,~\eg  discontinuity of segments, representation of the hierarchy of segments, \etc~ This reductionist methodology has the advantage of minimizing the interference of these other factors in analysing abilities of DNNs to capture long-range spatial dependencies. Note that the presence of other factors would obfuscate the specific causes of the network's behavior.




Let $\boldsymbol{ X} \in \{0, 1 \}^{N \times N}$ be an image or a matrix of size $N\times N$ pixels.
We use $X_{i,j}$ or $\left(\boldsymbol{X} \right)_{i,j}$, indistinguishably, to denote the value of the image  in position $(i,j)$.
We use this notation for indexing elements in any of the images and matrices that appear in the rest of the paper.
Also, in the figures we use white and black to represent $0$ and $1$, respectively.  


The \emph{insideness problem} refers to assigning pixels to the inside or the outside of a closed curve. 
We assume without loss of generality that there is only one closed curve in the image and that it is a digital version of a Jordan curve~\citep{Kong01},~\ie a closed curve without self-crosses nor self-touches and containing only horizontal and vertical turns,  as shown in Fig.~\ref{fig:Datasets}a. 
We further assume that the curve does not contain the border of the image. 
The curve is the set of pixels equal to $1$ and is denoted by $\mathcal{ F}_{\boldsymbol{X}}=\{(i,j)| X_{i,j}=1\}$.



The pixels in $\boldsymbol{X}$ that are not in $\mathcal{ F}_{\boldsymbol{X}}$ can be classified into two categories: the inside and the outside of the curve~\citep{Kong01}. We define the segmentation of $\boldsymbol{ X}$ as $\boldsymbol{ S}(\boldsymbol{ X}) \in \{0, 1 \}^{N \times N}$, where
\begin{align}
\left(\boldsymbol{S}(\boldsymbol{ X})\right)_{i,j}&=
\left \{
    \begin{array}{cll}
    0 & \text{if }  \text{${X}_{i,j}$ is inside}  \\
    1 & \text{if } \text{${X}_{i,j}$ is  outside} 
    \end{array}
\right.  ,
\label{ConditionOfAnswer}
\end{align}
and for the pixels in $\mathcal{ F}_{\boldsymbol{X}}$, the value of $\left(\boldsymbol{S}(\boldsymbol{ X})\right)_{i,j}$ can be either $0$ or $1$. Note that the definition of insideness 
is rigorously and uniquely determined by the input image itself.

The number of all digital Jordan curves is enormous even if the image size is relatively small,~\eg it is more than $10^{47}$ for the size $32 \times 32$ (Appendix~\ref{secSuppJordan}). 
In addition, insideness is a global problem; whether a pixel is inside or outside depends on the entire image, and not just on a local area around the pixel. 
Therefore, simple pattern matching,~\ie memorization, is impossible in practice.




\label{secJordan}

\section{Are state-of-the-art DNNs for Segmentation Sufficiently Complex to Solve Insideness?} \label{SecRay-Intersection}

\label{sec:dnns}
The universal approximation theorem~\citep{cybenko1989approximation} tells us that even a shallow neural network is able to solve the insideness problem. Yet, it could be that the amount of units is too large to be implementable in practice. We show that two commonly used DNN architectures for segmentation are able to perfectly solve the insideness problem, and they are easily implementable in practice.  One architecture is feed-forward with dilated convolutions~\citep{YuKoltun2016,chen2018deeplab} and the other is recurrent, based on ConvLSTMs~\citep{xingjian2015convolutional,visin2016reseg,li2018referring,alom2018recurrent}.

\noindent {\bf Dilated convolutions.}
Also called atrous convolutions, these are convolutions with upsampled kernels, which enlarge the receptive fields of the units but preserve the number of parameters of the kernel~\citep{YuKoltun2016,chen2018deeplab}. They facilitate capturing long-range dependencies which are key for segmentation~\citep{YuKoltun2016,chen2018deeplab}. To demonstrate that there are architectures with dilated convolutions that can solve the insideness problem, we borrow insights from the ray-intersection method. 
The ray-intersection method~\citep{Ullman84,Ullman96}, 
also known as the crossings test or the even-odd test~\citep{Haines94},  
is built on the following fact: Any ray that goes from a pixel to the border of the image alternates between inside and outside regions every time it crosses the curve. Therefore, the parity of the total number of such crossings determines the region to which the
pixel belongs. If the parity is odd then the pixel is inside, otherwise it is outside. In Appendix~\ref{secDilation}, we introduce a DNN with dilated convolutions that can implement the ray-intersection algorithm: a network with a number of dilated convolutional layers equal to $\log_2(N)$ (recall $N$ is the image size) with one kernel of  $3\times 3$ size, and two convolutional layers with $3N/2$ kernels of  $1\times 1$ size. This is the smallest network we could find that solves the insideness problem with dilated convolutions. Larger networks than the one we introduced can also solve the problem, as the network size can be reduced by setting kernels to zero and layers to implement the identity operation. 



\begin{figure}[t]
\centering
\footnotesize
\begin{tabular}{c@{\hspace{1cm}}c}
 \multirow{2}{*}{
 \begin{tabular}{c}
 \vspace*{-3.5cm}\\ \includegraphics[width=0.42\linewidth]{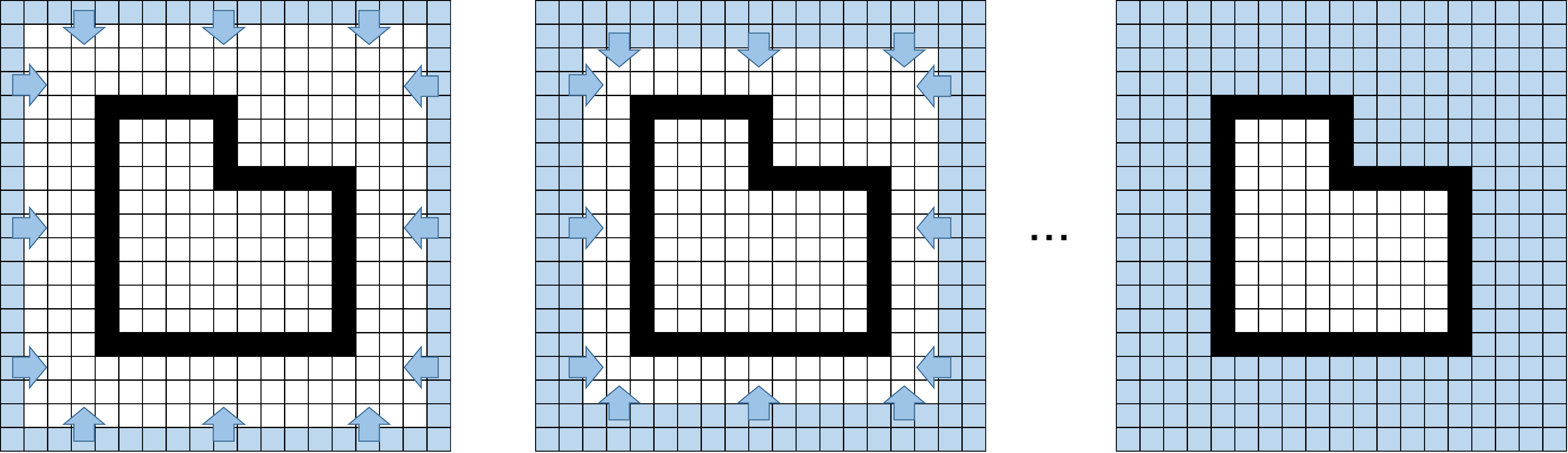}\\[1em] (a) 
 \end{tabular}} &
  \includegraphics[width=0.45\linewidth]{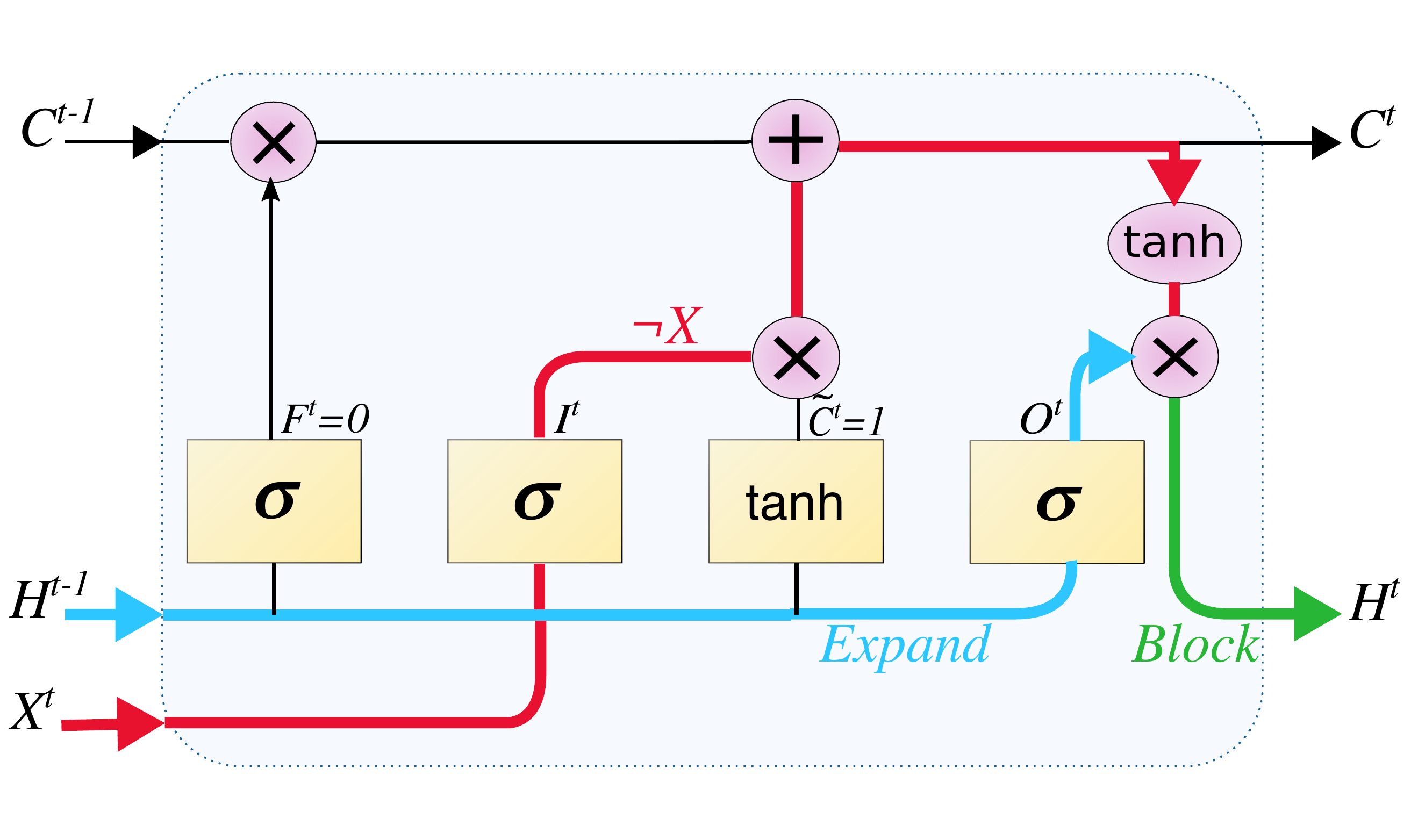}\\
 & (b) \\
\end{tabular}
  \caption{\emph{The Coloring Method with ConvLSTM}. (a) The coloring method consists  of  several iterations of  the \emph{coloring routine},~\ie expanding the outside region and blocking it on the curve. (b) Diagram of the ConvLSTM implementing the coloring method, we  highlight the connections between layers that are used for insideness. $\neg \boldsymbol{X}$ denotes the element-wise ``Boolean not'' of $\boldsymbol{X}$.} 
    \label{fig:fig/ColoringMethod.eps}
\end{figure}

\label{SecColoring}

\noindent {\bf Convolutional LSTM (ConvLSTM)}. 
ConvLSTM breaks the evaluation of long-range dependencies in a sequence of local operations, \ie one step of the ConvLSTM is a local operation that does not tackle long-range dependencies but applying the ConvLSTM multiple steps allows resolving longer-range dependencies. We now show that a ConvLSTM with just one kernel of size $3 \times 3$ is sufficient to solve the insideness problem, when the ConvLSTM is applied multiple steps.




Our demonstration is inspired by the coloring method~\citep{Ullman84,Ullman96},
 another algorithm for the insideness problem. The algorithm assigns neighbouring pixels to the same region until encountering the curve, which is also known as flood filling. 
  We introduce a version of this method that can be implemented as a ConvLSTM.
The coloring method consists of multiple iterations of two steps: \emph{(i)} expand the outside region from the borders of the image (which by assumption are in the outside region) and \emph{(ii)}~block the expansion when the curve is reached. 
The repeated application of these two steps solves the insideness problem, as depicted in Fig.~\ref{fig:fig/ColoringMethod.eps}a. We call one iteration of expanding and blocking the \emph{coloring routine}.


We use $\boldsymbol{E}^{t} \in \{0,1\}^{N\times N}$ (expansion) and $\boldsymbol{B}^{t} \in \{0,1\}^{N\times N}$ (blocking) to represent the result of the two operations after iteration $t$. The \emph{coloring routine} can then be written as \emph{(i)} $\boldsymbol{E}^{t} = \mbox{Expand}\left( \boldsymbol{B}^{t-1} \right)$ and \emph{(ii)} $\boldsymbol{B}^t = \mbox{Block}\left( \boldsymbol{E}^{t}, \mathcal{F}_{\boldsymbol{X}} \right)$. Let $\boldsymbol{B}^{t-1}$ maintain a value of $1$ for all pixels that are known to be outside and $0$ for all pixels whose region is not yet determined or belong to the curve. Thus, we initialize $\boldsymbol{B}^{0}$ to have value $1$ (outside) for all  border pixels of the image and $0$ for the rest.  In step \emph{(i)}, the outside region of $\boldsymbol{B}^{t-1}$ is expanded by setting also to $1$ (\emph{outside}) its neighboring pixels, and the result is assigned to $\boldsymbol{E}^{t}$. Next, in step \emph{(ii)}, the pixels in $\boldsymbol{E}^{t}$ that were labeled with $1$ (\emph{outside}) and belong to the curve, $\mathcal{F}_{\boldsymbol{X}}$, are reverted to $0$  (\emph{inside}), and the result is assigned to $\boldsymbol{B}^{t}$. This algorithm ends when the outside region can not expand anymore, which is always less than $N^2$ iterations (the number of pixels in the image). Therefore, we have $\boldsymbol{E}^{N^2}= \boldsymbol{S}(\boldsymbol{X})$.




\label{secLSTM}

In Appendix~\ref{secSuppEXPBLCK} we demonstrate that a 
ConvLSTM with one kernel applied on an image $\boldsymbol{X}$ can implement the coloring algorithm. In the following we 
provide a summary of the proof. Let $\boldsymbol{I}^{t}$,  $\boldsymbol{F}^{t}$, $\boldsymbol{O}^{t}$, $\boldsymbol{C}^{t}$,  and  $\boldsymbol{H}^{t}\in \mathbb{R}^{N\times N}$ be the activations of the input, forget, and output gates, and cell and hidden states of a ConvLSTM at step $t$, respectively.  By analyzing the equations of the ConvLSTM (\eqref{equationLSTM:o} and~\eqref{equationLSTM:H} in Appendix~\ref{secSuppEXPBLCK}) we can see that the output layer, $\boldsymbol{O}^{t}$, back-projects to the hidden layer, $\boldsymbol{H}^{t}$. In the coloring algorithm, $\boldsymbol{E}^{t}$ and $\boldsymbol{B}^{t}$ are related in a similar manner. Thus, we define $\boldsymbol{O}^{t}=\boldsymbol{E}^{t}$ (expansion) and $\boldsymbol{H}^{t}=\frac{1}{2} \boldsymbol{B}^{t}$ (blocking). The $\frac{1}{2}$ factor is a technicality due to  non-linearities, which is compensated in the output gate and has no relevance in this discussion.

We initialize $\boldsymbol{H}^{0}=\frac{1}{2}\boldsymbol{B}^{0}$ (recall $\boldsymbol{B}^{0}$ is $1$ for all pixels in the border of the image and  $0$ for the rest). The output gate expands the hidden representations using one $3\times 3$ kernel. 
To stop the outside region from expanding to the inside of the curve, $\boldsymbol{H}^{t}$ takes the expansion output $\boldsymbol{O}^{t}$ and sets the  pixels at the curve's location to $0$ (inside). This is same as the element-wise product of  $\boldsymbol{O}^{t}$ and the
``Boolean Not'' of $\boldsymbol{X}$, which is denoted as $\neg \boldsymbol{X}$. Thus, the blocking operation can be implemented as   $\boldsymbol{H}^{t} = \frac{1}{2}(\boldsymbol{O}^{t} \odot \neg \boldsymbol{X})$, and
can be achieved if $\boldsymbol{C}^{t}$ is equal to $\neg \boldsymbol{X}$. In Fig.~\ref{fig:fig/ColoringMethod.eps}b we depict these computations.

In Appendix~\ref{secSuppEXPBLCK} we show that the weights of a ConvLSTM with just one kernel of size $3\times 3$ can be configured to reproduce these computations. A key component is that many of the weights  use a value that tends to infinity. This value is denoted as $q$ and it is used to saturate the non-linearities of the ConvLSTM, which are hyperbolic tangents and sigmoids. Note that it is common in practice to have weights that asymptotically tend to infinity,~\eg when using the cross-entropy loss to train a network~\citep{soudry2018implicit}. In practice, we found that saturating  non-linear units using $q=100$ is enough to solve the insideness problem for all curves in our datasets.
Note that only one kernel is sufficient for ConvLSTM to solve the insideness problem, regardless of image size.  Furthermore, networks with multiple stacked  ConvLSTM and more than one kernel  can implement the coloring method by setting 
    unnecessary  ConvLSTMs to implement the identity operation (Appendix~\ref{secSuppEXPBLCK}) and the unnecessary kernels to $0$.

Finally, we point out that there are networks with a much lower complexity than LSTMs that can solve the insideness problem, although these networks rarely find applications in practice. In Appendix~\ref{secRNN}, we show that a convolutional recurrent network as small as having one sigmoidal hidden unit per pixel, with a $3\times 3$ kernel, can also solve the insideness problem for any given curve.


 

\section{Do state-of-the-art DNNs for Segmentation Learn General Solutions for Insideness?} 
\label{SecExperiments}

After having identified DNNs with an implementable number of units that have sufficient complexity to solve the insideness problem, we focus on analyzing whether these solutions can be learnt from examples. 
In the following, we first describe the experimental setup and then show that DNNs trained in standard manner
fail to learn general solutions of insideness.

\subsection{Experimental Setup}
Recall that the goal of the network is to predict for each pixel in the image whether it is inside or outside of the curve. Thus, we have a binary classification problem for each pixel. Since the insideness problem is the same problem for all pixels and we use convolutional layers, each pixel in the image provides one training example of insideness. 

\noindent {\bf Datasets.}  Given that the number of Jordan curves explodes exponentially with the image size, a procedure that could provide curves without introducing a  well-defined bias for learning 
is unknown. We introduce three algorithms to generate different types of Jordan curves. For each dataset, we generate $95K$ images for training, $5K$ for validation and $10K$ for testing.  All the datasets are constructed to fulfill the constraints introduced in Section~\ref{secJordan}. We construct three different datasets of $42\times 42$ pixel image size, called Polar, Spiral and Digs. Fig.~\ref{fig:Datasets}a shows examples of curves for each dataset, see Appendix~\ref{secSuppDataset} for the description on how the curves are generated. Note that the Polar dataset has different complexities depending on the number of vertices of the shape. We refer to each of these datasets as \emph{Polar} with a prefix with the amount of vertices,~\eg $24$-Polar.

\noindent {\bf Evaluation metrics.} Insideness is evaluated for every pixel except the pixels in the curve $\mathcal{F}_{\boldsymbol{X}}$.  We use the following  metrics: \emph{per pixel accuracy} (average of the accuracy for inside and outside, evaluated separately, such that it weights the two categories equally as there is an imbalance of inside and outside pixels) and \emph{per image accuracy} (each image is considered correctly classified if  all the pixels in the image are correctly classified).

\noindent {\bf Architectures.} We evaluate the networks that we analyzed theoretically and also other common architectures: \\
\noindent - \emph{Feed-forward Architectures:} We use the dilated convolutional DNN (\emph{Dilated}) introduced in 
Section~\ref{sec:dnns}, based on the architecture by \cite{YuKoltun2016}. 
We also evaluate two variants of \emph{Dilated}, which are the Ray-intersection network (\emph{Ray-int.}), which uses a receptive field of $1\times N$ instead of the dilated convolutions (see Appendix~\ref{secDilation}), and a convolutional network (\emph{CNN}), which has all the dilation factors set to $1$. 
Finally, we also evaluate \emph{UNet}, which is a popular architecture with skip connections and de-convolutions and has achieved state-of-the-art accuracy of segmentation of gray-scale medical images~\citep{RFB15}. \\
\noindent - \emph{ Recurrent Architectures:}  We test the ConvLSTM (\emph{1-LSTM}) corresponding to the architecture introduced in Section~\ref{secLSTM}. We initialize the hidden and cell states to $0$ (inside) everywhere except the border of the image which is initialized to $1$ (outside), such that the network can learn to color by expanding the outside region. We also evaluate a 2-layers ConvLSTM (\emph{2-LSTM}) by stacking one \emph{1-LSTM} after another, both with the initialization of the hidden and cell states of the \emph{1-LSTM}. Finally, to evaluate the effect of such  initialization, we test the \emph{2-LSTM} without it (\emph{2-LSTM w/o init.}),~\ie with the hidden and cell states   
initialized all
to $0$. We use backpropagation through time by unrolling $60$, $30$ or $10$ time steps for training (we select the best performing one). For testing, we unroll until there is no change in the output labeling.

\noindent {\bf Learning.} We test a large set of  hyperparameters and architecture variants (we trained several thousands of networks per dataset), which we report in detail in Appendix~\ref{secHyper}. In the following we report the testing accuracy  for setup that achieved the highest per image accuracy at the validation set. 

\renewcommand{\arraystretch}{0}

\begin{figure}[t]
  \scriptsize
  \begin{tabular}{@{\hspace{-0cm}}c@{\hspace{0.1cm}}c@{\hspace{0.1cm}}c}
    \multirow{2}{*}{\begin{tabular}{@{\hspace{-0.1cm}}c}
        \vspace*{-2.8cm}\\
    {\bf \footnotesize Examples of Jordan Curves of Each Dataset }\\
        \includegraphics[width=0.42\textwidth]{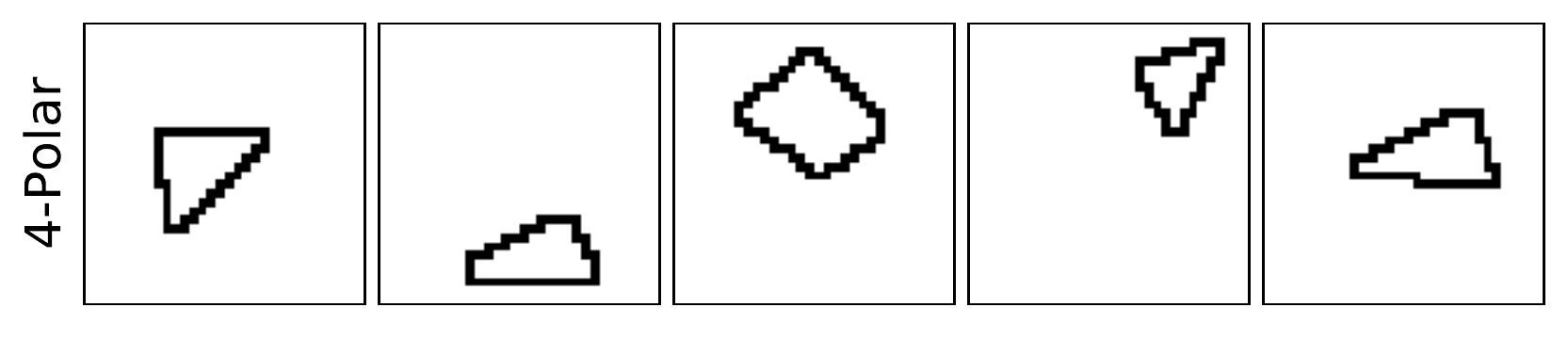}\\ [-0.09cm]
    \includegraphics[width=0.42\textwidth]{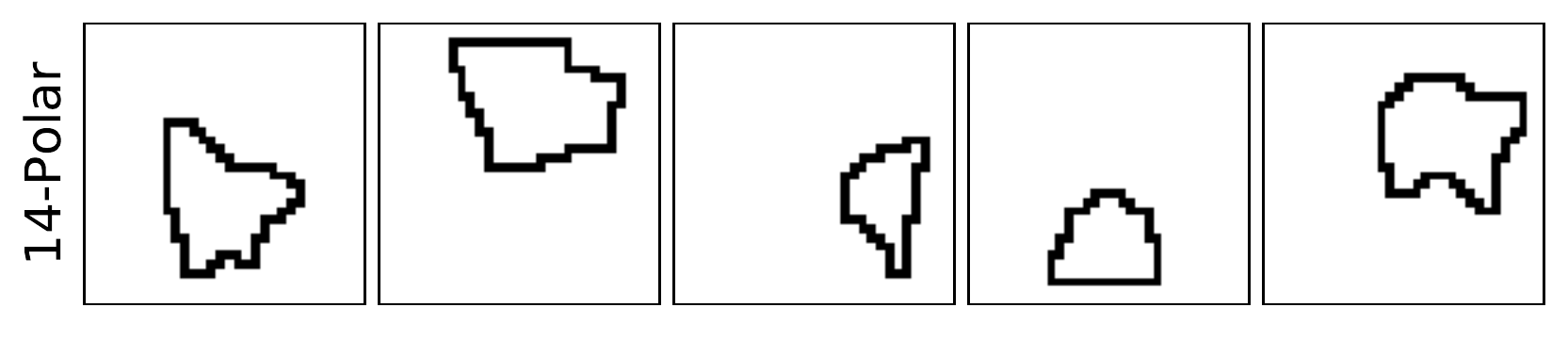}\\ [-0.09cm]
    \includegraphics[width=0.42\textwidth]{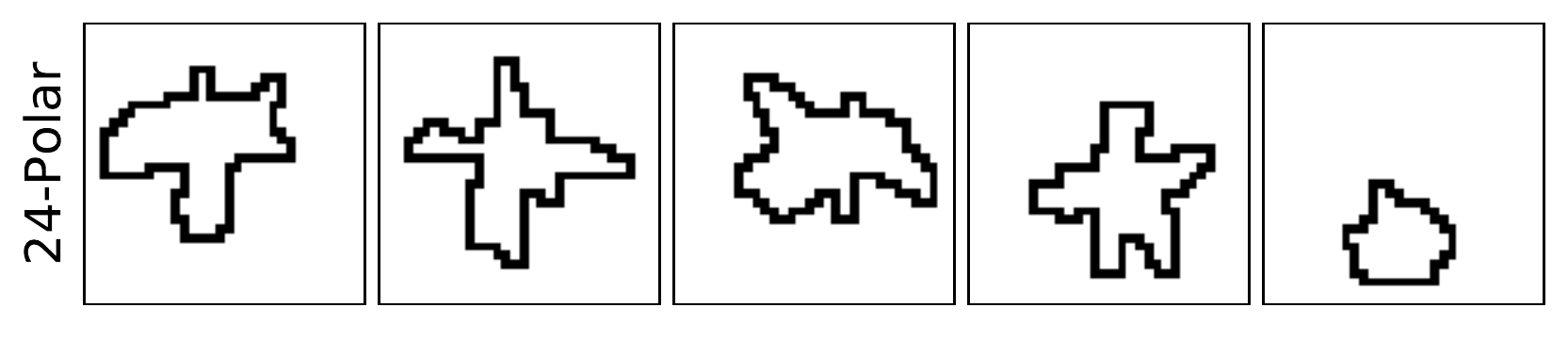}\\ [-0.09cm]
    \includegraphics[width=0.42\textwidth]{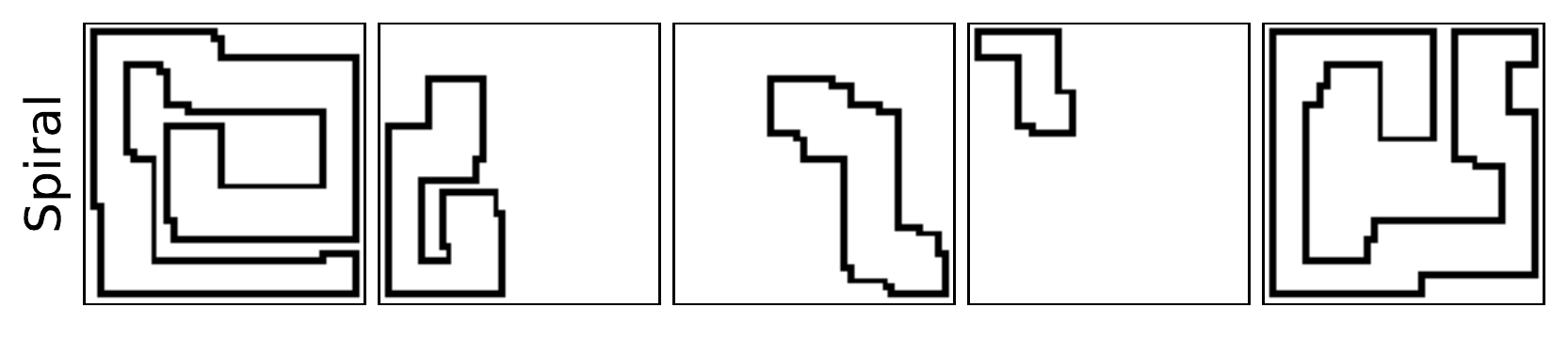}\\ [-0.09cm]
    \includegraphics[width=0.42\textwidth]{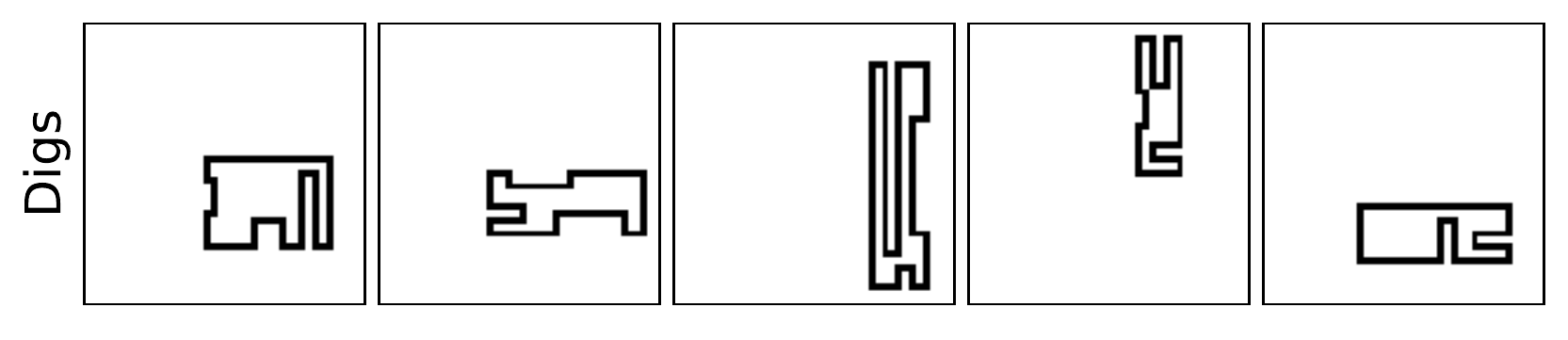}\\
    \footnotesize (a)
      \end{tabular}} &
    \includegraphics[width=0.27\textwidth]{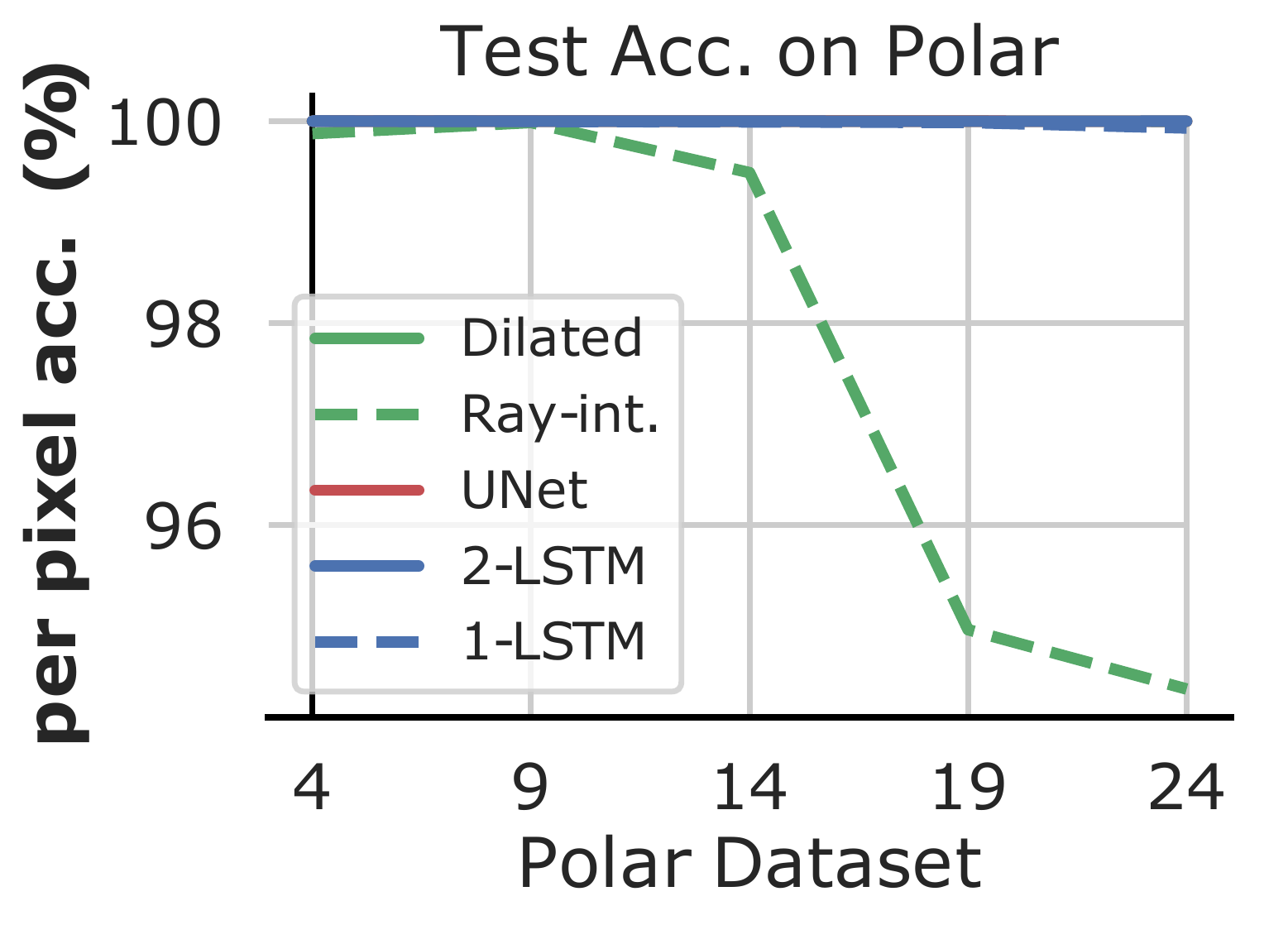}&
    \includegraphics[width=0.27\textwidth]{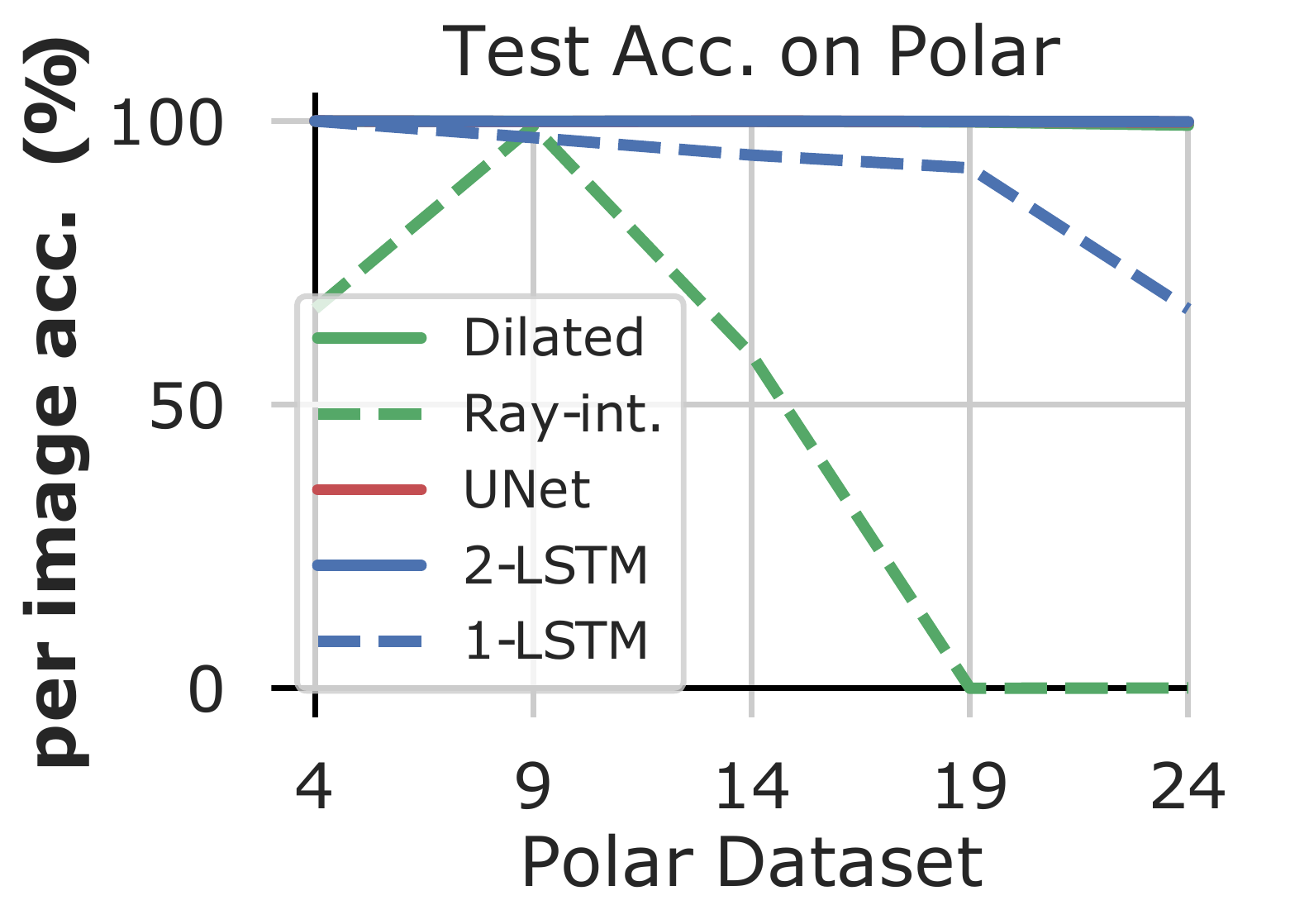}\\

     &
     \footnotesize (b)&
    \footnotesize (c)\\[0.3cm]
    &
    \includegraphics[width=0.27\textwidth]{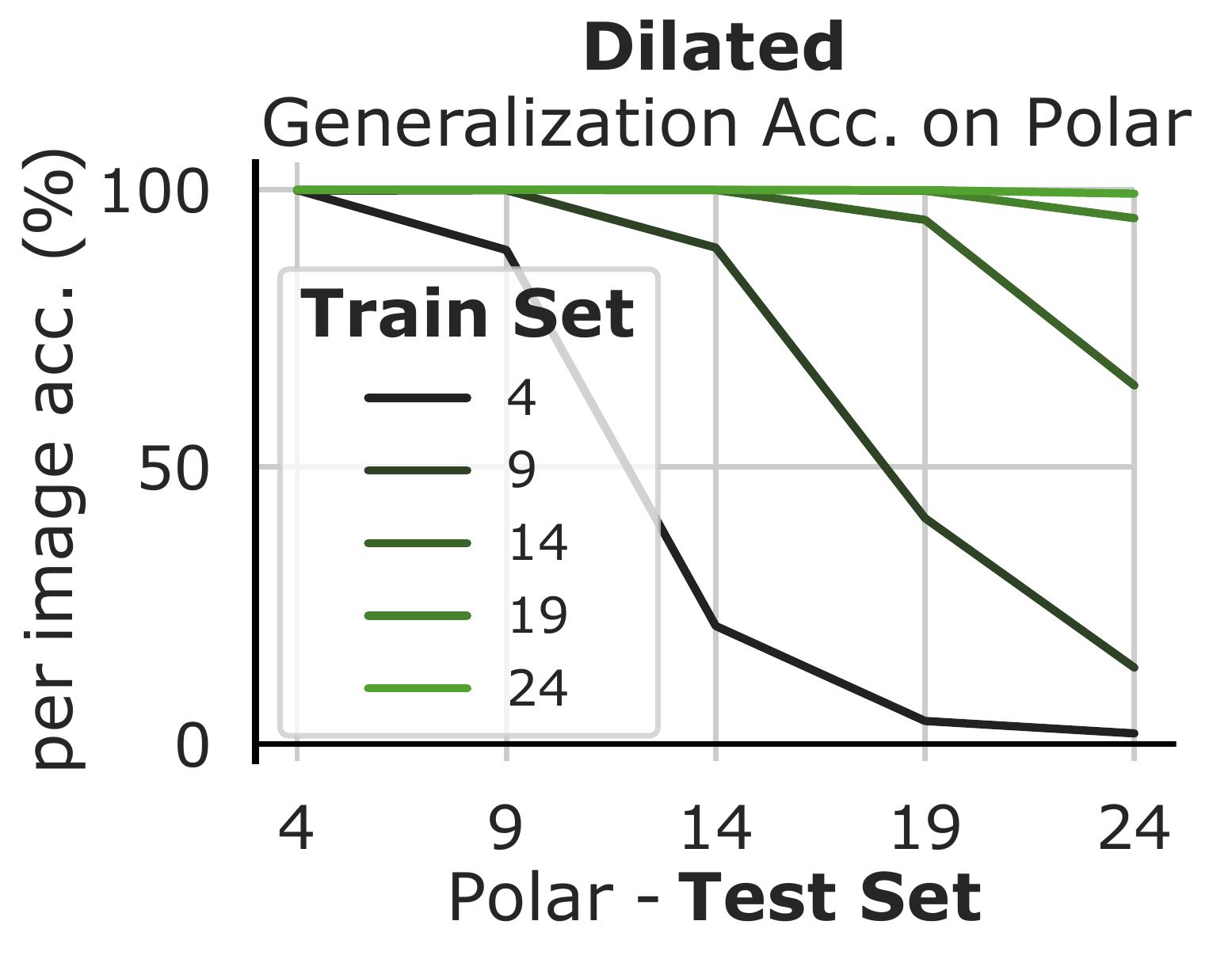}&
    \includegraphics[width=0.27\textwidth]{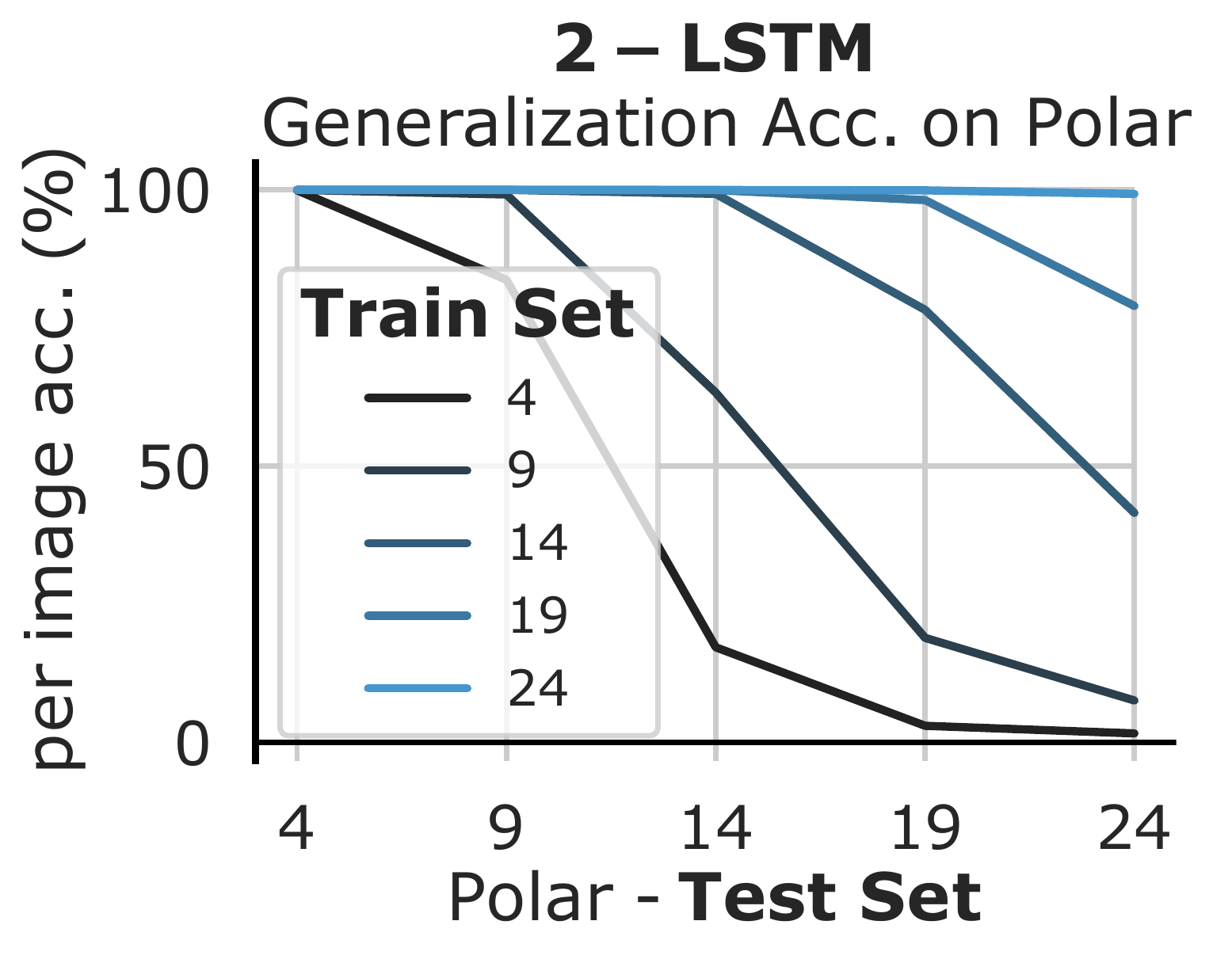}\\
     &  \footnotesize (d) &  \footnotesize (e) \\
    \end{tabular}
  \caption{\emph{Datasets and Results in Polar.} (a) Images of the curves used to train and test the DNNs. Each row correspond to a different dataset.
  (b) and (c) Intra-dataset evaluation using per pixel accuracy and  per image accuracy, respectively. (d) and (e) Evaluation using the testing set of each Polar datasets for \emph{Dilated} and \emph{2-LSTM} networks, respectively.}\vspace{-0.cm}
\label{fig:Datasets}
\end{figure}

\renewcommand{\arraystretch}{0}

\begin{figure*}[t]
  \footnotesize
  \begin{tabular}{@{\hspace{0.25cm}}c@{\hspace{-0.2cm}}c@{\hspace{-0.2cm}}c@{\hspace{-0.2cm}}c@{\hspace{-0.25cm}}c}
        \includegraphics[width=1.42in]{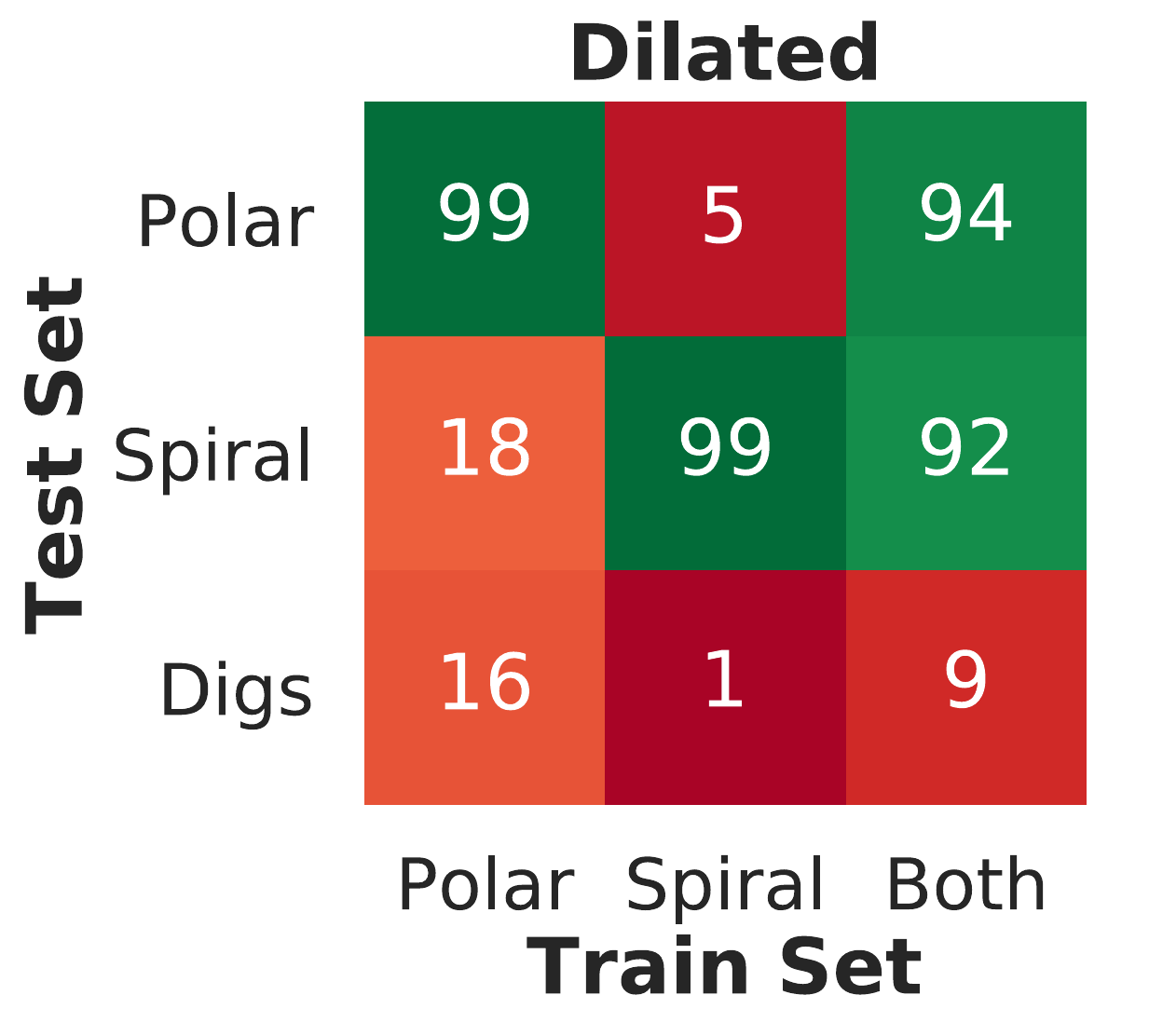}&
        \includegraphics[width=1.05in]{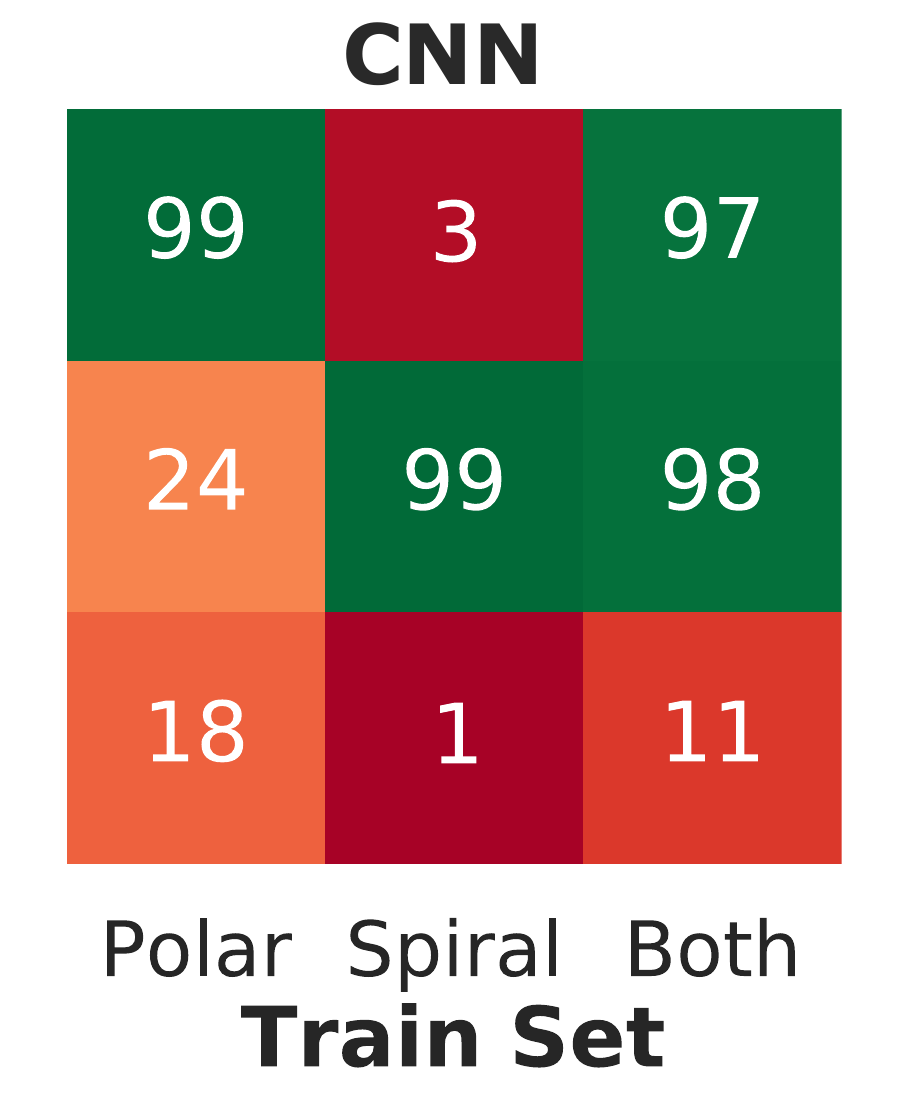}&
        \includegraphics[width=1.05in]{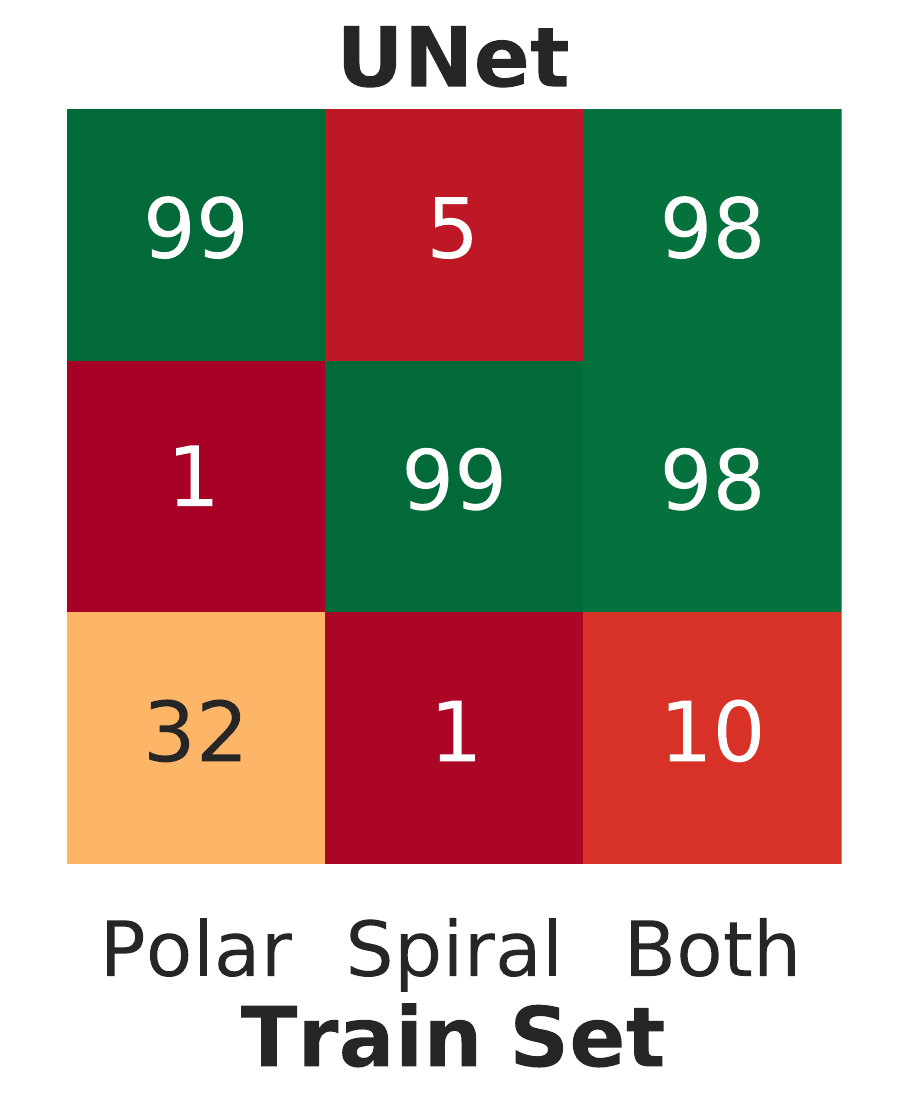}&
        \includegraphics[width=1.05in]{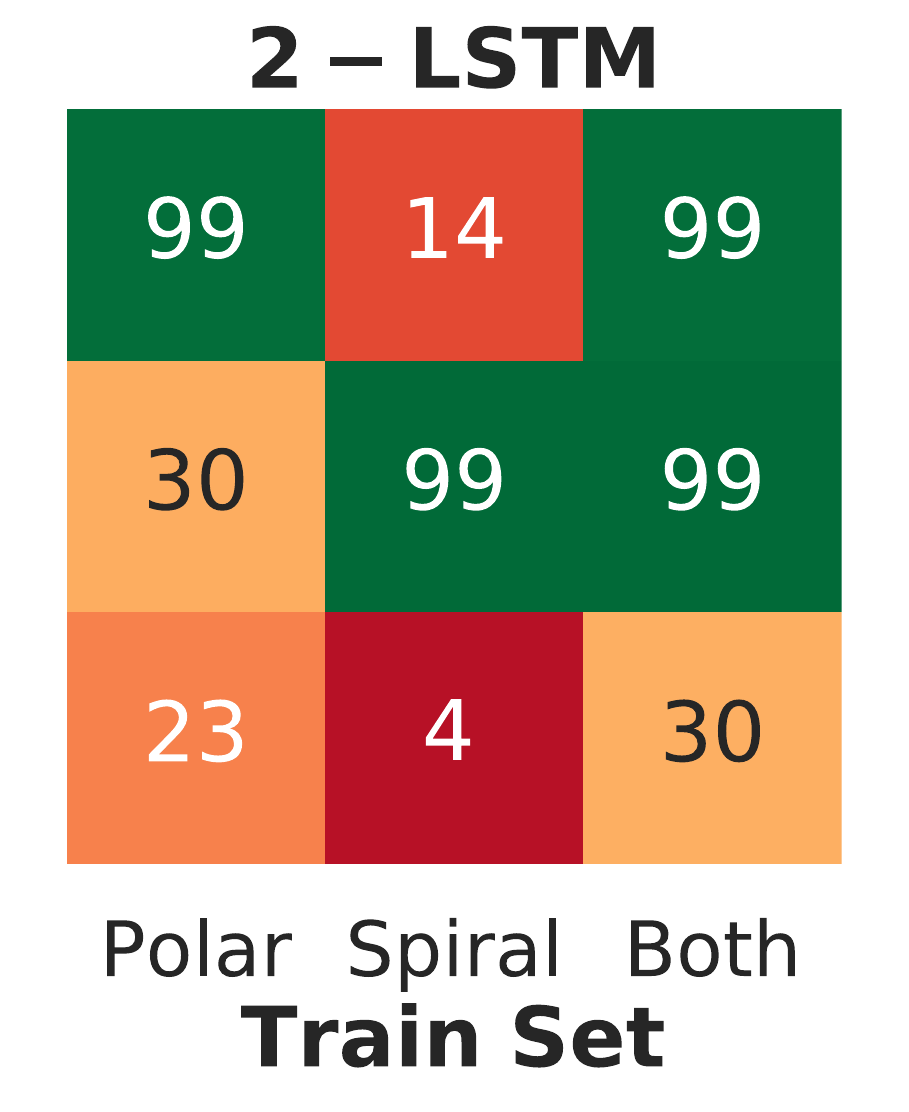}&
        \includegraphics[width=1.33in]{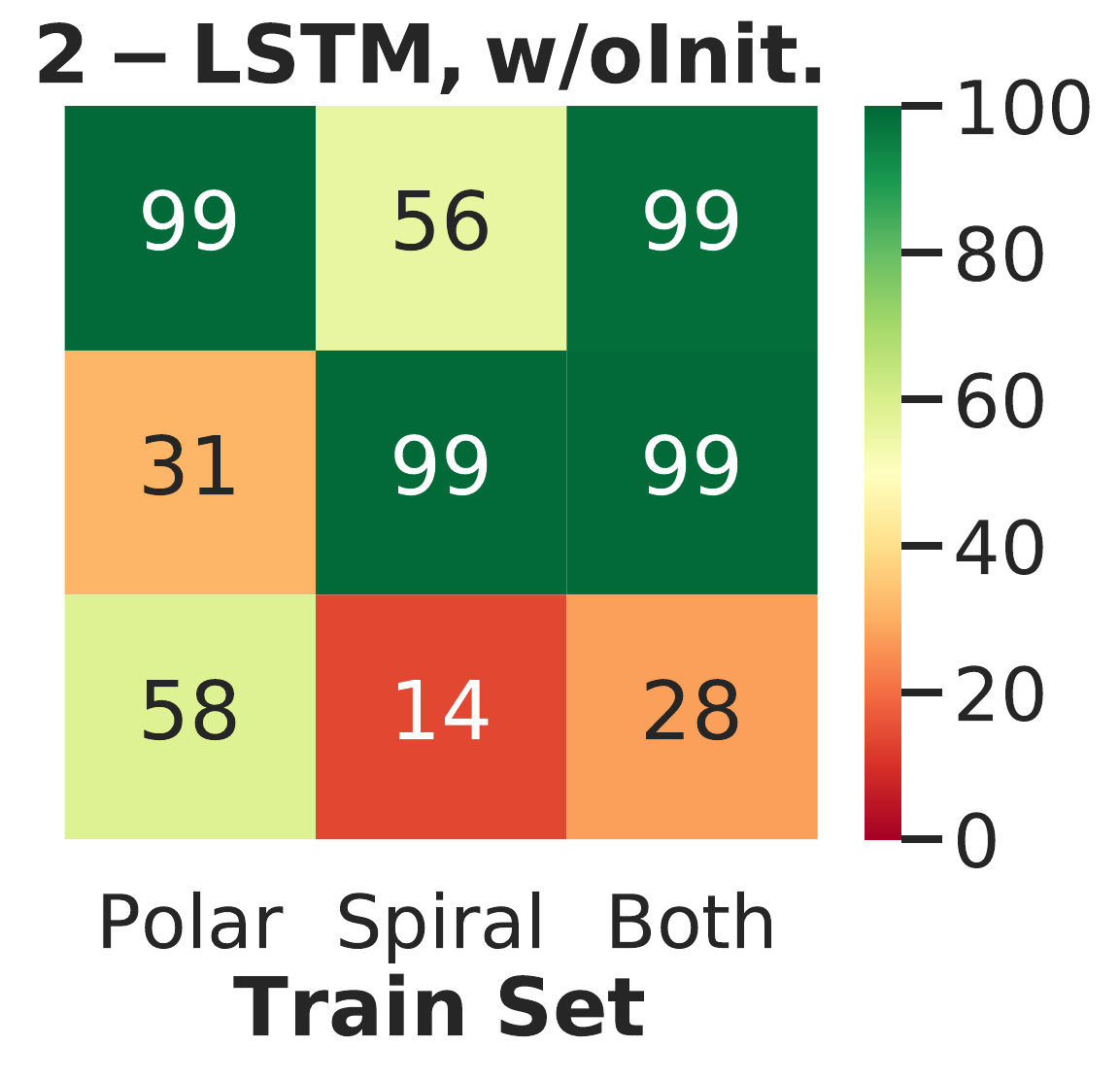}\\
  \end{tabular}
\caption{\emph{Cross-dataset Results.} Evaluation of the networks trained in $24$-Polar, Spiral and both $24$-Polar and Spiral datasets. The tesing sets are $24$-Polar, Spiral and Digs datasets.}
\label{fig:General2}
\end{figure*}

\subsection{Results}

We evaluate the DNNs trained in the standard manner,~\ie using backpropagation with the labeled datasets as described previously, to predict insideness for every pixel.

\noindent {\bf Intra-dataset Evaluation.} In Figs.~\ref{fig:Datasets}b and c we show per pixel and per image accuracy for the networks trained on the same Polar dataset that are being tested. \emph{Dilated}, \emph{2-LSTM} and \emph{UNet} achieve a testing accuracy very close to $100\%$, but \emph{Ray-int.} and \emph{1-LSTM} perform much worse. Training accuracy of \emph{Ray-int.} and \emph{1-LSTM} is the same as their testing accuracy (Figs.~\ref{resSuppQuant1}a and~b in Appendix~\ref{Additional_Figures_and_Visualizations}). This indicates an optimization problem similar to the cases reported by~\cite{SOS17}, as both \emph{Ray-int.} and \emph{1-LSTM} are complex enough to generalize. It is an open question to understand why backpropagation performs so differently in each of these architectures that can all generalize in theory. Finally, note that the per pixel accuracy is in most cases very high, and from now on, we only report the per image accuracy.


\noindent {\bf Cross-dataset Evaluation.} We evaluate if the networks that have achieved very high accuracies (\emph{Dilated}, \emph{2-LSTM} and \emph{UNet}), have learnt the general solution of insideness that we introduced in Section~\ref{SecRay-Intersection}. To do so, we train on one dataset and test on the different one. In Figs.~\ref{fig:Datasets}d and e, we observe that \emph{Dilated} and \emph{2-LSTM} do not generalize to Polar datasets with larger amount of vertices than the Polar dataset on which they were trained. Only if the networks are trained in 24-Polar, the networks generalize in all the Polar datasets. The same conclusions can be extracted for~\emph{UNet} (Fig.~\ref{resSuppQuant1}c). 


\renewcommand{\arraystretch}{0}

\begin{figure}[t!]
  \begin{tabular}{ll}
    \footnotesize
     {\bf  \quad  Dilated  \ \ \ \ \ \ \ \ CNN   \ \ \ \ \ \ \ \ \ UNet  \  \ \quad  2-LSTM} & 
    \footnotesize
     {\bf  \quad  Dilated  \ \ \ \ \ \ \ \ CNN   \ \ \ \ \ \ \ \ \ UNet  \  \ \quad  2-LSTM} \\ 
    \includegraphics[width=0.47\textwidth]{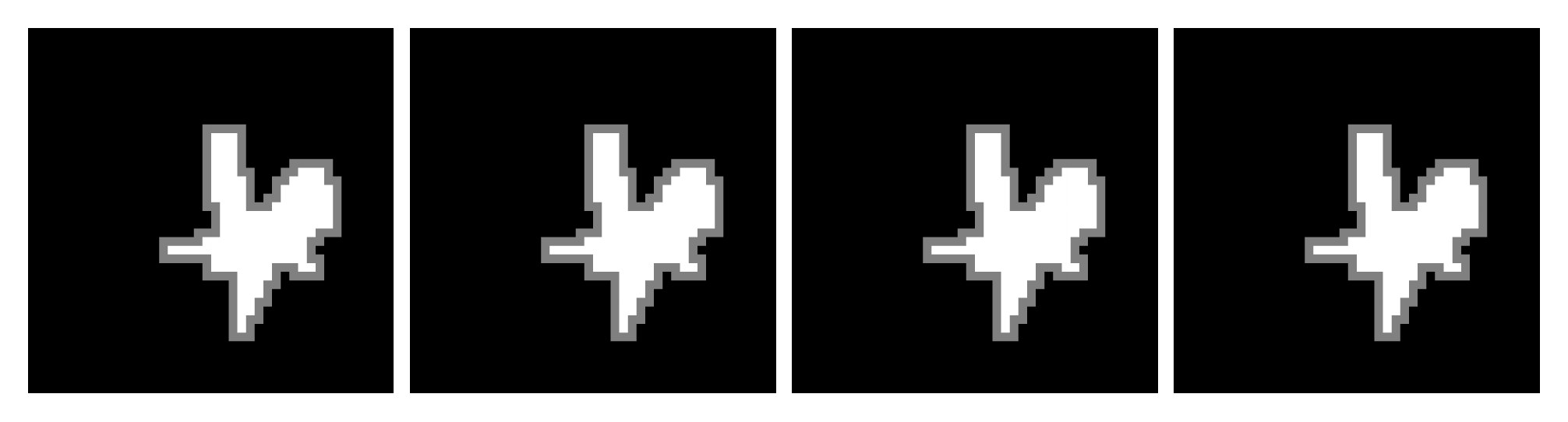} & 
    \includegraphics[width=0.47\textwidth]{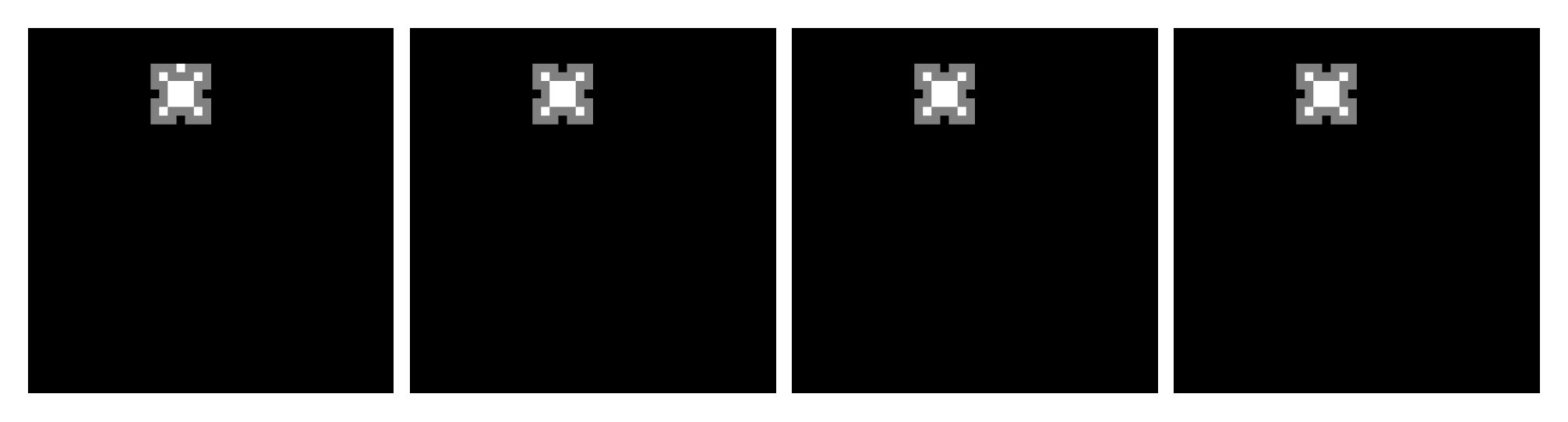} \\ [-0.15cm]
    \includegraphics[width=0.47\textwidth]{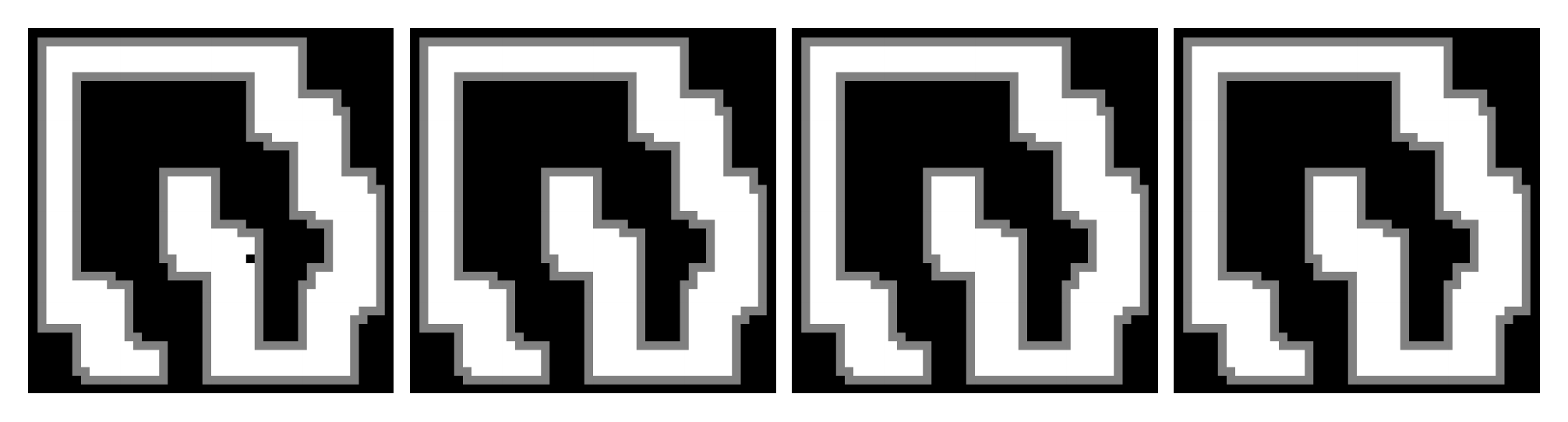} & 
    \includegraphics[width=0.47\textwidth]{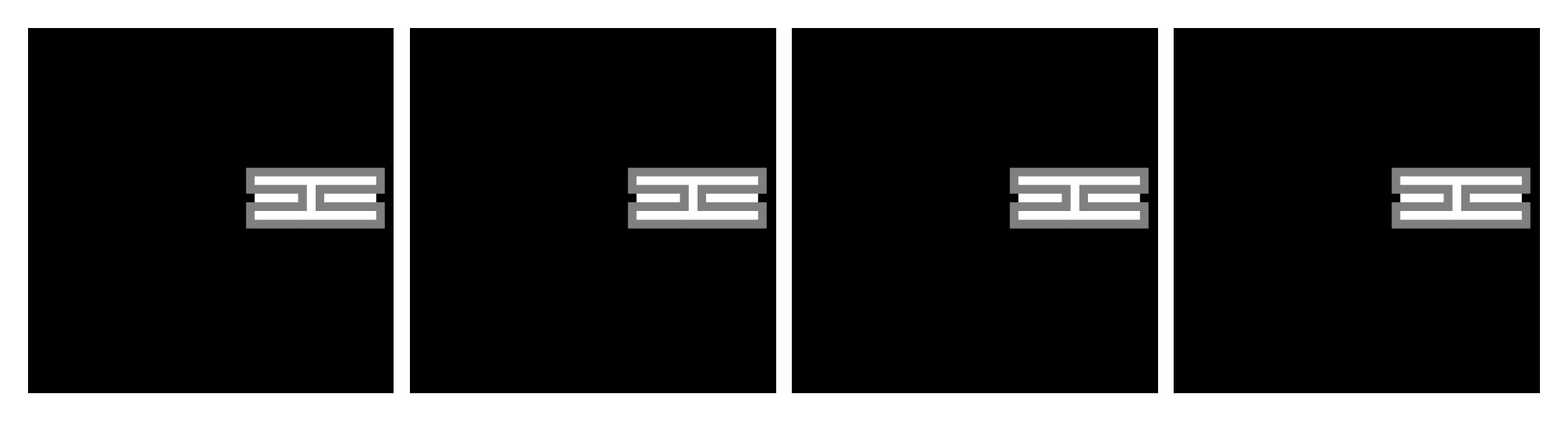}\\ [-0.15cm]
    \includegraphics[width=0.47\textwidth]{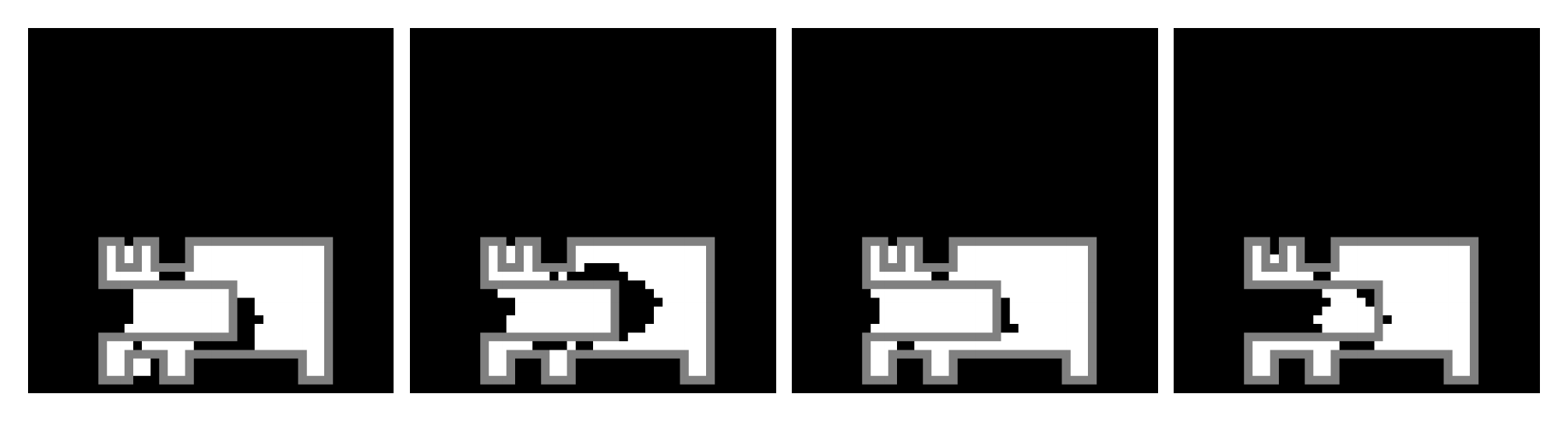} & 
    \includegraphics[width=0.47\textwidth]{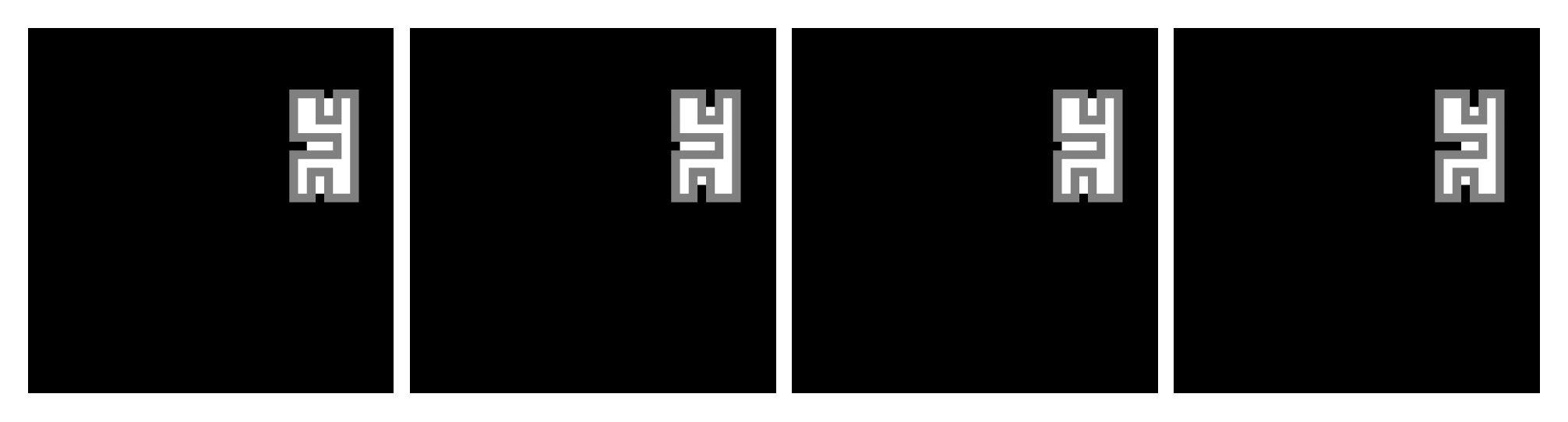}\\ 
  \end{tabular}
\caption{\emph{Qualitative Examples.} Networks trained in $24$-Polar and Spiral dataset fail to segment in the Digs dataset.}
 \label{exSuppQuak}
\end{figure}

We further test generalization capabilities of these networks beyond the Polar dataset. In this more broad analysis, we also include the \emph{CNN} and \emph{2-LSTM w/o init}, by training them on $24$-Polar, Spiral and both $24$-Polar and Spiral, and test them on $24$-Polar, Spiral and Digs separately. We can see in Fig.~\ref{fig:General2} that all tested networks generalize to new curves of the same family as the training set. Yet, the networks do not generalize to curves of other families. In Fig.~\ref{exSuppQuak}, we show qualitative examples of failed segmentations produced by networks trained on $24$-Polar and Spiral and tested on the Digs dataset.

Furthermore, note that using a more varied training set (``Both'') does not necessarily lead to better cross-dataset accuracy. For example, for \emph{UNet} and \emph{2-LSTM w/o init.}, training on Polar achieves better accuracy in Digs than when training on ``Both''. Also, for \emph{Dilated}, training on ``Both'' harms its accuracy: the accuracy drops more than $6\%$ in $24$-Polar and Spiral. In this case, the training accuracy is close to $100\%$, which indicates a problem of overfitting. We tried to address this problem by regularizing using weight decay, but it did not improve the accuracy (Appendix~\ref{secSuppDilated}).   Finally, we also tried fine-tuning to our datasets and training from scratch the state-of-the-art networks in segmentation benchmarks in natural images (DEXTR~\citep{maninis2018deep} and DeepLabv3+~\citep{chen2018deeplabv3+}), but they also failed to generalize (Appendix~\ref{secSuppNatural}). In Section~\ref{sec:small}, we will show that the lack of generalization in recurrent networks can be alleviated. Yet, for feed-forward networks it remains an open question whether there is a regularizer or an inductive bias that could facilitate learning the general solution based on the ray-intersection method.

\renewcommand{\arraystretch}{0}

\begin{figure*}[t]
  \footnotesize
  \centering
  \begin{tabular}{@{\hspace{-0.04cm}}c@{\hspace{-0.09cm}}c|c@{\hspace{-0.09cm}}c}
\multicolumn{2}{c}{\bf Layer 2} &\multicolumn{2}{c}{\bf Layer 6}\\[0.1cm]
    { Unit 3}&  { Unit 19}& { Unit 15}& { Unit 27}\\
        \includegraphics[width=0.245\textwidth]{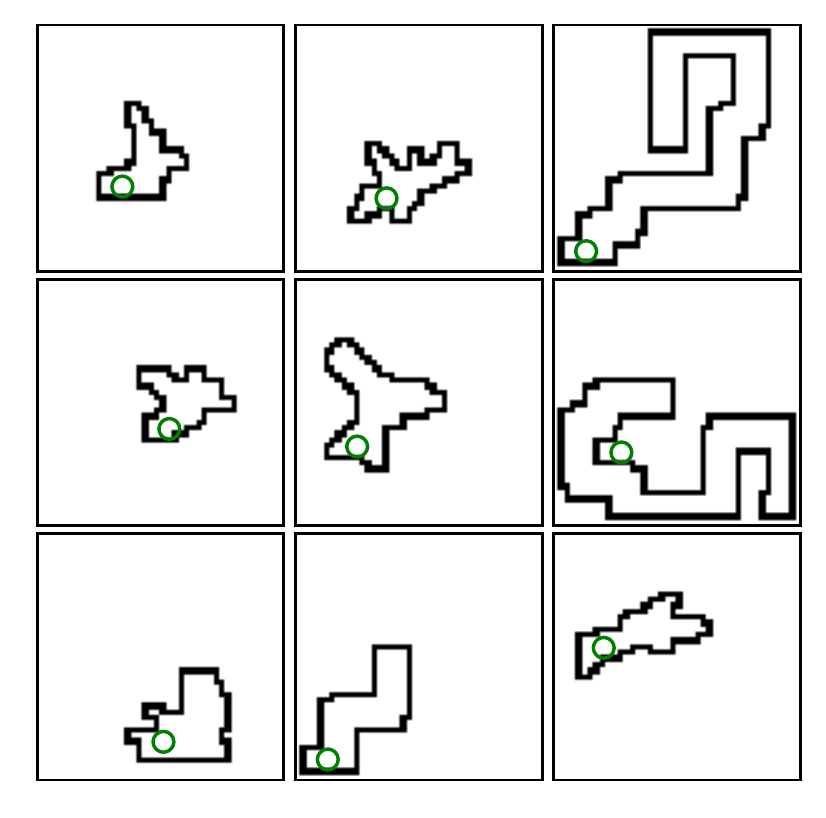}&
    \includegraphics[width=0.245\textwidth]{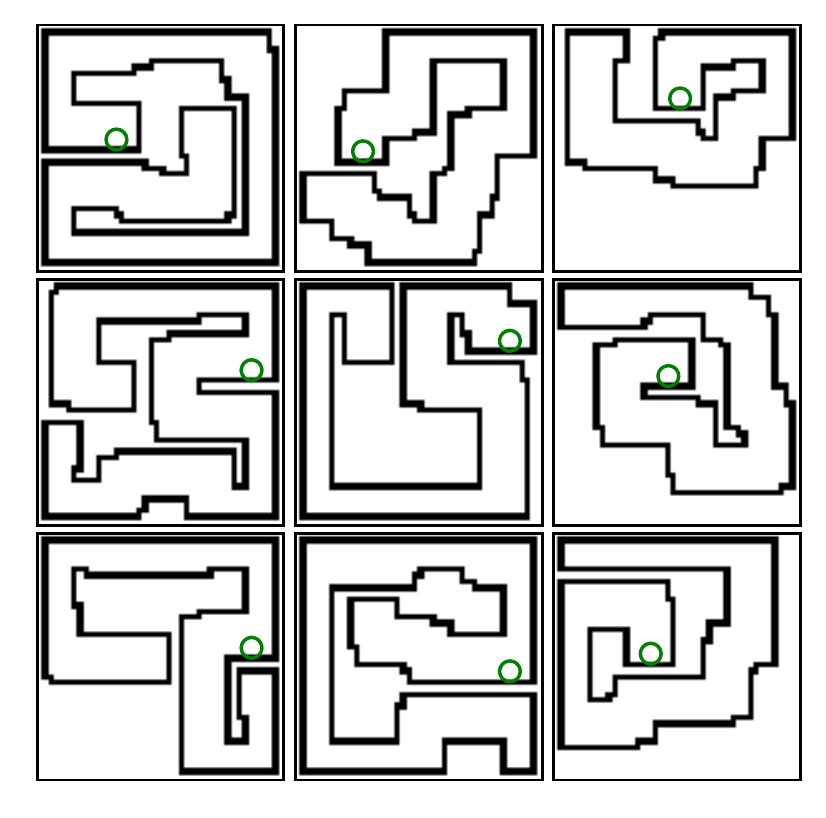}&
    \includegraphics[width=0.245\textwidth]{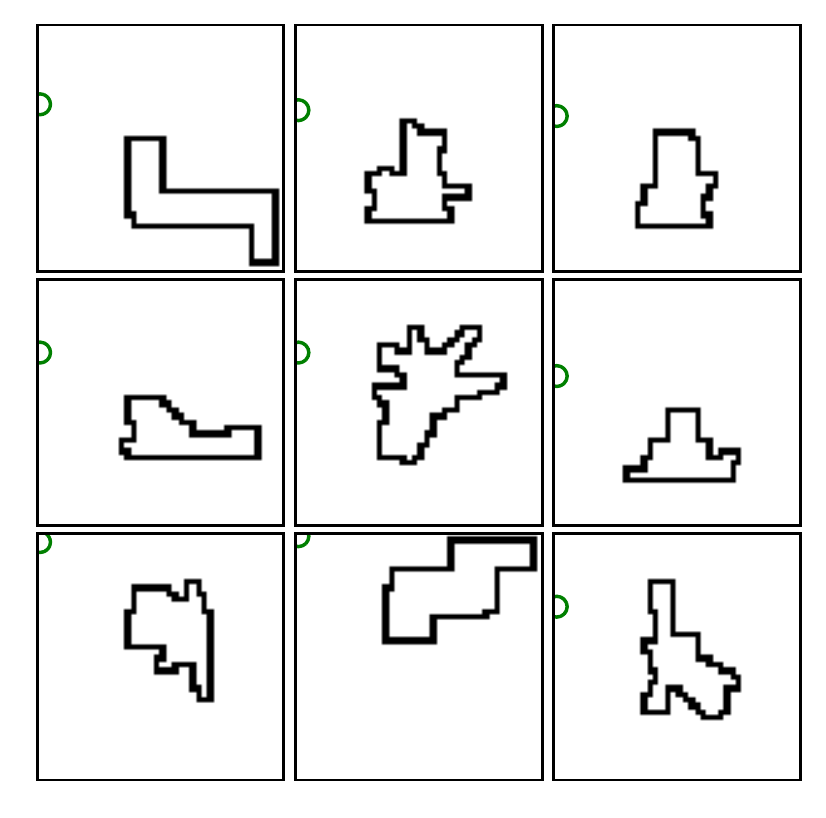}&
    \includegraphics[width=0.245\textwidth]{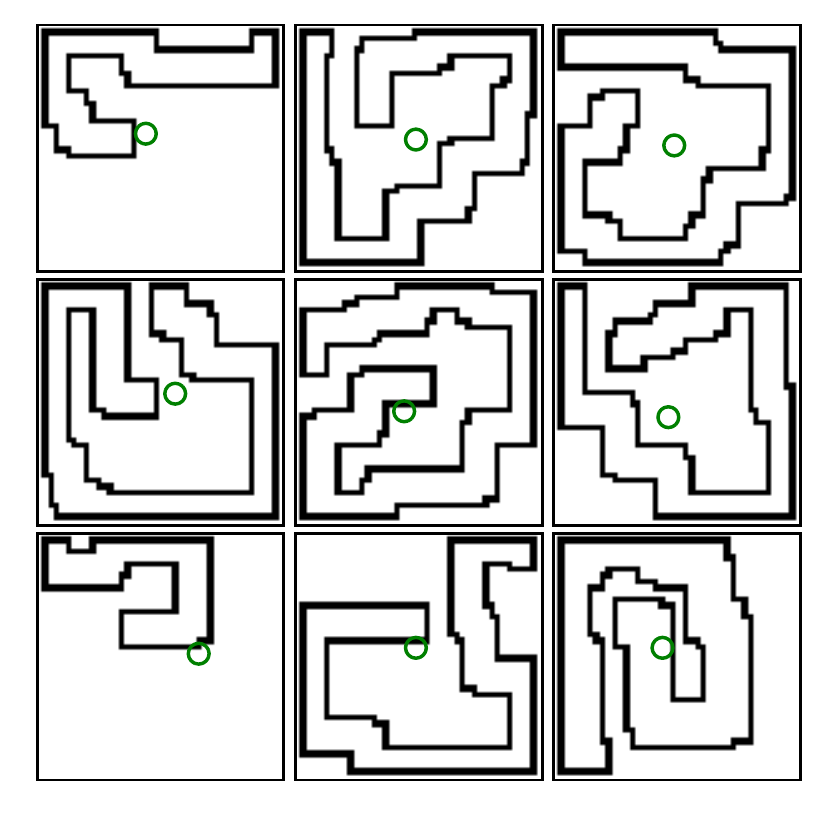}
  \end{tabular}
\caption{\emph{Visualization of the Units Learnt by~\emph{Dilation}.} Each block are the $9$ images that produce the maximum activation of a units in a convolutional layer across the test set. The green dot indicates the location of the unit in the feature map. Fig.~\ref{visSuppDilation} in Appendix~\ref{Additional_Figures_and_Visualizations} shows more examples.}
\label{visDilation}
\end{figure*}

\noindent {\bf Understanding the Lack of Generalization.} We visualize the networks to understand why the representations learnt do not generalize. 
In Fig.~\ref{visDilation} and Fig.~\ref{visSuppDilation} (Appendix~\ref{Additional_Figures_and_Visualizations}), 
we analyze different units of \emph{Dilated} trained on $24$-Polar and Spiral. We display  units of the same kernel from the second, four and sixth layers, by showing the nine images in the testing set that produce the unit to be most active across all images~\citep{zeiler2014visualizing}. For each image, we indicate the unit location in the feature map by a green circle. The visualizations suggest that units of the second layer are tuned to local features (\eg Unit~$19$ is tuned to close parallel lines), while in layer 6 they are tuned to more global ones (\eg Unit~$27$ captures the space left in the center of a spiral). Thus, the units are tuned to  characteristics of the curves in the training set rather than to features that could capture the the long-range dependencies to solve insideness, such as the ones we derived theoretically. 

\renewcommand{\arraystretch}{0}

\begin{figure}[t!]

  \centering
   {\bf Input Images}
   
    \includegraphics[width=0.5\textwidth]{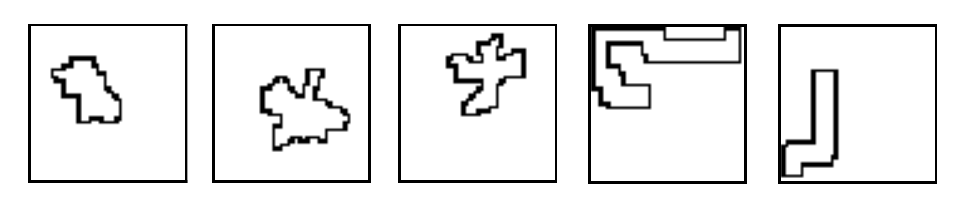}
    
    \centering
      \vspace*{0.3cm}
    \begin{tabular}{cc}
    {\bf Analytically-derived Coloring Routine} &  {\bf 2-LSTM Trained on Polar and Spiral}   \\[0.25cm]
    
    \scriptsize
   {\bf t=0}  \ \ \ \   \quad {\bf t=4}  \ \ \ \ \ \ \ \ {\bf t=8} \quad \   \ \ \ {\bf t=12} \quad  \ \ \ {\bf t=16} \quad   \ \ {\bf t=20} &  
    \scriptsize
     {\bf t=0}  \ \ \ \  \quad {\bf t=4}  \ \ \ \ \ \ \ \ {\bf t=8} \quad \   \ \ \ {\bf t=12} \quad  \ \ \ {\bf t=16} \quad   \ \ {\bf t=20} \\
     \includegraphics[width=0.47\textwidth]{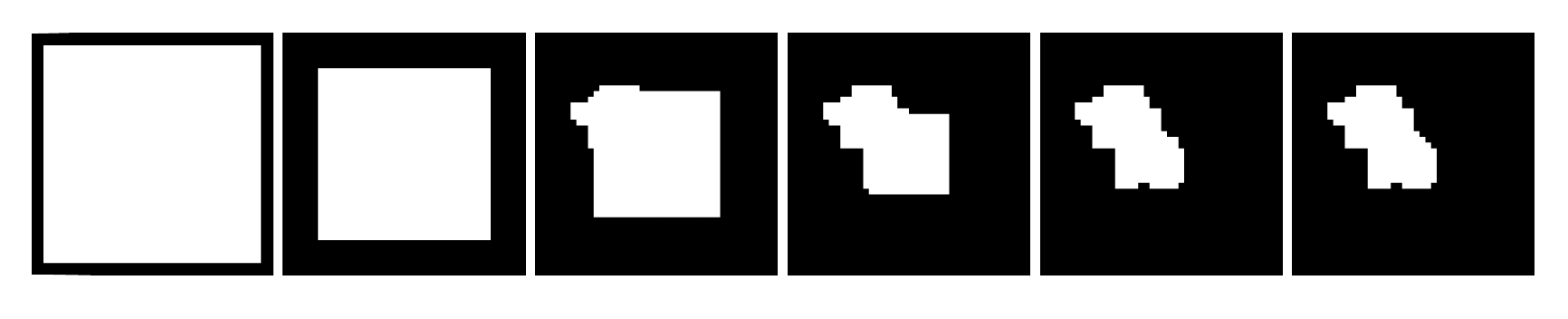} &\includegraphics[width=0.47\textwidth]{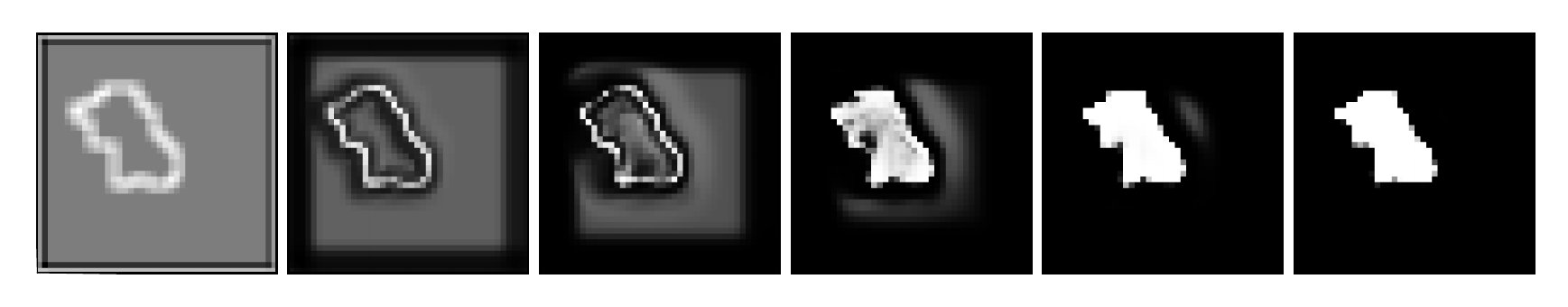}\\  [-0.2cm]
     
  \includegraphics[width=0.47\textwidth]{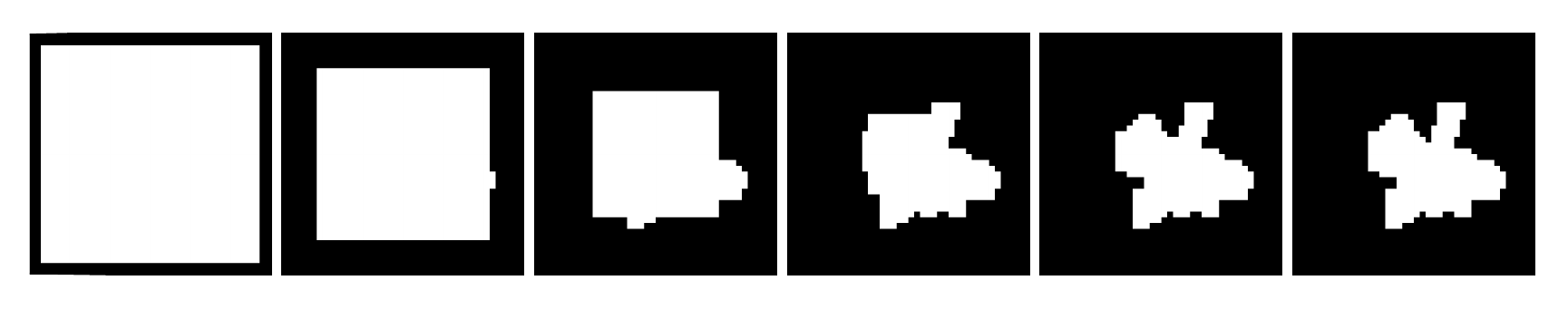} &\includegraphics[width=0.47\textwidth]{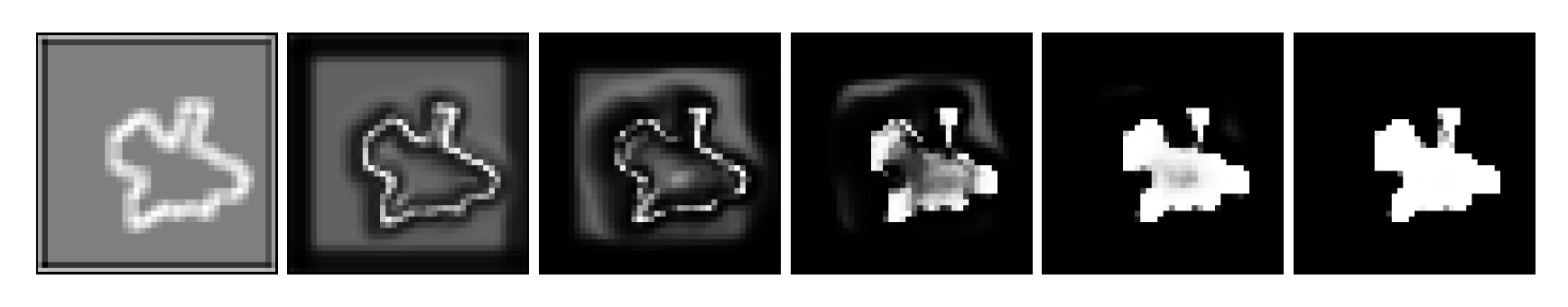}\\  [-0.2cm]
  
   \includegraphics[width=0.47\textwidth]{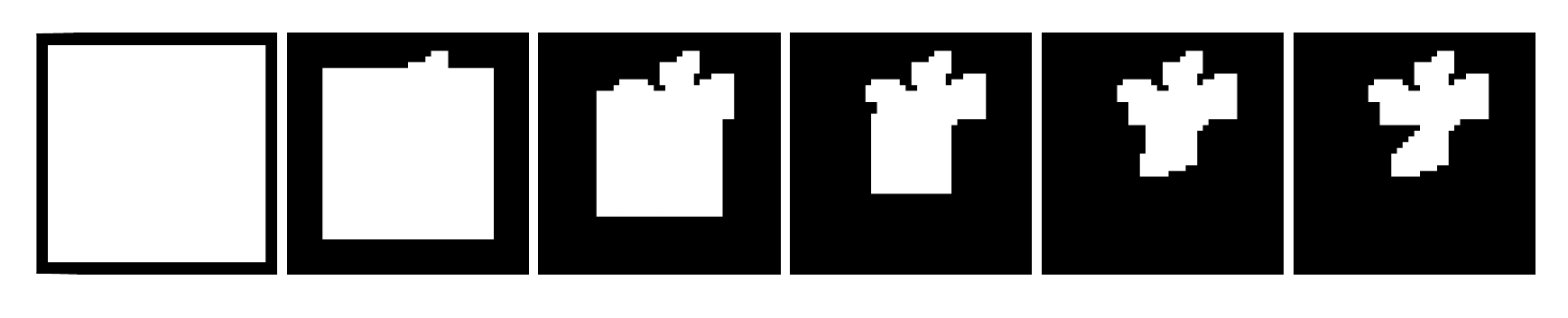} &\includegraphics[width=0.47\textwidth]{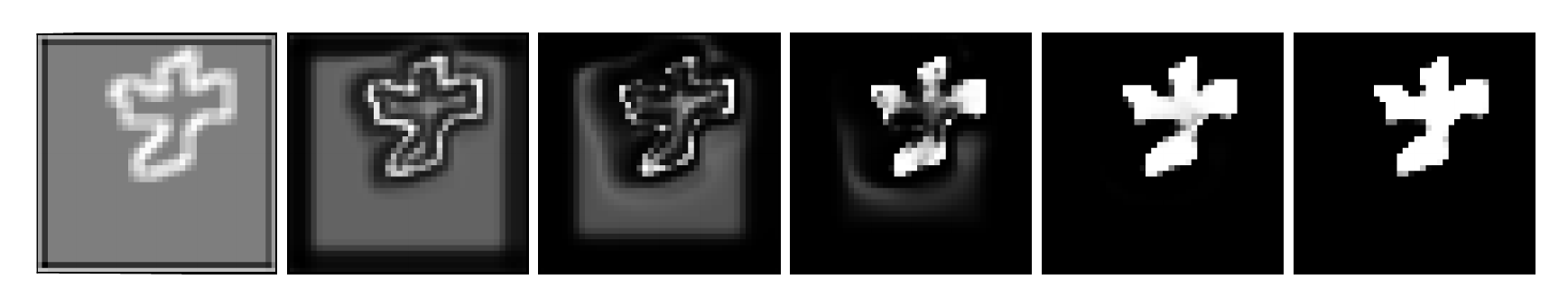}\\  [-0.2cm]
   
\includegraphics[width=0.47\textwidth]{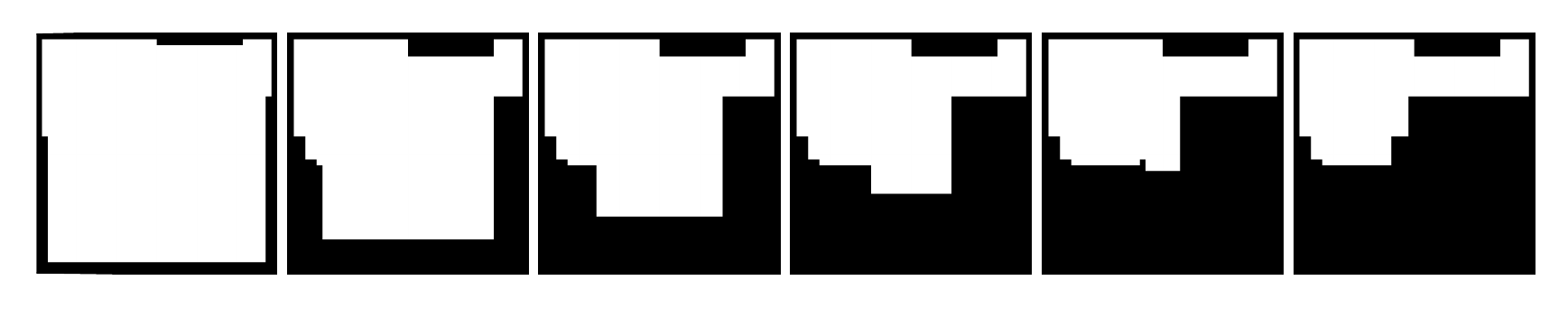} &\includegraphics[width=0.47\textwidth]{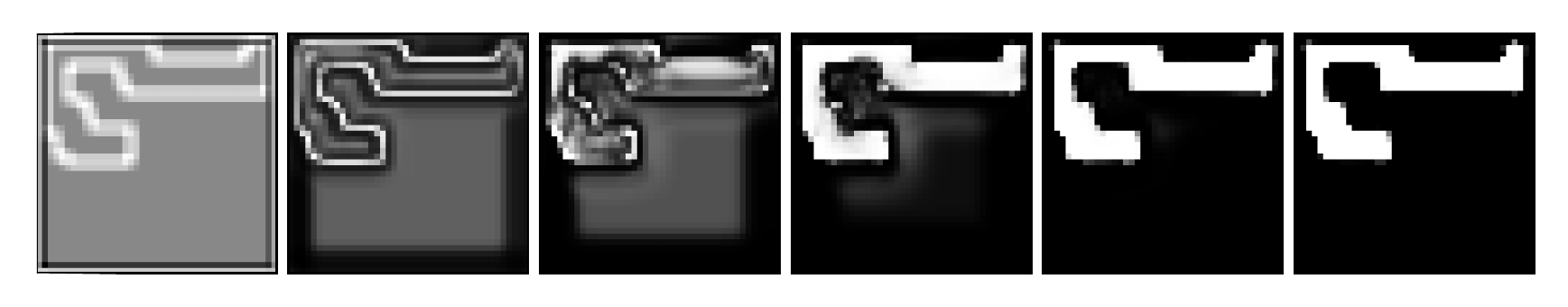}\\  [-0.2cm]

\includegraphics[width=0.47\textwidth]{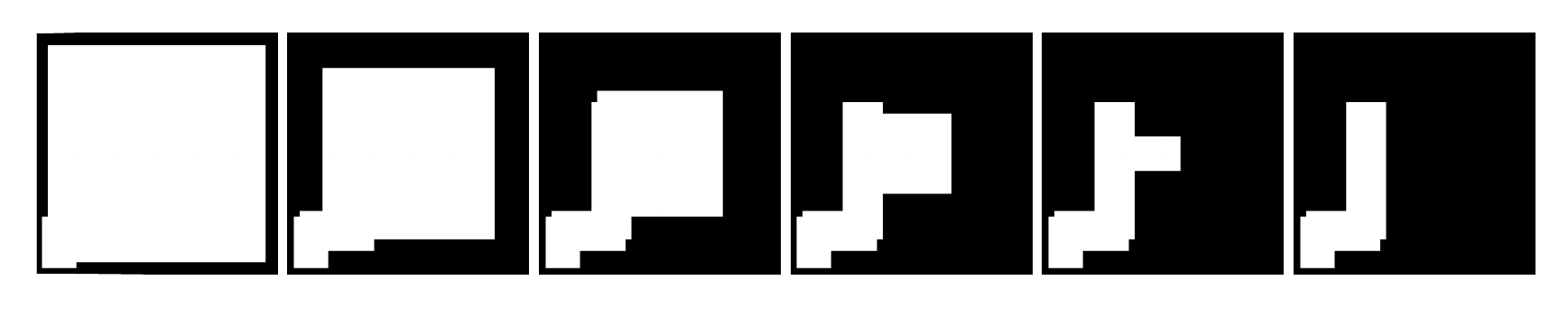} &\includegraphics[width=0.47\textwidth]{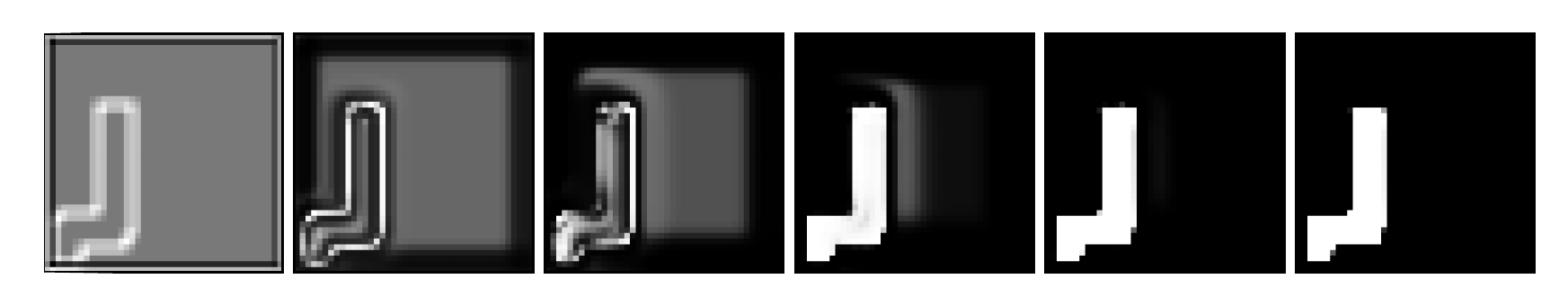}\\
  \end{tabular}

  \centering
  \vspace*{0.3cm}
  \begin{tabular}{@{\hspace{-0.05cm}}c}
  {\bf 2-LSTM Trained on Small Images}\\[0.25cm]
    \scriptsize
    \hspace{-1cm}
  \ \  \ \ \ \ \ \ \
 \ \  \ \   \  \ \quad {\bf t=0}
      \ \ \ \ \quad {\bf t=5}
       \ \ \ \quad {\bf t=10}
       \ \ \quad {\bf t=15}
       \ \ \ \quad {\bf t=20}
       \ \ \ \quad {\bf t=25} 
       \ \ \ \quad {\bf t=30}
       \ \ \ \quad {\bf t=35}
       \ \ \ \quad {\bf t=40}
       \ \ \ \quad {\bf t=45}
       \ \ \quad {\bf t=50}
       \ \ \ \quad {\bf t=55}\\
    \hspace{-1.2cm} \includegraphics[width=0.99\textwidth]{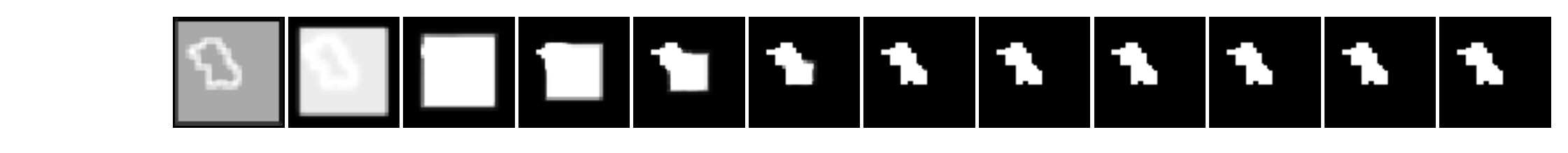} \\ [-0.2cm]
   \hspace{-1.2cm}      \includegraphics[width=0.99\textwidth]{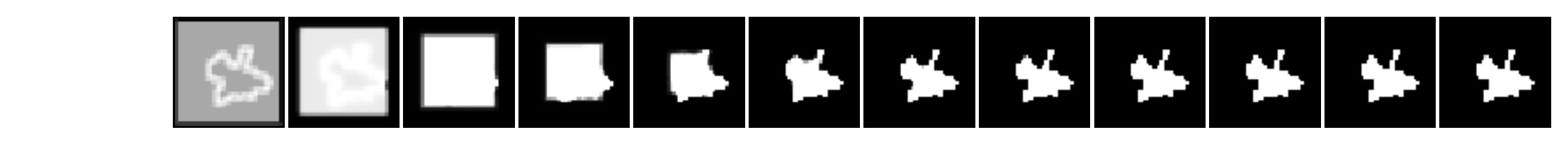} \\[-0.2cm]
    \hspace{-1.2cm}     \includegraphics[width=0.99\textwidth]{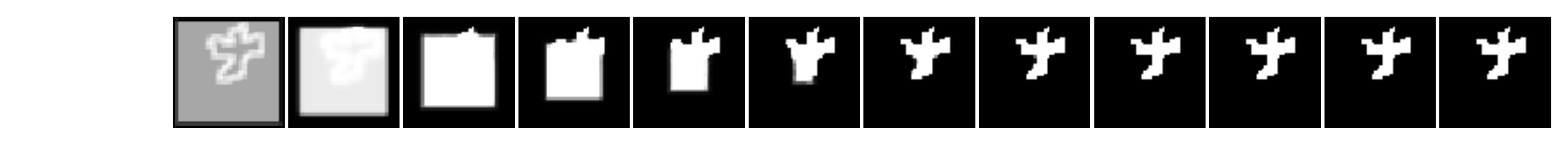} \\[-0.2cm]
  \hspace{-1.2cm} \includegraphics[width=0.99\textwidth]{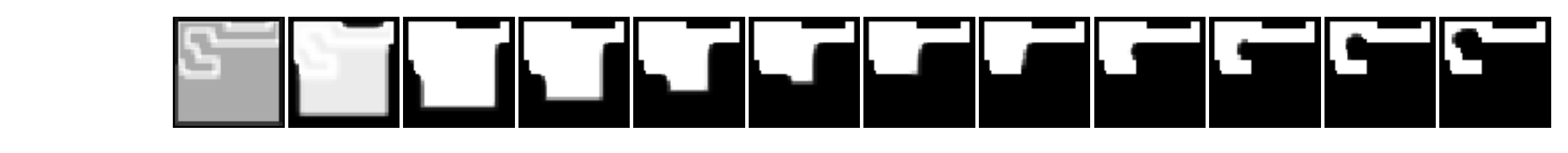} \\[-0.2cm]
    \hspace{-1.2cm}    \includegraphics[width=0.99\textwidth]{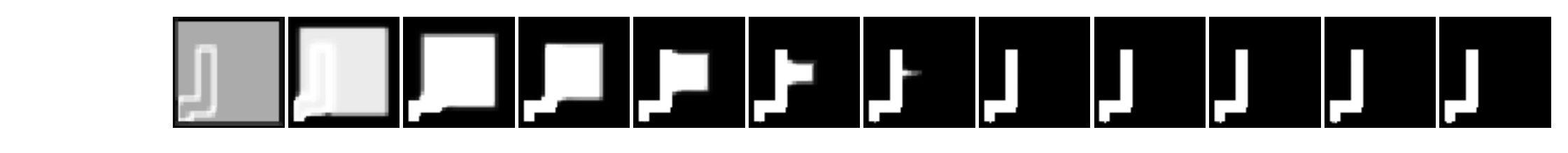} \\

  \end{tabular}
\caption{\emph{Output of the \emph{Coloring Routine}, the \emph{2-LSTM} Trained on Polar and Spiral Datasets, and the  \emph{2-LSTM} Trained on Small Images ($18\times 18$ pixels).} Each row corresponds to the output of the network for different images and each column to a different time step. Note that the 2-LSTM trained with small images implements the coloring routine because it expands the ``outside'' region and blocks the expansion at the border of the curve, while the other 2-LSTM does not effectively implement the coloring routine.}
\label{visSuppLSTM}
\end{figure}


In Fig.~\ref{visSuppLSTM} top row, we display the feature maps of  \emph{2-LSTM} trained on $24$-Polar and Spiral and the analytically-derived coloring routine. The figure shows the output of the network at different time steps. We can see that the network expands the borders of the image, which have been initialized to outside. Yet, the network expands the curve in both directions, indicating that the blocking operation has not been learned. Also, note that it is impossible to know the direction of expansion from the curve with only  local operations, as in one step of the ConvLSTM. 
The analytical solution only expands the borders of the image. 





\clearpage
\section{Is There a Training Paradigm That Leads to Learning General Solutions for Insideness?}

\label{sec:small}
We now introduce a strategy to train convolutional recurrent networks that leads to excellent generalization across datasets as an implementation of the \emph{coloring routine} emerges in the network.

Recurrent networks trained by backpropagation through time can be thought of as feed-forward networks with time steps being applied as subsequent layers,~\ie backpropagation is applied to the  ``unrolled''  version of the recurrent network.
Recall that for the insideness problem the prediction error is evaluated after the last unrolling step. Thus, depending on the image size and the shape of the curve a large number of unrolling steps may be required to train the network (\eg~the number of unrolling steps for the networks shown in Section~\ref{SecColoring} is bounded by $N^2$). It is well known that for a large number of unrolling steps backpropagation through time has difficulties capturing long-range relationships~\citep{bengio1994learning}. Moreover,  the memory requirements are proportional to the number of unrolling steps, making it difficult  to scale in practice.

A plethora of attempts have been made to alleviate the problems of learning with a large number of unrolling steps~\citep{HS97,pascanu2013difficulty,gruslys2016memory}. Here we introduce a simple yet effective strategy for reducing the number of unrolling steps required to learn insideness. Our strategy consists of training the network with small images as they require a smaller number of unrolling steps. To test the network with bigger image sizes, we just unroll it a larger number of steps. Since the convolutional recurrent network can learn the \emph{coloring routine} (Section~\ref{SecColoring}), it is possible to generalize to larger images and more complex curves. As we have shown before, this routine does not contain long-range operations because it takes into account only $3\times 3$ neighbourhoods; the  long-range  relationships  are  captured  by  applying  the \emph{coloring  routine} multiple  times.  

Note that by using small images there is a risk that the  dataset lacks enough variability and the network does not generalize beyond  the few cases seen during training. Thus, there is a trade-off between the variability in the dataset and the number of unrolling steps required. In the following section, we explore this trade-off and show that  adjusting the image size and the number of unrolling steps leads to large gains of generalization accuracy.
 
\subsection{Experimental Setup}

\begin{figure}[t]
  \footnotesize
  \begin{tabular}{@{\hspace{+0.1cm}}c@{\hspace{-0.1cm}}c@{\hspace{0.1cm}}c}
    \multirow{2}{*}{\begin{tabular}{@{\hspace{-0.1cm}}c}
        \vspace*{-2.4cm}\\
    {\bf \footnotesize Examples of Curves of the} \\ {\bf \footnotesize Random Walk Dataset }\\[1em]
        \includegraphics[width=0.44\textwidth]{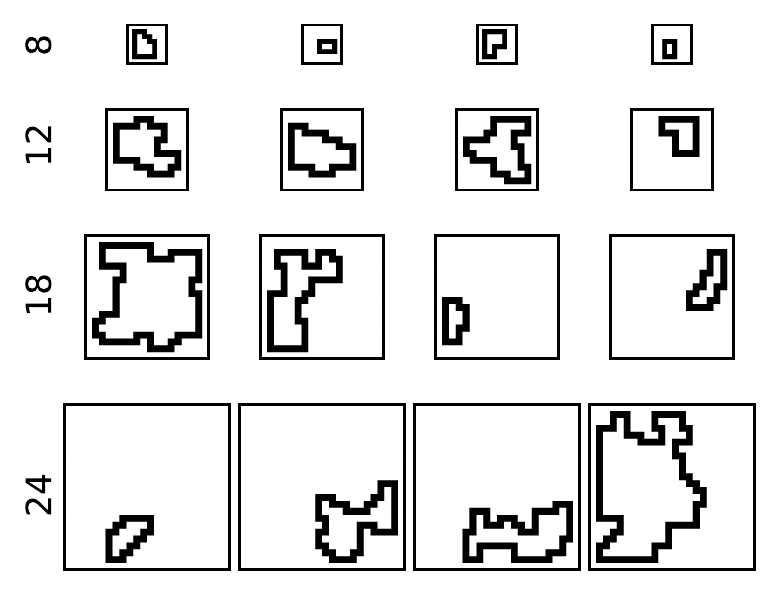}\\ 
    \footnotesize (a)
      \end{tabular}} &
    \includegraphics[width=0.27\textwidth]{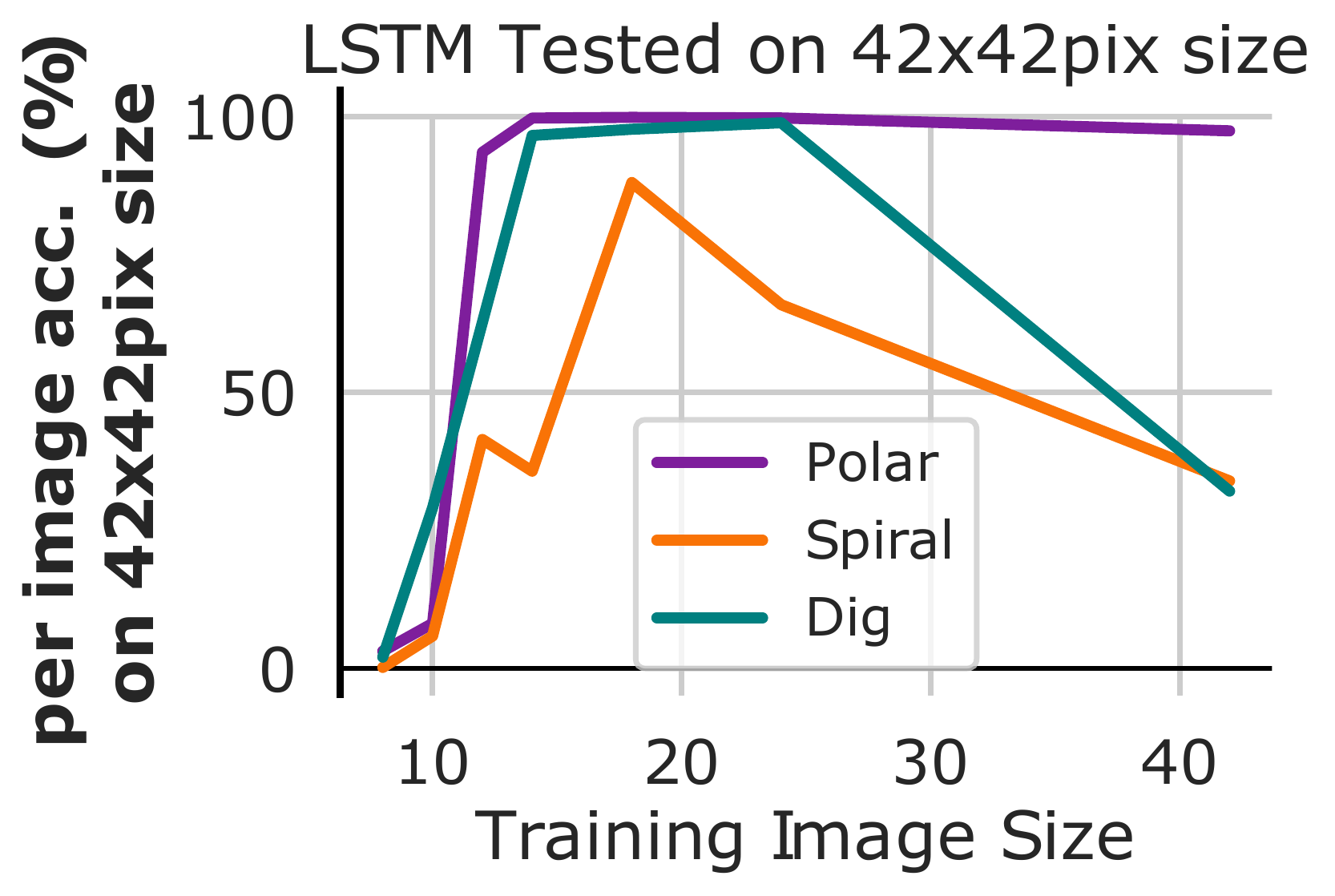}&
    \includegraphics[width=0.27\textwidth]{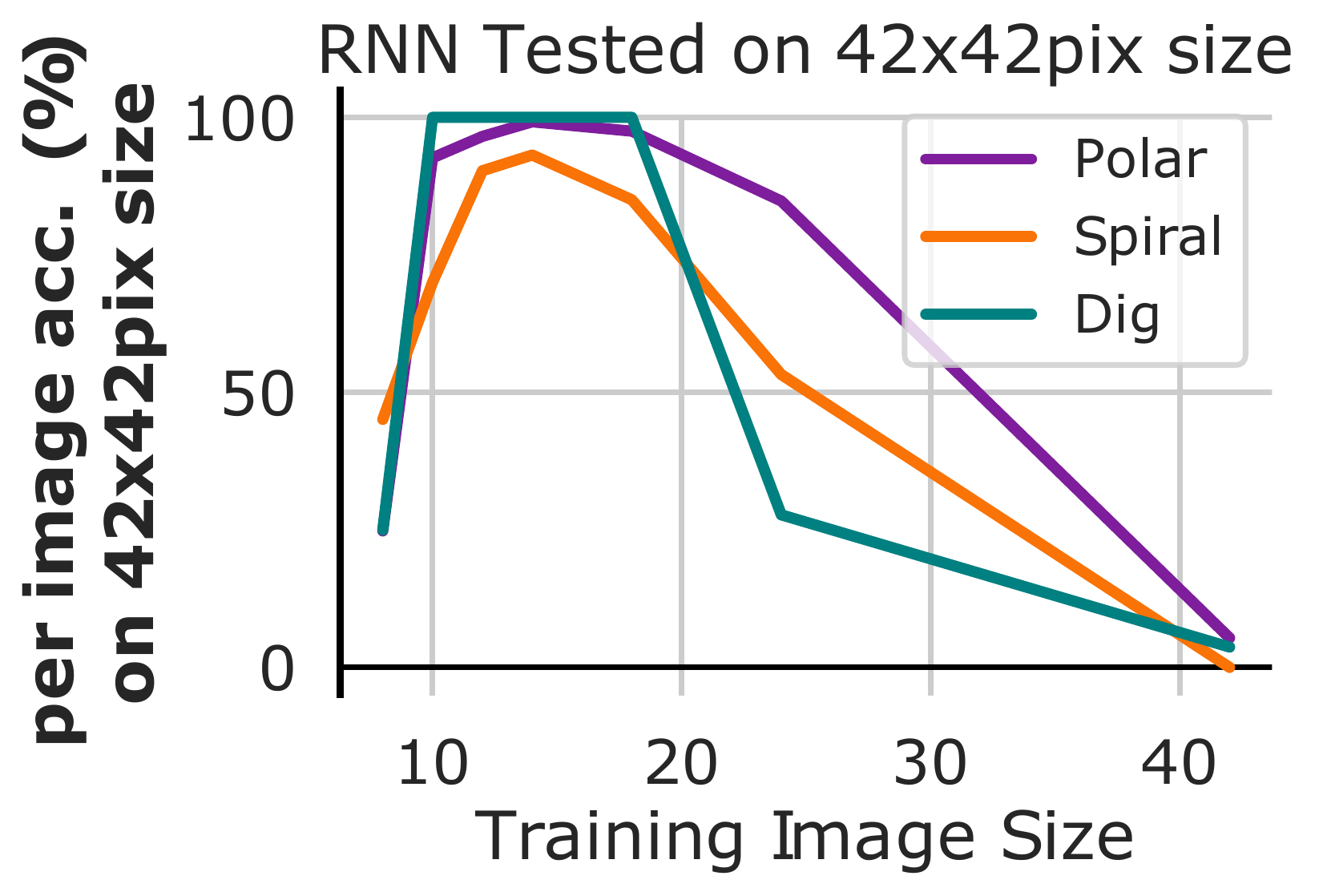}\\
 
     &
     \footnotesize (b)&
    \footnotesize (c)\\[0.3cm]
    &
    \includegraphics[width=0.27\textwidth]{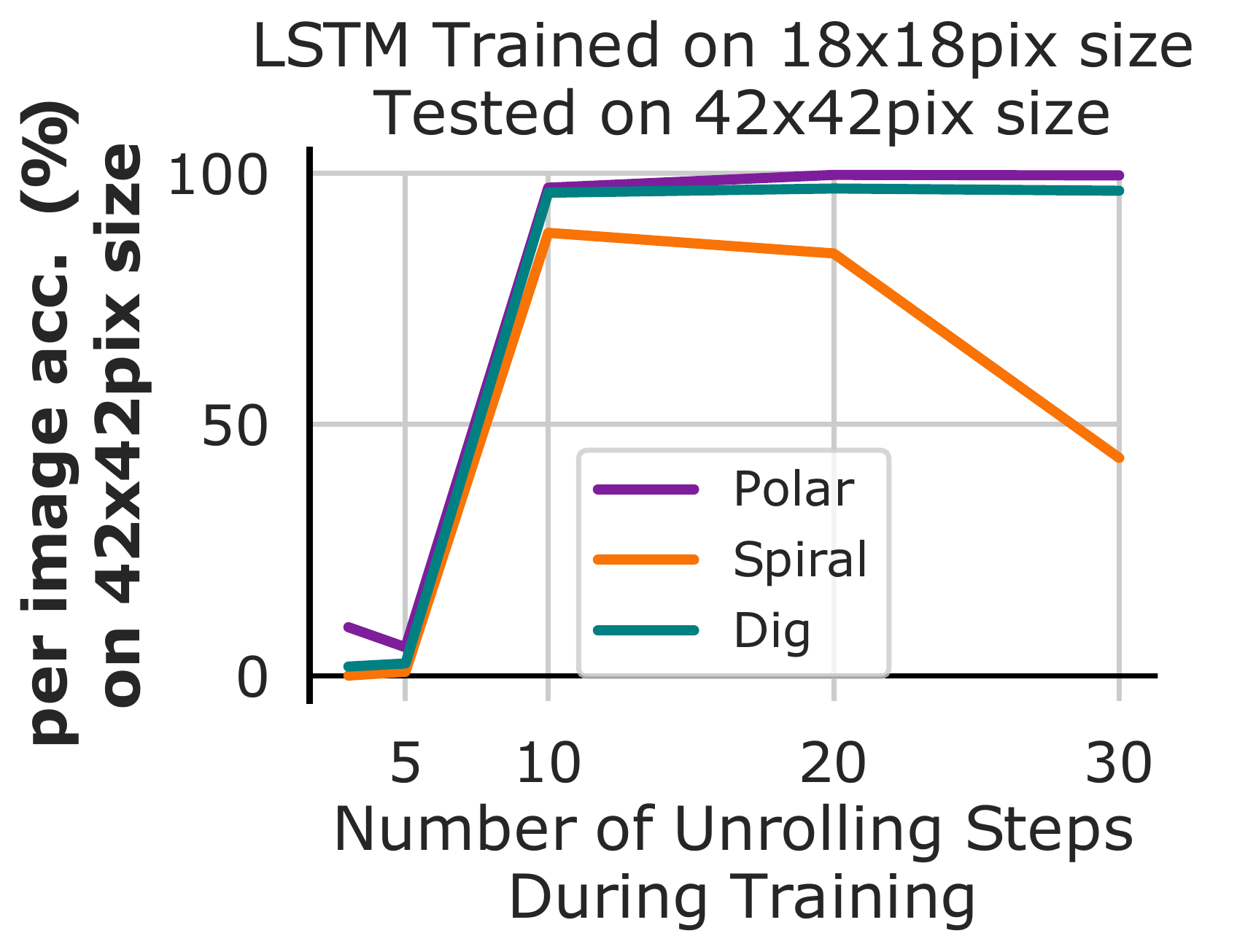}&
    \includegraphics[width=0.27\textwidth]{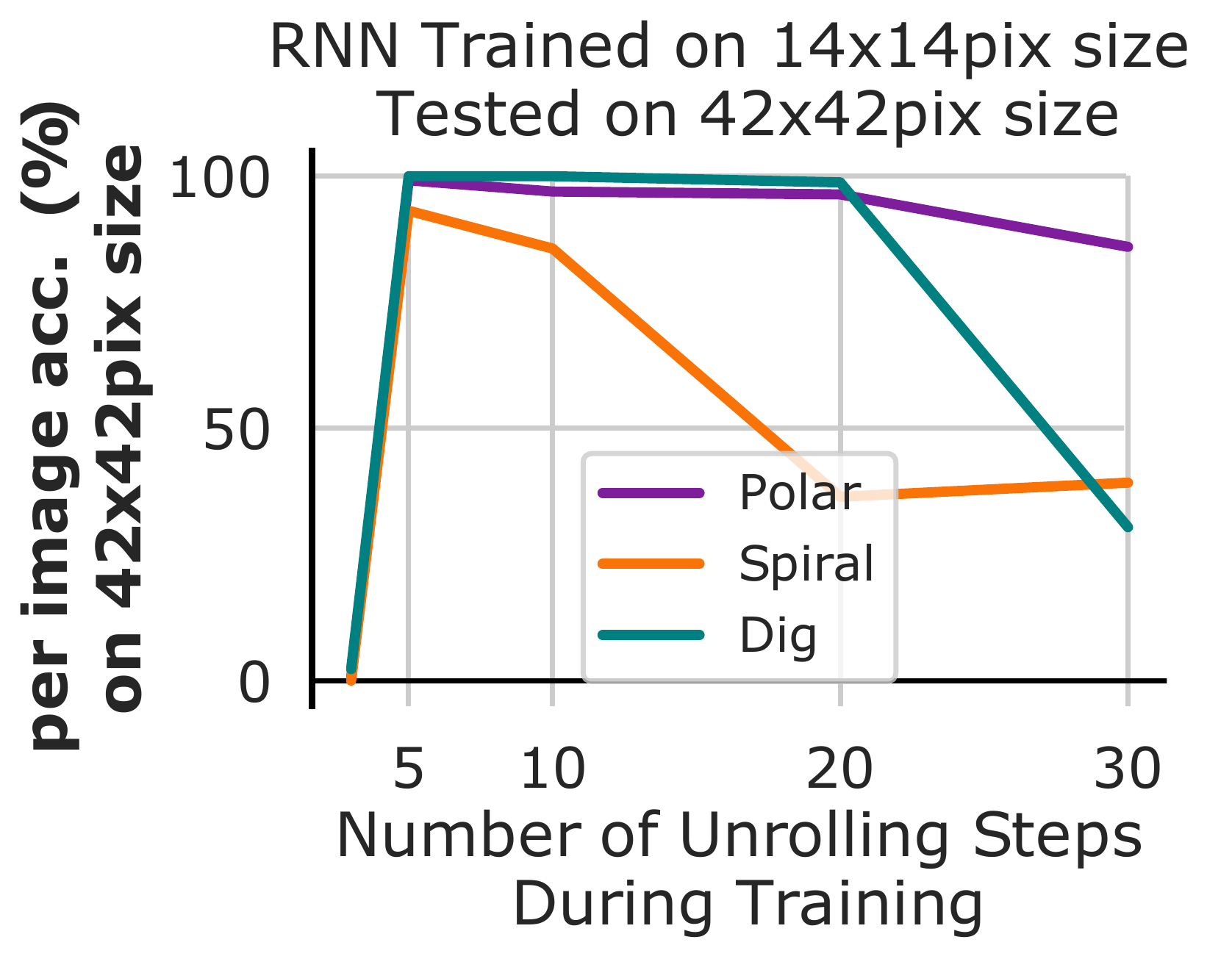}\\
     &  \footnotesize (d) &  \footnotesize (e) \\
    \end{tabular}
  \caption{\emph{Datasets and Results of Controling the Number of Unrolling Steps.} (a) Images of the curves of the Random Walk dataset used to train the recurrent networks with different image sizes. Each row correspond to a different image size.
 (b) and (c) Cross-dataset evaluation (per image accuracy) using the Polar, Spiral and Digs testing sets for ~\emph{2-LSTM} and ~\emph{RNN} networks, respectively. (d) and (e) Evaluation of generalization accuracy based on increasing numbers of unrolling steps for \textit{2-LSTM} and \textit{RNN} network, respectively.}
\label{fig:Curriculum}
\end{figure}

\noindent \textbf{The Random Walk Dataset.}
Polar, Spiral and Digs datasets are hard to adapt to small images without constraining the dataset variability. We introduce a new family of curves which we call the Random Walk dataset. 
The algorithm to generate the curves is based on a random walk following the connectivity conditions of the Jordan curve we previously introduced. It starts by selecting a random location in the image and at every iteration, choosing with equal probability any of the valid directions (not intersecting the curve constructed so far, or the border). If there are no available directions to expand in, the algorithm backtracks until there are, and continues until the starting point is reached. In Fig.~\ref{fig:Curriculum}a, we show several examples of the curves in images of different sizes. 

\noindent \textbf{Architecture.}
In order to show the generality of our results, we test the \textit{2-LSTM} and another recurrent neural network  which we denote as \textit{RNN}. We conducted an architecture search to find a recurrent network that succeeded: a convolutional recurrent neural network with a sigmoidal hidden layer and an output layer that is backwardly connected to the hidden layer. The kernel sizes are $3\times 3$ and $1\times 1$ for the hidden and output layers respectively, with $5$ kernels. Observe that this network is sufficiently complex to solve the insideness problem, because it is the network 
introduced in Appendix~\ref{secRNN} with an additional layer and connections. 

\noindent \textbf{Learning.}
We train both the \textit{2-LSTM} and the \textit{RNN} on the Random Walk dataset on $7$ different image sizes  (from $8\times 8$ to $42\times 42$ pixels) and test it with the original images of the 24-Polar, Spiral and Digs datasets ($42\times 42$ pixels).  To explore the effect of the number of unrolling steps, for each training image size we train the networks using $5$, $10$, $20$, $30$ and $60$ unrolling steps.

\renewcommand{\arraystretch}{0}

\subsection{Results}

In Figs.~\ref{fig:Curriculum}b and c, we show the cross-dataset accuracy for the \textit{2-LSTM} and \textit{RNN}, respectively, for different training image sizes. The optimal number of unrolling steps is selected for each training image size. To evaluate the cross-dataset generalization accuracy we  apply the trained models to the full size images of the  Polar, Spiral, and Digs datasets, using $60$ unrolling steps (larger number of unrolling steps does not improve the accuracy). Note that when the networks are trained with small images ($14\times 14$ is best for  \textit{RNN}, $18\times 18$ is best for \textit{2-LSTM}), at least $80\%$ of the images in all the datasets are correctly segmented. This is a massive improvement of the cross-dataset accuracy compared with the networks trained with large image sizes, as shown in previous section (Fig.~\ref{fig:General2}) and confirmed  here again with the Random Walk dataset (less than $40\%$ of the images are correctly segmented in Spiral and Digs datasets, Fig.~\ref{fig:Curriculum} b).

In Fig.~\ref{fig:Curriculum}d and e, we show the performance of \textit{2-LSTM} and \textit{RNN}, trained with their optimal training image size, when varying the number of unrolling steps. We can observe that both networks generalize  with a small number of unrolling steps  and fail to generalize as the number of unrolling steps is increased during training. Also, note that as expected, \textit{2-LSTM} is more robust than the \textit{RNN} to large numbers of unrolling steps during training.
These positive results demonstrate that training with smaller images and number of unrolling steps is extremely useful in enabling different types of recurrent networks to learn a general solution of insideness across different families of curves not seen during training. 

Furthermore, we observe that the solution implemented by the network is the~\emph{Coloring Routine} (Fig.~\ref{visSuppLSTM}). In Fig.~\ref{figMAthLSTM} and Fig.~\ref{visSuppLSTM_shaping} in Appendix~\ref{Additional_Figures_and_Visualizations}, we  visualize the internal feature maps and observe that the network implements the~\emph{Coloring Routine} in a different way from the analytically-derived solution, as the internal features maps are different. This suggests that there may be other ways to implement the~\emph{Coloring Routine} in a \emph{2-LSTM} than the analytical solution we introduced. 

Note that the~\emph{Coloring Routine} emerges in the network and we do not enforce it in any way besides training the network with small images and small number of unrolling steps. Note that we could enforce  the network to learn the~\emph{Coloring Routine} by providing the ground-truth produced by the routine at each step rather than waiting until the last step. In Appendix~\ref{sec:perstep} we show that this per-step strategy also leads to successful cross-dataset generalization accuracy.  This result is however less interesting  as it requires the per-step ground-truth derived from the analytical solution.

%


%








\section{Conclusions}

We have shown that state-of-the-art DNNs for segmentation based on dilated convolutions and convolutional LSTM networks are sufficiently complex, with a number of units implementable in practice, to solve the insideness problem. Yet, when using the standard training strategies, the units in these networks become specialized to detect characteristics of the curves in the training set and only generalize to curves of the same family as the training. We introduced a strategy to alleviate this limitation based on convolutional recurrent networks that facilitates the emergence of an implementation of the \emph{coloring routine}. The \emph{coloring routine} breaks the evaluation of long-range relationships into local operations and when trained with small images, it generalizes substantially better to new families of curves as it alleviates the well-known difficulties of training recurrent networks with a large number of unrolling steps. 

While our results on learning general solutions of insideness are promising, the \emph{Coloring Routine} may require a large number of unrolling steps to segment complex curves, as it expands only one pixel in each unrolling step. More research is needed to investigate mechanisms that facilitate reducing the number of unrolling steps both at training and testing phases. A potential source of inspiration to
ameliorate this issue can be found in constraints and mechanisms found in  brains.  \cite{jeurissen2016serial} provided evidence to support that the brain's algorithm for insideness can be described with the growth-cone model. This is a more sophisticated version of the coloring
method where coloring spreads fast through large homogeneous regions and slows down in narrow, difficult parts of figures. Such an algorithm could be implemented for segmentation in brains by units processing larger receptive fields being able to propagate information to those with small receptive fields and vice-versa. Analogously, by building
networks following the visual hierarchy, recurrent filling-in additions can be added at different layers of a deeper feed-forward network to process features at different scales. This could lead to the emergence of the growth-cone model, which would help to reduce the necessary number of unrolling steps and improve generalization without the need of training in small images.

Finally, we hope that the insights introduced in this paper inspire new segmentation training procedures and help future investigations of other important aspects of segmentation beyond insideness (\eg the discontinuity of segments and the hierarchical structure of segments).




\vspace{\baselineskip}
\noindent {\bf Acknowledgements. } We are grateful to Tomaso Poggio for his insightful advice and warm encouragement. We also thank Shimon Ullman and Pawan Sinha for helpful feedback and discussions. 
This work is supported by the Center for Brains, Minds and Machines (funded by NSF STC award CCF-1231216),  Fujitsu Laboratories Ltd. (Contract No. 40008401) and the MIT-Sensetime Alliance on Artificial Intelligence.

\vspace{\baselineskip}
\noindent {\bf Code and Data Sets. } They are available at \url{https://github.com/biolins/insideness }

\clearpage

\appendix
\renewcommand\thefigure{\thesection.\arabic{figure}}    
\setcounter{figure}{0}  

\section{Number of Digital Jordan Curves}
\label{secSuppJordan}

\begin{figure}[t]
\footnotesize
	\centering
  \begin{tabular}{cccccc}
      {} &
      {} &
      {} & 
	\includegraphics[width=0.125\linewidth]{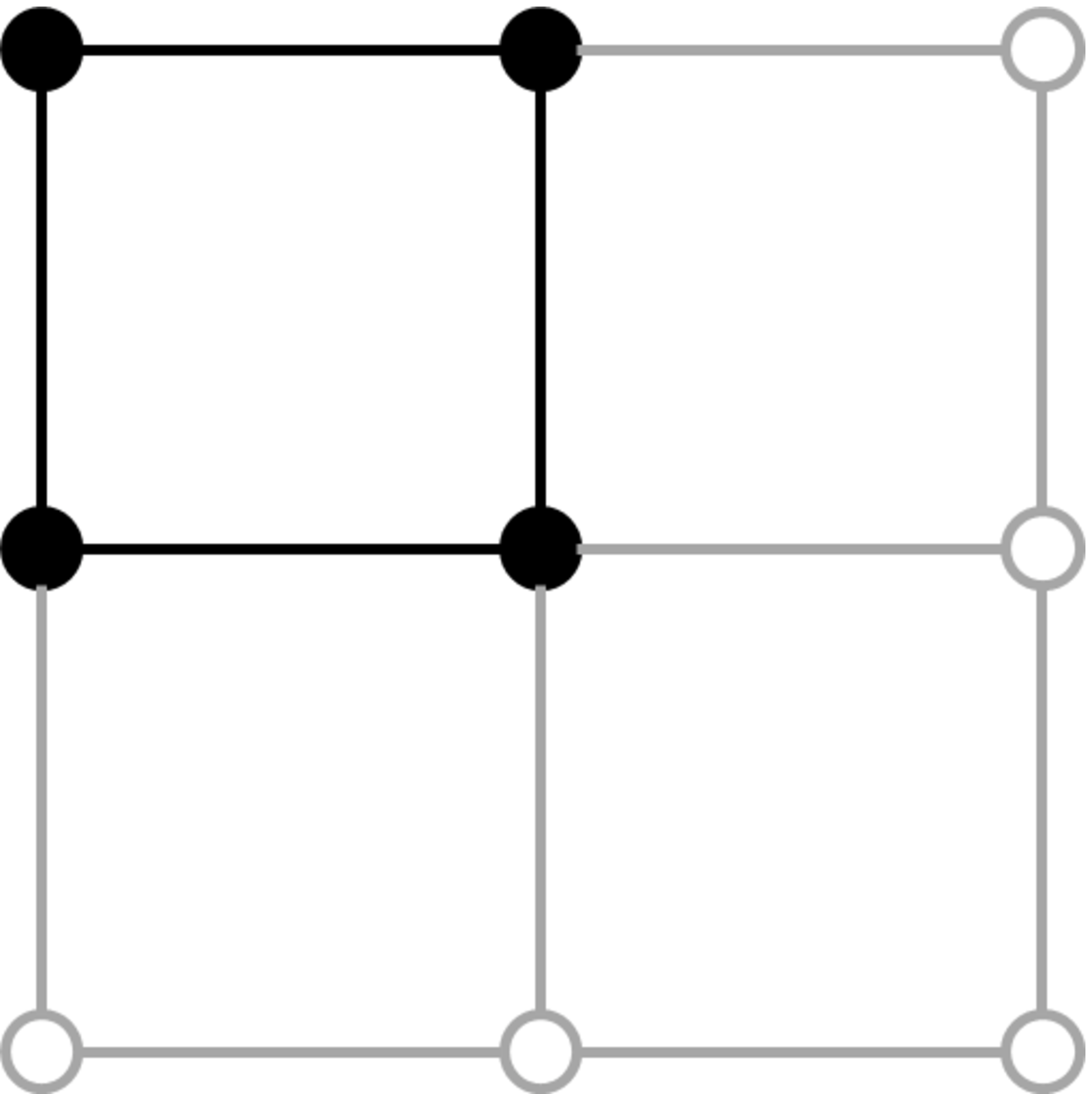}&
	\includegraphics[width=0.125\linewidth]{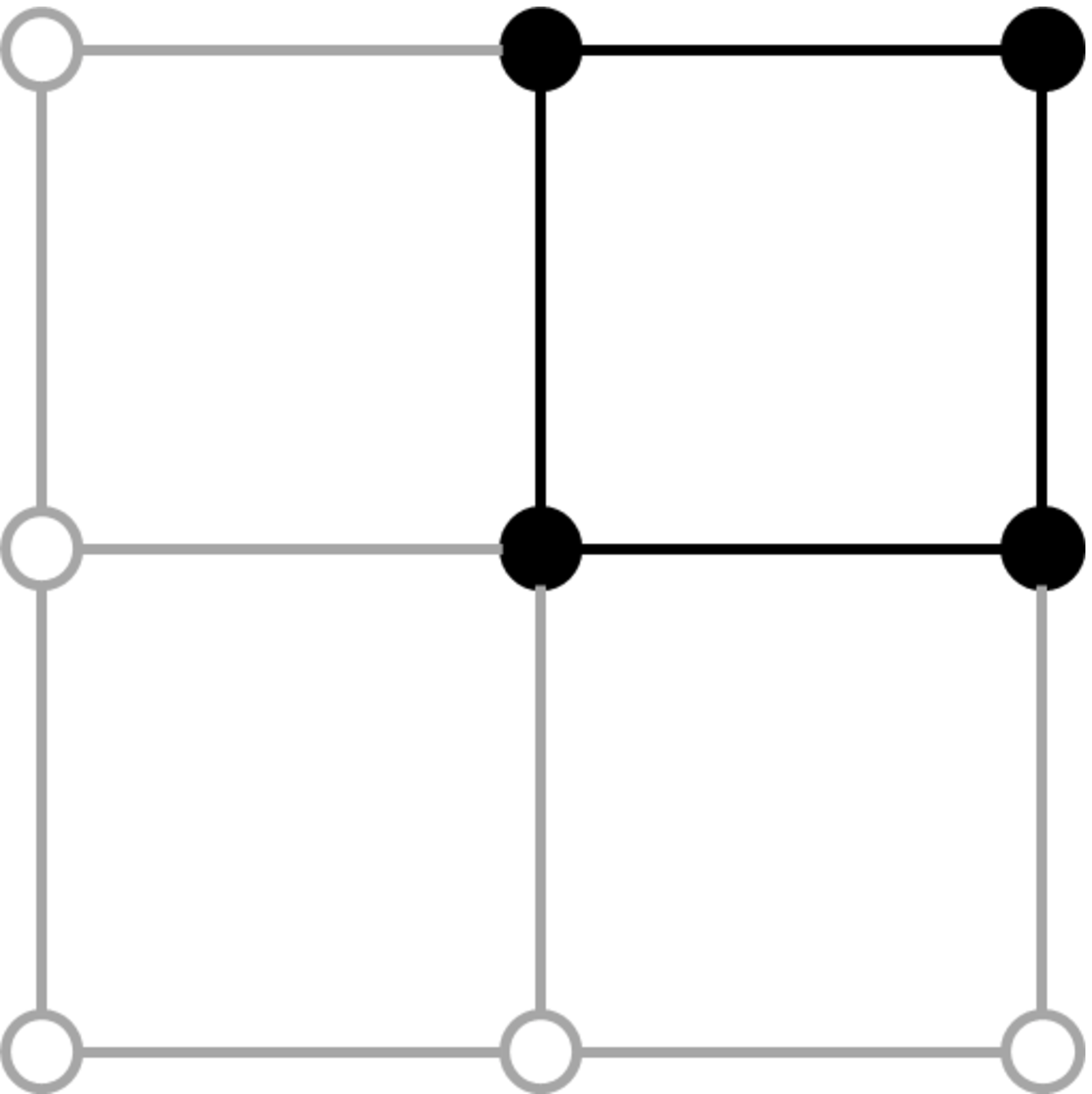}&
	\includegraphics[width=0.125\linewidth]{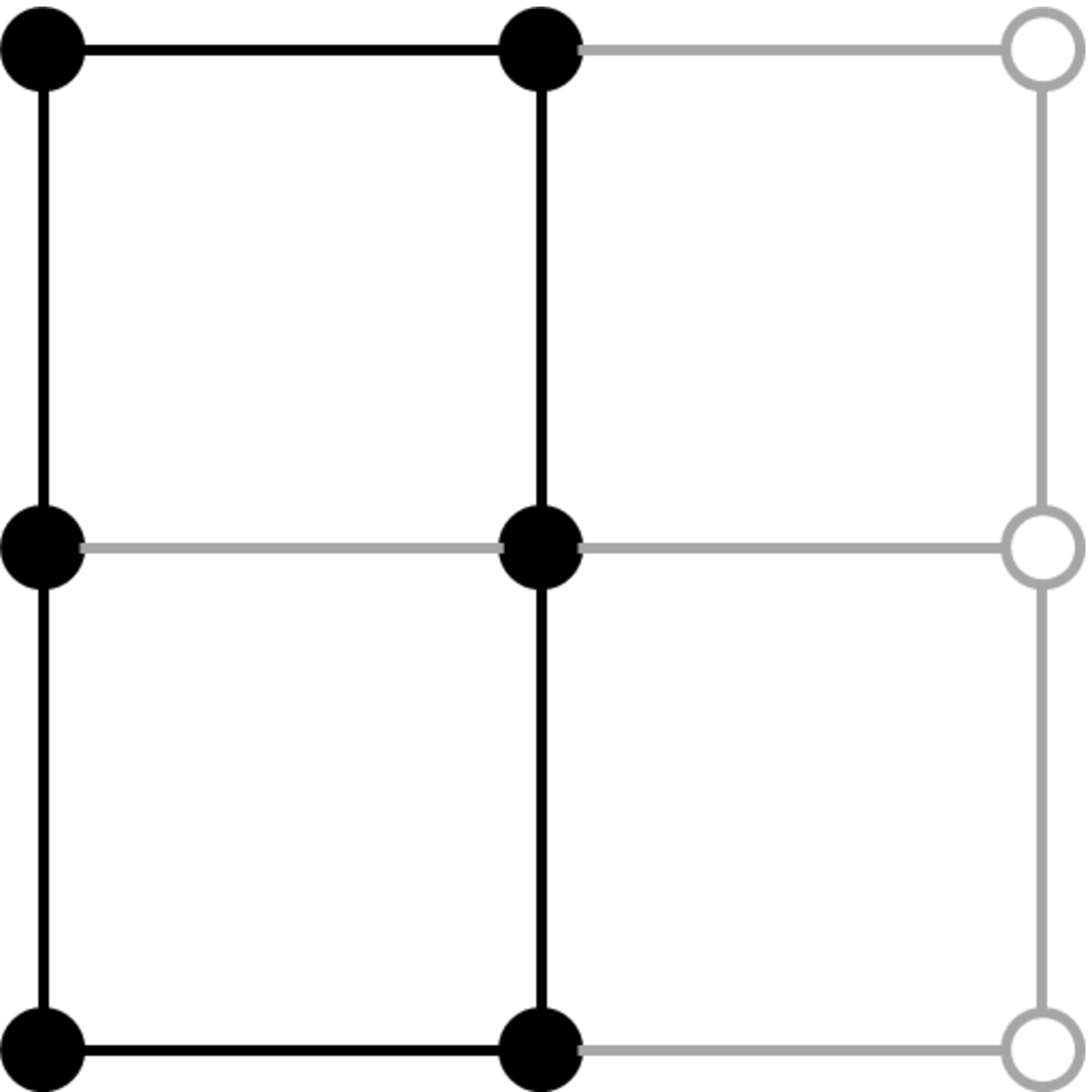} \\
        {} & 
        {} & 
        {} &
        {} & 
        {} & 
        {} \\
      \includegraphics[width=0.15\linewidth]{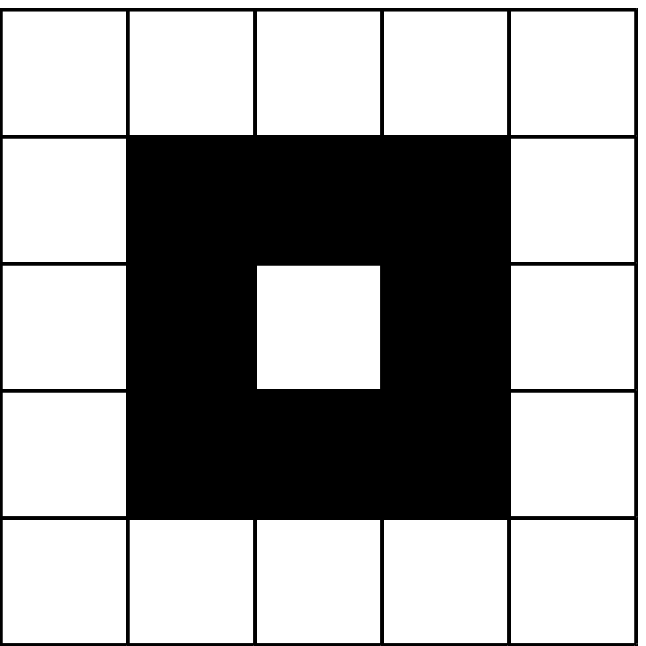} &
      \raisebox{0.06\linewidth}{\scalebox{1.5}{$\Leftrightarrow$}}  &
      \includegraphics[width=0.15\linewidth]{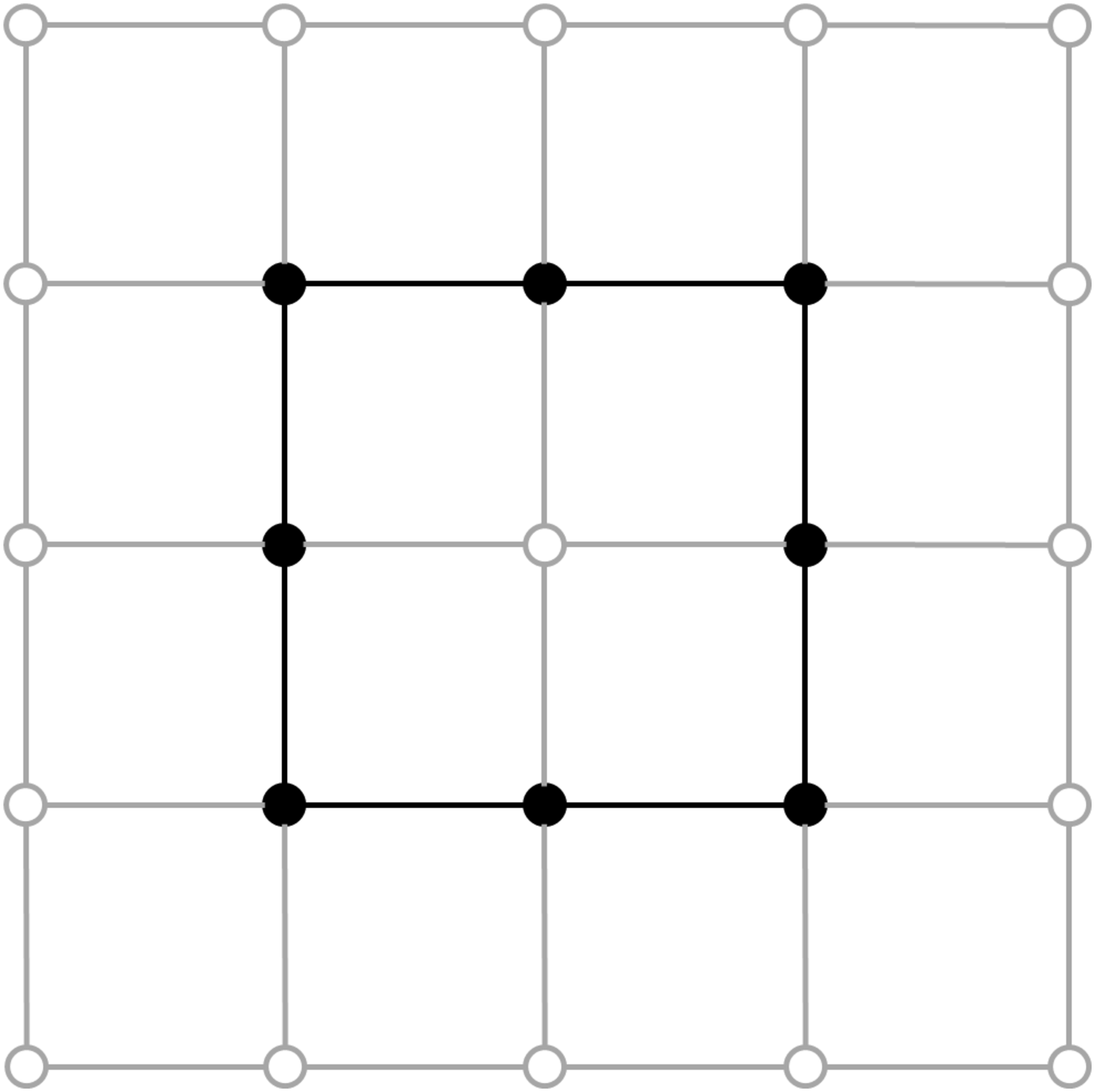} &
 	\includegraphics[width=0.125\linewidth]{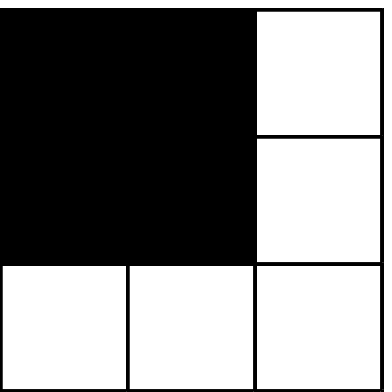}&
	\includegraphics[width=0.125\linewidth]{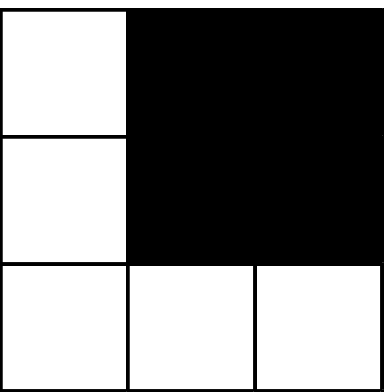}&
	\includegraphics[width=0.125\linewidth]{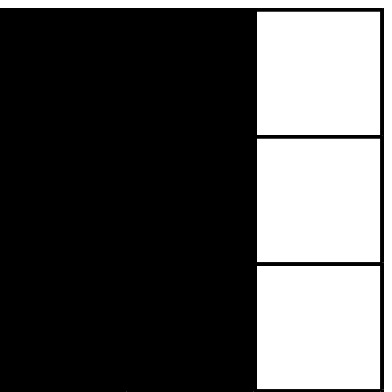} \\
    [+0.5cm] 
        & 
        (a) & 
        &
        & 
        (b) & 
        {} 
  \end{tabular}
\caption{ \emph{Subgraph Representations of Figures.} (a) A figure in an image of size 5 $\times$ 5 pixels (left) and its subgraph representation in a grid graph of  5 $\times$ 5  vertices (right). (b) Cycles that are not digital Jordan curves (top) and their correspondents (bottom).}
\label{NotJordan}
\end{figure}

Here we introduce a procedure to derive a lower bound of the number of Jordan curves in an image.
We represent an image of size $N \times N$ pixels by using a grid graph  (square lattice)  with $N \times N$ vertices.  
We employ $4$-adjacency for black pixels and corresponding grid points, and $8$-adjacency for white pixels and their counterpart.
Then, a curve (the set of black pixels) corresponds to a subgraph of the base grid graph (Fig.~\ref{NotJordan}a).

In this representation, 
a digital Jordan curve 
is defined as a subgraph specified by a sequence of vertices 
$(\mathtt{v}_{0}, \mathtt{v}_{1}, \dots, \mathtt{v}_{L})$ 
satisfying the following conditions~\citep{Rosenfeld70,Kong01}:  
\begin{enumerate}
\item \label{adjacency-condition} $\mathtt{v}_{r}$ is 
$4$-adjacent to 
$\mathtt{v}_{s}$ if and only if  $r  \equiv s \pm 1 \pmod{L + 1}$, 
\item \label{closedness-condition} $\mathtt{v}_{r} = \mathtt{v}_{s}$ if and only if  $r = s$, and 
\item \label{length-condition} $L \ge 4$.
\end{enumerate}
The ``if'' part of condition \ref{adjacency-condition} means that the black pixels lie consecutively and the curve formed by the black pixels is closed. 
The ``only if'' part of conditions \ref{adjacency-condition} and \ref{closedness-condition} assures that the curve never crosses or touches itself. 
The condition \ref{length-condition} excludes exceptional cases which satisfy conditions \ref{closedness-condition} and \ref{length-condition} but cannot be considered as digital versions of the Jordan curve (see \citet{Rosenfeld70} for details).
Note that any digital Jordan curve is a cycle~\citep{harary1969graph} in a grid graph but not vice versa.  
Figure \ref{NotJordan}b shows examples of cycles that are not digital  Jordan curves.



The numbers of all cycles in  grid graphs of different sizes were computed up to $27 \times 27$ vertices~\citep{Iwashita+13b,A140517}, 
and we utilize this result to get lower bounds for the number of digital Jordan curves with the following considerations.

\begin{figure}[t]
\footnotesize
	\centering
  \begin{tabular}{cccccc}
	\includegraphics[width=0.15\linewidth]{./fig/GridPoint3x3-NotJordan-1.eps}&
	\raisebox{0.075\linewidth}{\scalebox{1.5}{$\mapsto$}} &
	\includegraphics[width=0.15\linewidth]{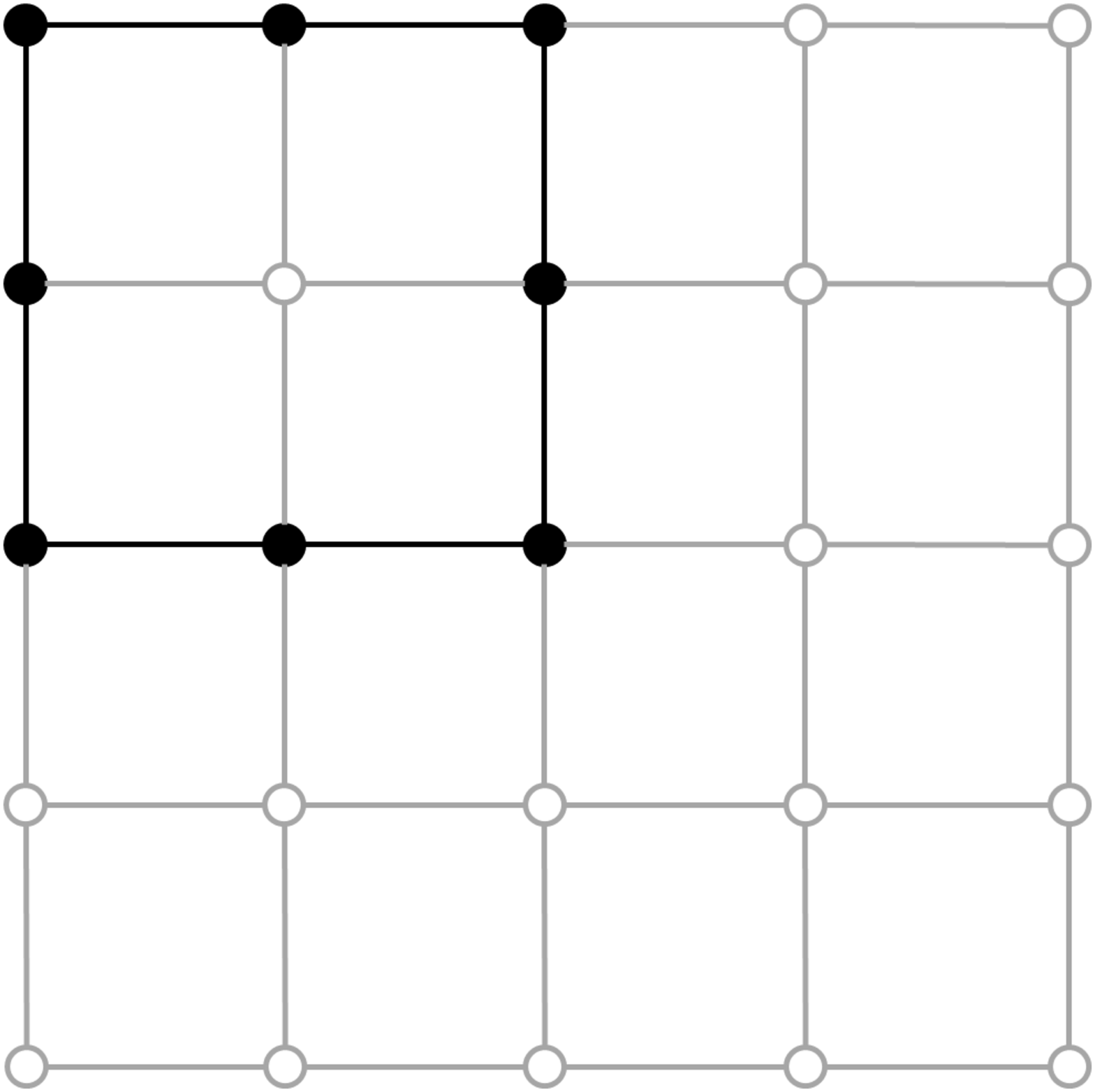} & 
      {} & 
      \includegraphics[width=0.15\linewidth]{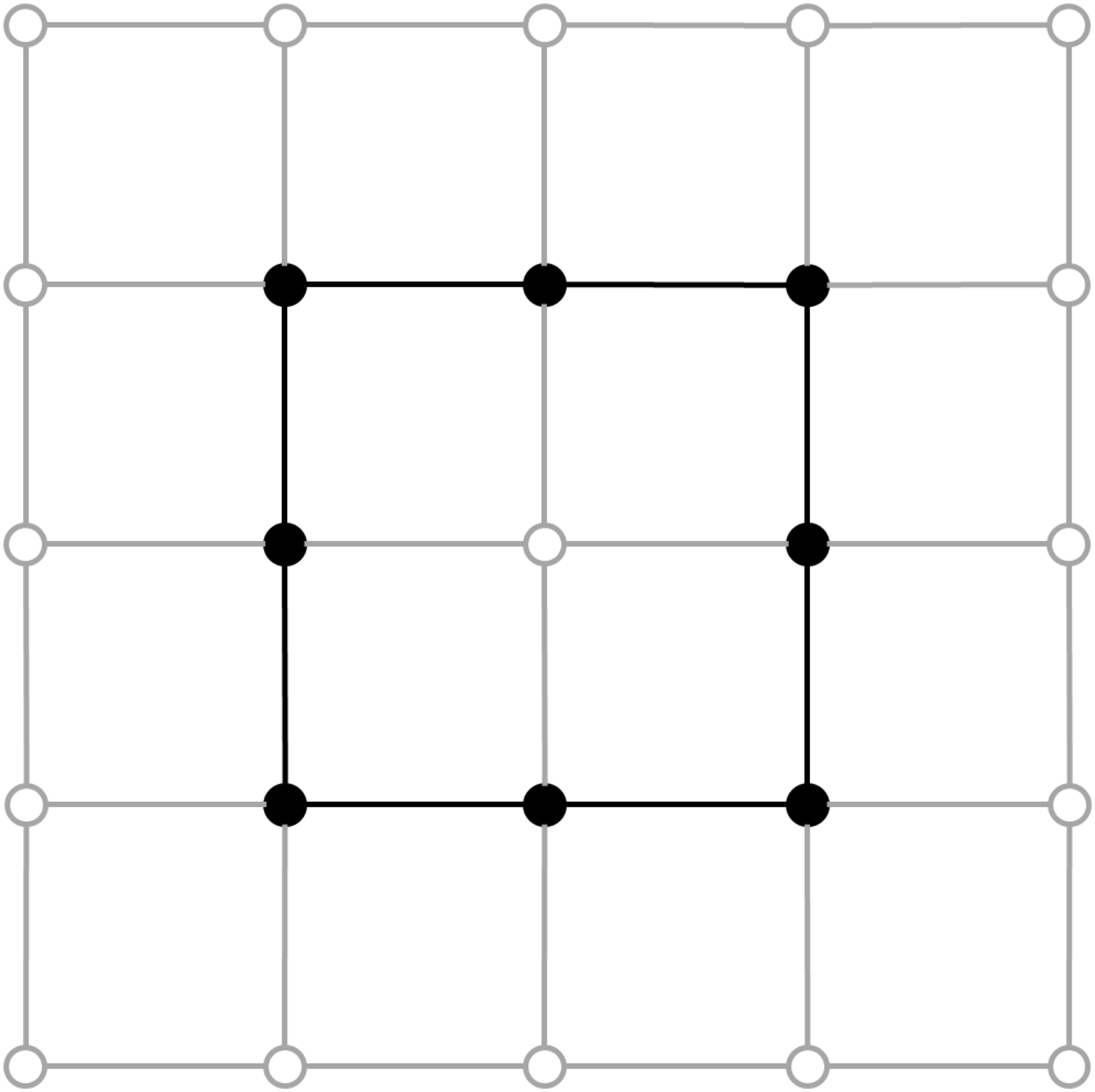} & 
      \includegraphics[width=0.15\linewidth]{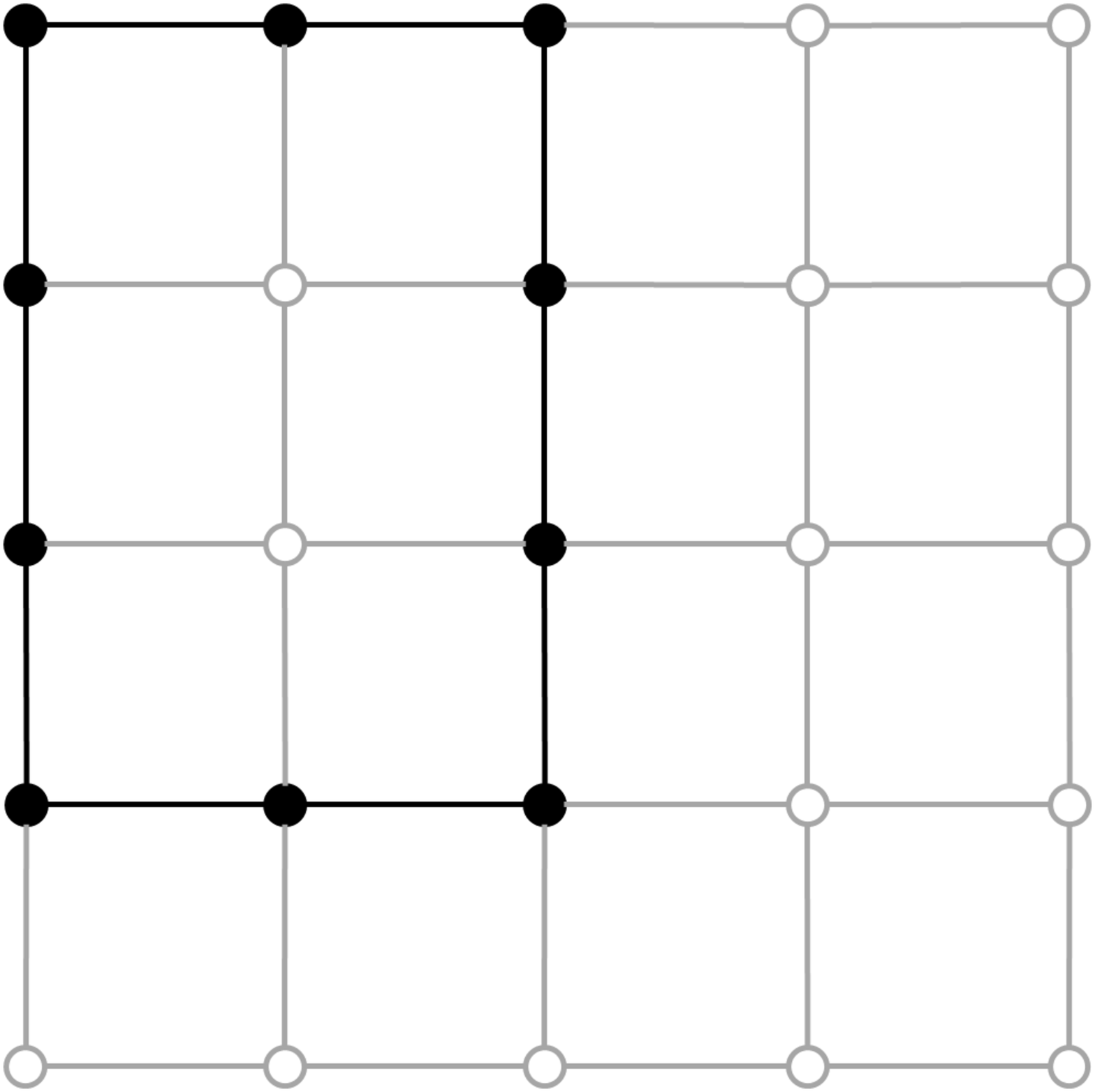}  \\
	[+0.5cm] 
	\includegraphics[width=0.15\linewidth]{./fig/Pixel3x3-NotJordan-1.eps}&
	\raisebox{0.075\linewidth}{\scalebox{1.5}{$\mapsto$}} &
	\includegraphics[width=0.15\linewidth]{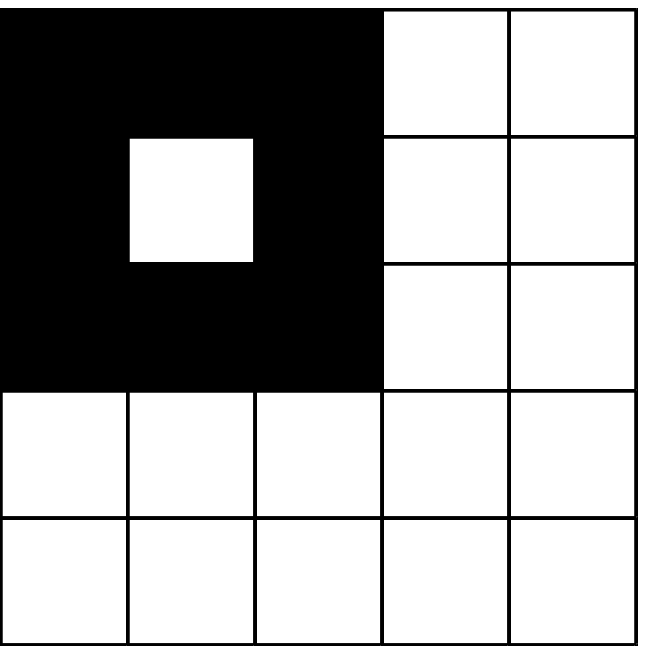} & 
      {} &
      \includegraphics[width=0.15\linewidth]{./fig/Pixel5x5.eps} & 
      \includegraphics[width=0.15\linewidth]{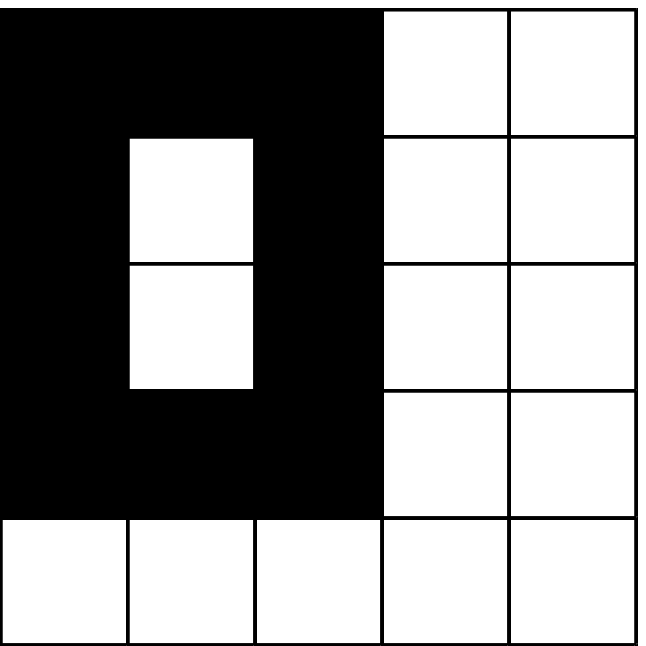}  \\
	[+0.5cm]
      {} & 
      (a) & 
      {} &  
      \multicolumn{3}{c}{(b)}
  \end{tabular}
\caption{ \emph{``Upsampling" Operation and Its Limitations.} (a) Depiction of ``upsampling" operation. (b) Digital Jordan curves that cannot be obtained by the upsampling shown in (a). (Left)  The issue is the place of the  digital Jordan curve. We can get the same curve on the upper-left, upper-right, lower-left and lower-right corners but cannot get the one in the center. (Right) The issue is the length of the side. We cannot get a side with 4 vertices (4 pixels) nor with even number vertices (pixels) in general. }
\label{Upsampling}
\end{figure}

Although a cycle in a grid graph is not necessarily a digital Jordan curve as shown above,   
we can obtain a digital Jordan curve in a larger image from any cycle by  ``upsampling" as shown in Fig.~\ref{Upsampling}a. 
Note that there are other digital  Jordan curves than the ones obtained in this manner (therefore we get a lower bound with this technique). 
See Fig.~\ref{Upsampling}b for examples. 

\begin{figure}[t]
\footnotesize
	\centering
  \begin{tabular}{ccc}
	\includegraphics[height=0.2\linewidth]{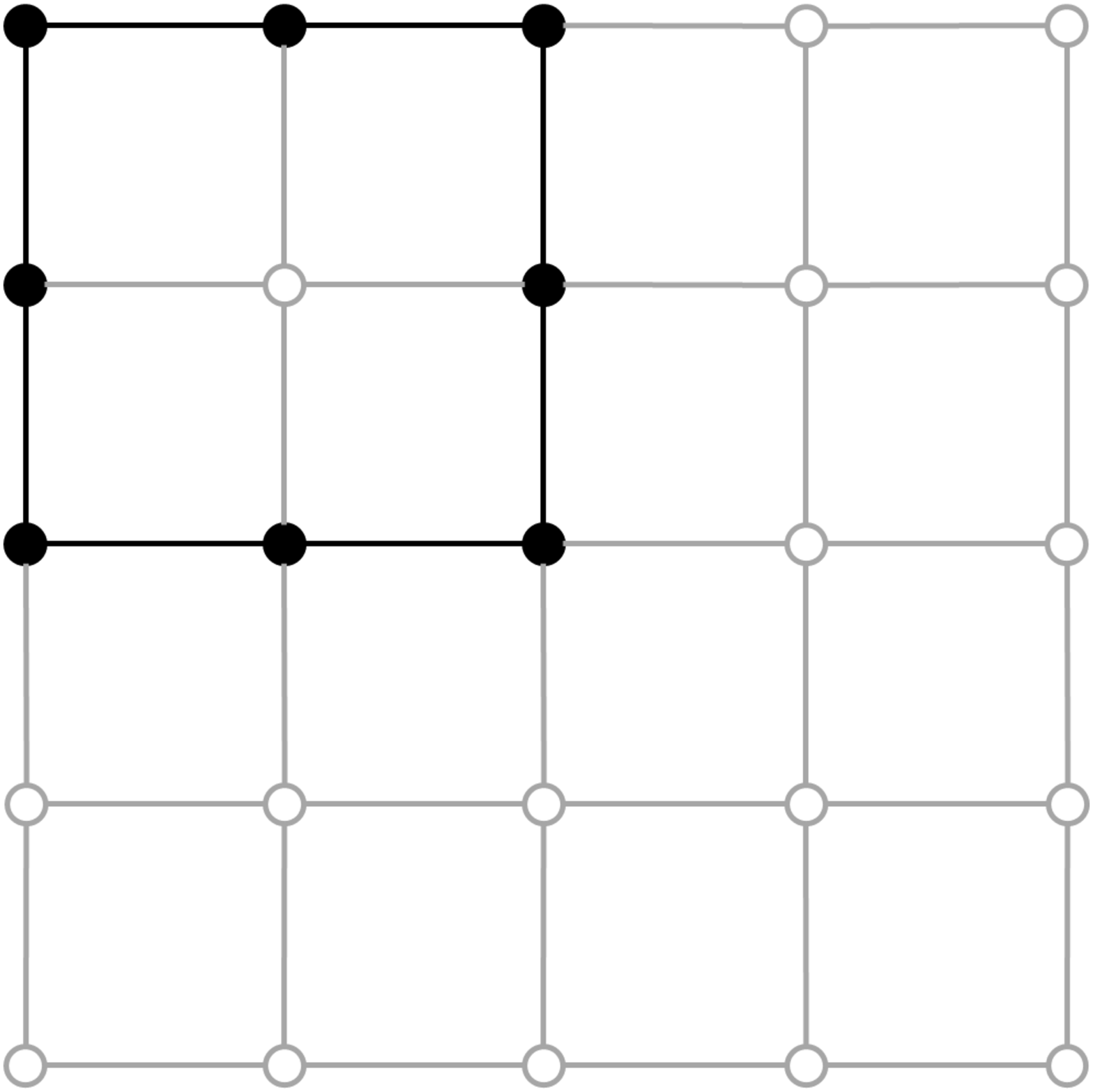}&
	\raisebox{0.085\linewidth}{\scalebox{1.5}{$\mapsto$}} &
	\includegraphics[height=0.2\linewidth]{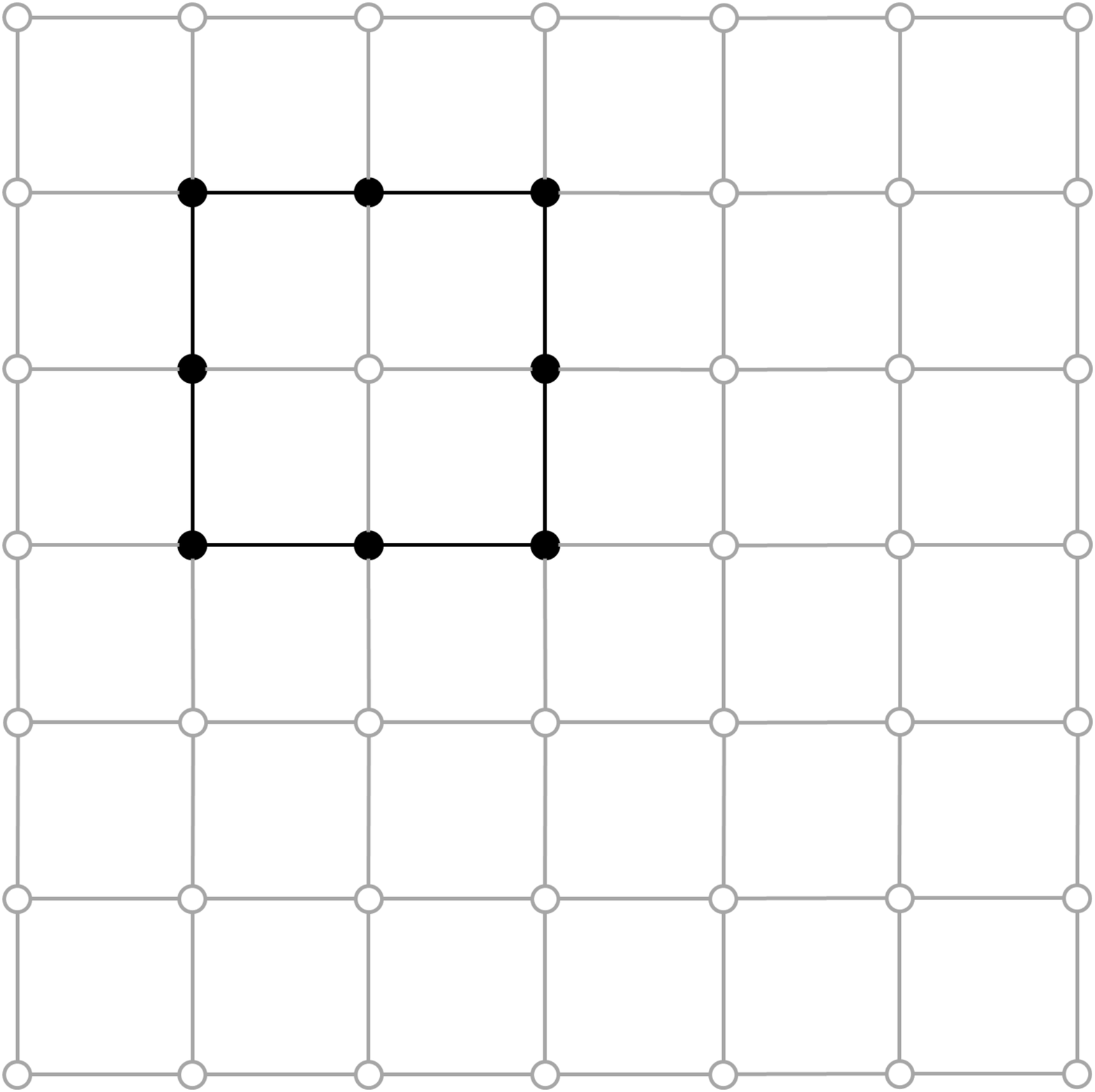} 
  \end{tabular}
\caption{\emph{Depiction of ``Padding"}.}
\label{Padding}
\end{figure}

We also consider ``padding" shown in Fig.~\ref{Padding} to assure that a digital Jordan curve does not contain the border of a image (this is what we assume in the main body of the paper). 

Taking everything into consideration, 
we can obtain a lower bound of the number of digital Jordan curves in an $N \times N$ image that does not contain border pixels 
utilizing the above-mentioned result~\citep{Iwashita+13b,A140517}, upsampling and padding. Table~\ref{TableJordan} shows lower bounds obtained in this way.
For example, starting with the row 2 of~\citet{A140517-Table} in~\citet{A140517} (this represents the number of all cycles in the grid graph with $3 \times 3$ vertices),  we get a lower bound 13 for the number of digital Jordan curves in $5 \times 5$ images by considering the upsampling and get the same number as a lower bound  for the number of digital Jordan curves that do not contain border pixels in $7 \times 7$ images by considering the padding.



\renewcommand{\arraystretch}{1.4}
\begin{table*}[t]
\begin{center}
\caption{\emph{Lower bounds (LBs) of the number of digital Jordan curves in $N \times N$ images that do not contain border pixels}.}
\small 
\begin{tabular}{|c|c|c|c|c|c|c|c|c|c|c|}
\hline
$N$  & 5 & 7  & 9 & $\cdots$ & 31  & 33  & 35 & $\cdots$ & 55 \\
\hline\hline
LB &  1  &  13 & 213 & $\cdots$ &  $1.203 \times 10^{47}$ & $1.157 \times 10^{54}$   &  $ 3.395 \times 10^{61}$  & $\cdots$ & $6.71 \times 10^{162}$ \\
\hline
\end{tabular}
\label{TableJordan}
\end{center}
\end{table*}
\renewcommand{\arraystretch}{1.0}

\section{Implementing the Ray-intersection algorithm with Dilated Convolutional Networks }
\label{secAppendixDilation}

\label{secDilation}

\begin{figure}[t]
    \centering
    \footnotesize
    \begin{tabular}{c@{\hspace{0.3cm}}c@{\hspace{0.3cm}}c}
    \includegraphics[width=0.14\linewidth]{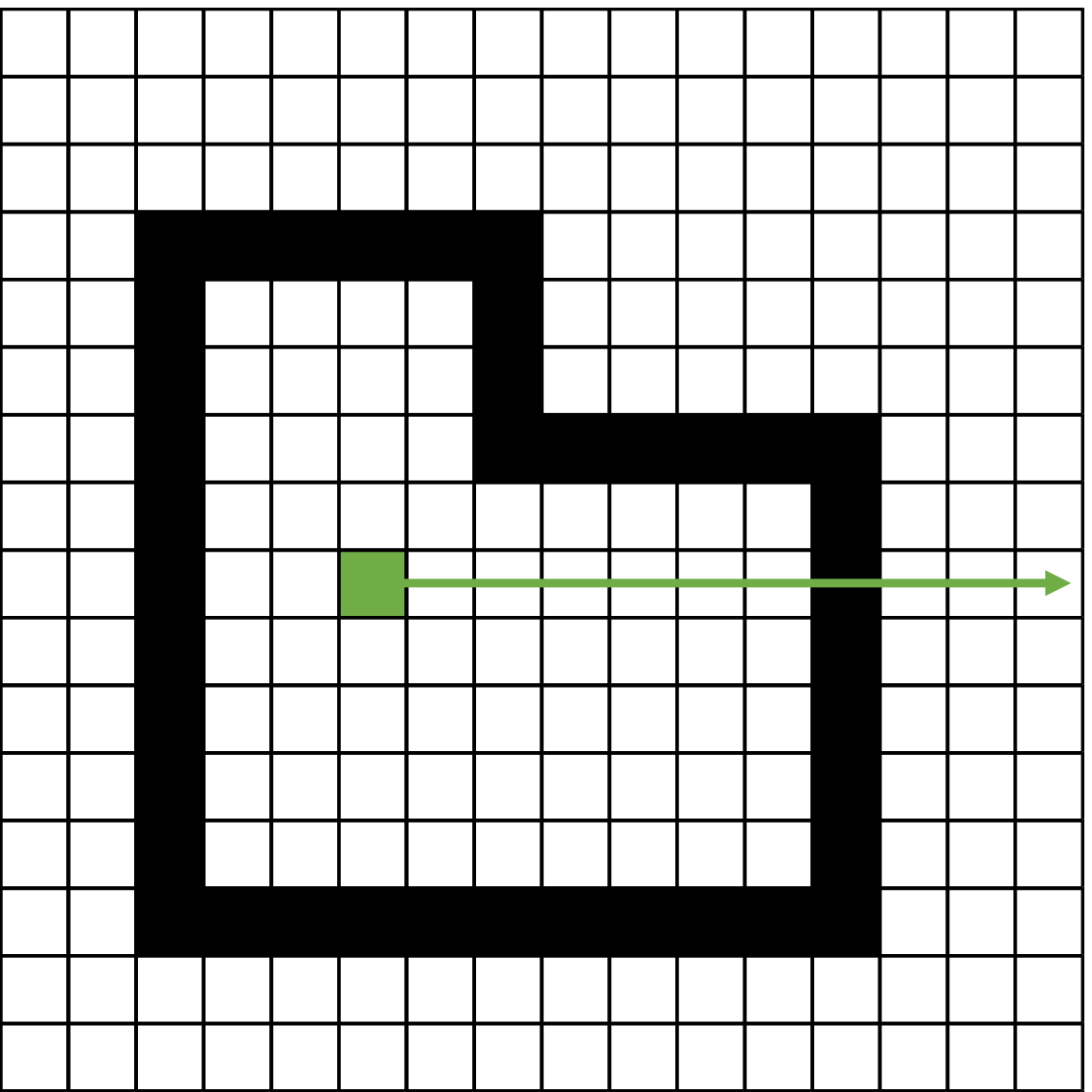} &
    \includegraphics[width=0.14\linewidth]{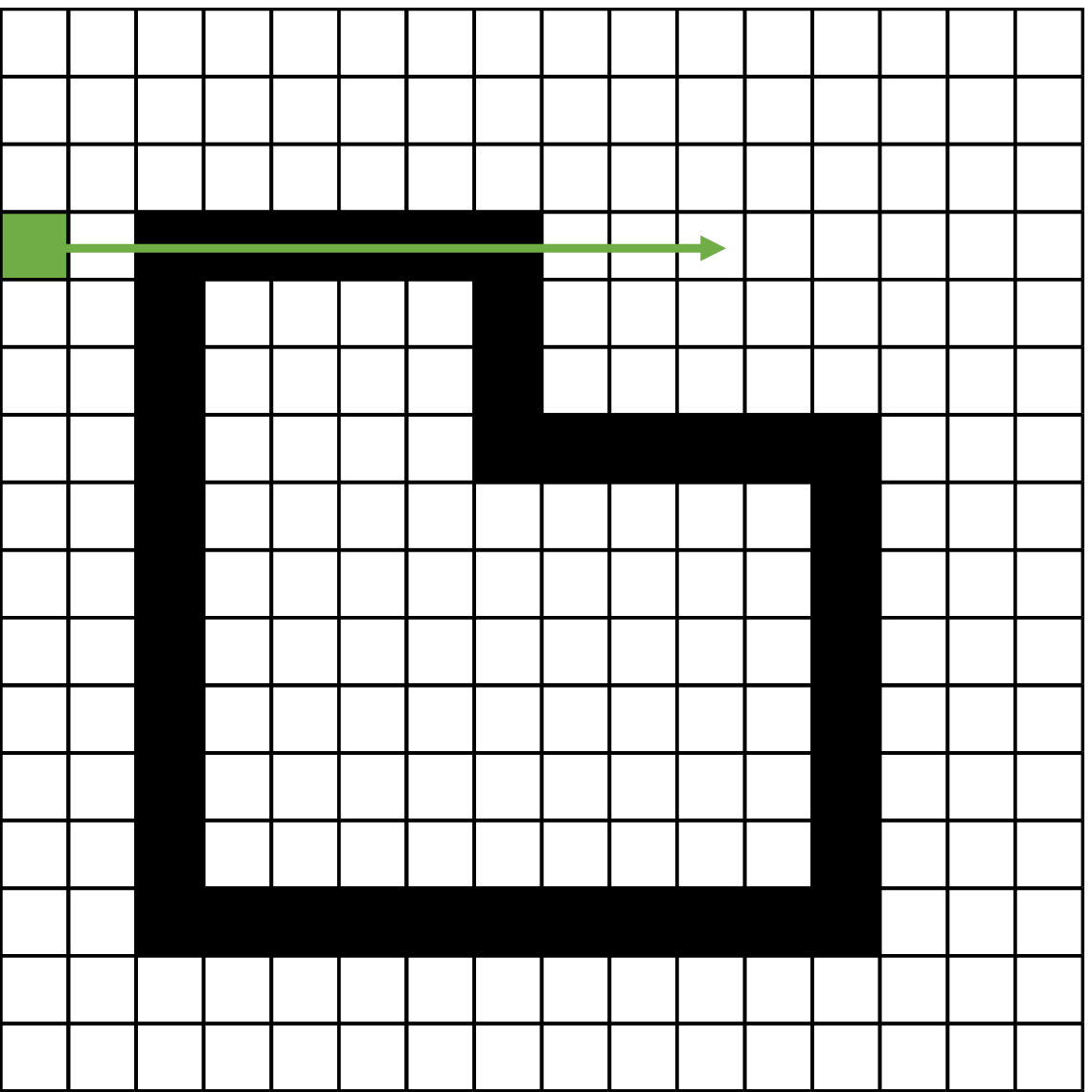} &
            \includegraphics[width=0.62\linewidth]{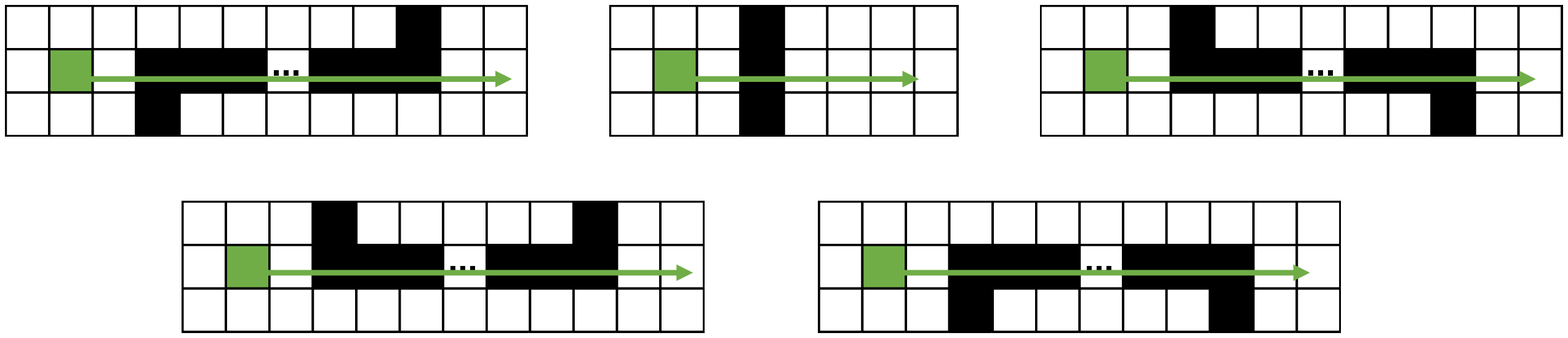}
\\
    (a) & (b) & (c) \\
    \end{tabular}
    \caption{\emph{Intersections of the Ray and the Curve.} (a)~Example of ray going from one region to the opposite one when crossing the curve. (b)~Example of ray staying in the same region after intersecting the curve. (c)~All cases in which a ray could intersect a curve. In the three cases above the ray travels from one region to the opposite one, while in the two cases below the ray does not change regions.}%
    \label{fig:fig/RayIntersectionMethod.eps}
\end{figure}

Dilated convolutions facilitate capturing long-range dependencies which are key for segmentation~\citep{YuKoltun2016,chen2018deeplab}. To demonstrate that there are architectures with dilated convolutions that can solve the insideness problem, we borrow insights from the ray-intersection method.
The ray-intersection method~\citep{Ullman84,Ullman96}, 
also known as the crossings test or the even-odd test~\citep{Haines94},  
is built on the following fact: Any ray that goes from a pixel to the border of the image alternates between inside and outside regions every time it crosses the curve. Therefore, the parity of the total number of such crossings determines the region to which the
pixel belongs. If the parity is odd then the pixel is inside, otherwise it is outside (see Fig.~\ref{fig:fig/RayIntersectionMethod.eps}a).

The definition of a crossing should take into account cases like the one depicted in Fig.~\ref{fig:fig/RayIntersectionMethod.eps}b, in which the ray intersects the curve, but does not change region after the intersection.  
To address these cases, we enumerate all possible intersections of a ray and a curve, and analyze which cases should count as crossings and which ones should not. Without loss of generality, we consider only horizontal rays. As we can see in Fig.~\ref{fig:fig/RayIntersectionMethod.eps}c, there are only five cases for how a horizontal ray can intersect the curve. 
The three cases at the top 
of Fig.~\ref{fig:fig/RayIntersectionMethod.eps}c, are crosses because the ray goes from one region to the opposite one, while the two cases at the bottom (like in Fig.~\ref{fig:fig/RayIntersectionMethod.eps}b) are not considered crosses because the ray remains
in the same region.

Let $\vec{ X }(i,j)\in \{0,1\}^{1 \times N}$ be a horizontal ray starting from pixel $(i,j)$, which we define as
\begin{align}
    \vec{ X }(i,j) = [X_{i,j}, X_{i,j+1}, X_{i,j+2}, \ldots, X_{i,N}, 0, \ldots, 0],
\end{align}
where zeros are padded to the vector if the ray goes outside the image, such that $\vec{ X }(i,j)$ is always of dimension $N$.
Let $\vec{X}(i, j) \cdot \vec{X}(i+1, j)$ be the inner product of the ray starting from $(i,j)$ and the ray starting from the pixel below, $(i+1,j)$. Note that the contribution to this inner product from the three cases at the top of Fig.~\ref{fig:fig/RayIntersectionMethod.eps}c (the crossings) is odd, whereas the contribution from the other two intersections is even. Thus, the parity of $\vec{X}(i, j) \cdot \vec{X}(i+1, j)$ is the same as the parity of the total number of crosses and determines the insideness of the pixel $(i,j)$,~\ie 
\begin{align}
\left(\boldsymbol{S}(\boldsymbol{ X})\right)_{i,j} = \mbox{parity}\left(\vec{X}(i, j) \cdot \vec{X}(i+1, j)\right).
\label{eq:ParityRay}
\end{align}

In the following we prove that~\eqref{eq:ParityRay} can be  implemented with a neural network with dilated convolutions.   The demonstration is based on implementing the dot product in~\eqref{eq:ParityRay} with multiple layers of dilated convolutions, as they enable capturing the information across the ray, and then, calculating the parity with another neural network. We first introduce a feed-forward convolutional DNN for which there exist parameters that reproduce~\eqref{eq:ParityRay}. Then, we show that one of the layers in this network can be better expressed with multiple dilated convolutions. Finally, we introduce the network to calculate the parity.

\subsection{Overview of the Network to Implement the Ray-Intersection Method}

\begin{figure}[t]
\centering
\footnotesize
\begin{tabular}{cc}
  \includegraphics[width=0.49\linewidth]{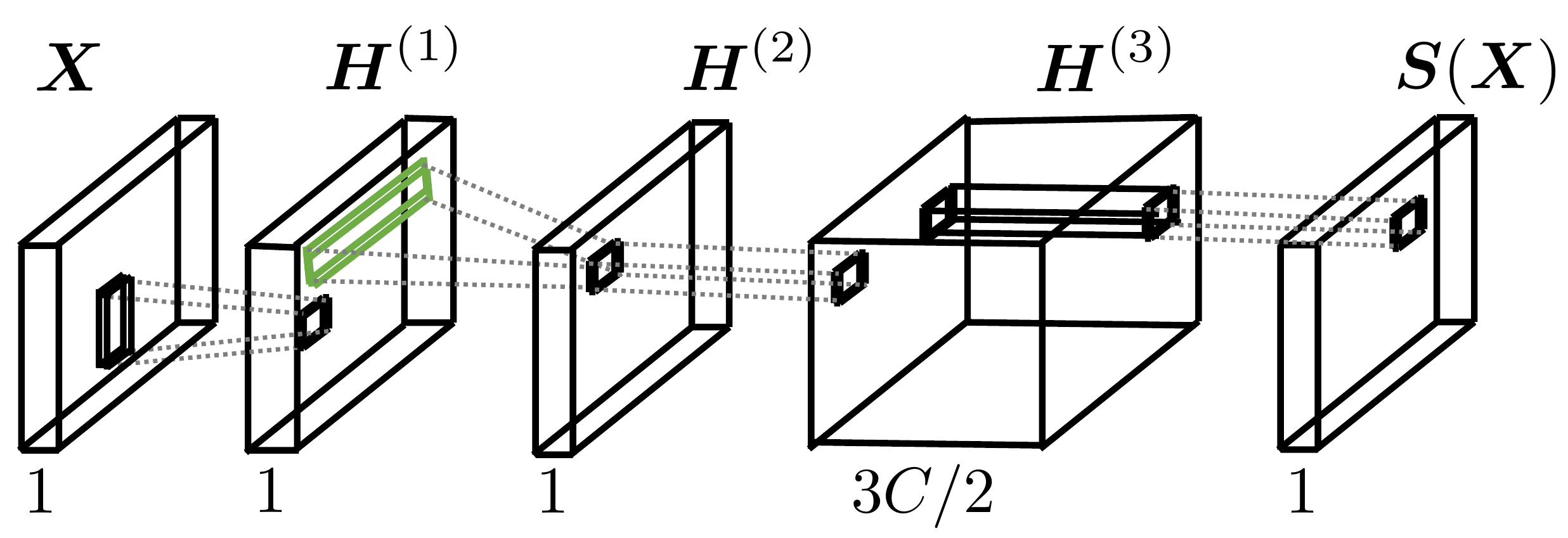} &
\includegraphics[width=0.43\linewidth]{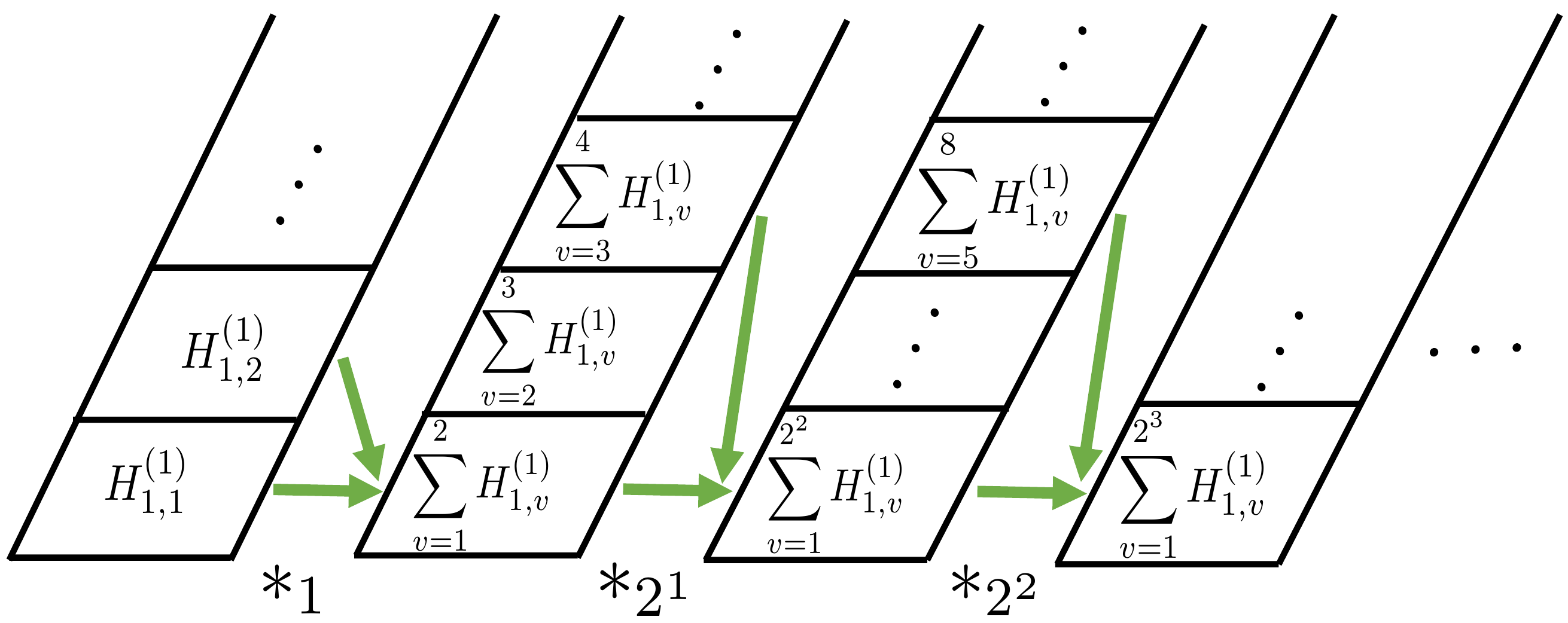}
  \\
  (a) & (b)
  \end{tabular}
    \caption{\emph{Dilated Convolutions from the Ray-Intersection Method}. (a)~The  receptive field colored in green has size $1\times N$, and it can be substituted by an equivalent network composed of multiple dilated convolutions. (b)~The $1\times N$ kernel of the ray-intersection network is equivalent to multiple dilated convolutional layers. The figure shows an horizontal ray of activations from different layers, starting from the first layer $H^{(1)}$. The green arrows indicate the locations in the ray that lead to the desired sum of  activations to implement the $1\times N$ kernel,~\ie the sum of the ray. 
  }
    \label{fig:Type1/2rays.eps}
\end{figure}

The smallest CNN that we found that implements the ray-intersection method has 4-layers. Fig.~\ref{fig:Type1/2rays.eps}a depicts the architecutre.
As we show in the following, the first two layers compute $\vec{X}(i, j) \cdot \vec{X}(i+1, j)$, and the last two layers compute the parity. We use $\boldsymbol{ H}^{(k)}\in \mathbb{R}^{N \times N}$  to denote the activations of the units at the $k$-th layer. These are obtained after applying the \emph{ReLU} activation function, which is denoted as $[ \; ]_+$.

\noindent {\bf First and Second Layer: Inner product.} For the sake of simplicity, we only use  horizontal rays, but the network that we introduce can be easily adapted to any ray direction. The first layer implements all products needed for the inner products across all rays in the image,~\ie $X_{i,j}\cdot X_{i+1,j}, \forall (i,j)$. Note that there is exactly one product per pixel, and each product can be reused for multiple rays. For convenience, $H^{(1)}_{i, j}$ represents the product in pixel $(i,j)$,~\ie $H^{(1)}_{i, j} = X_{i,j}\cdot X_{i+1,j}$.
Since the input consists of binary images, each product can be reformulated as 
\begin{align}
 H^{(1)}_{i, j} =
    \begin{cases}
    1 &\quad \text{if}\quad X_{i,j}=X_{i+1,j}=1\\
    0 & \quad \text{otherwise}
    \end{cases}.
\label{eq:AND}
\end{align}
This equality can be implemented with a \emph{ReLU}: 
$H^{(1)}_{i, j} = [1\cdot X_{i,j} +  1\cdot X_{i+1,j} -1]_+$. Thus, $\boldsymbol{H}^{(1)}$ is a convolutional layer with a $2\times 1$ kernel that detects the intersections shown in Fig.~\ref{fig:fig/RayIntersectionMethod.eps}c. This layer can also be implemented with a standard convolutional layer with a $3\times 3$ kernel, by setting 
the unnecessary elements 
of the kernel to $0$.

The second layer sums over the products of each ray. To do so, we use a kernel  of dimension $1\times N$ with weights equal to $1$ and  bias equal to $0$,~\ie 
${ H}^{(2)}_{i,j} = \boldsymbol{ {\it 1}}_{1 \times N} \cdot {H}^{(1)}_{i, j}=\vec{X}(i, j) \cdot \vec{X}(i+1, j)$, in which
 $\boldsymbol{ {\it 1}}_{I \times J}$ denotes the  matrix of size $I\times J$ with all entries equal to $1$. Zero-padding is used to keep the kernel size constant across the image.
 
 Note that the shape of the kernel, $1\times N$, is not common in the DNN literature. Here, it is necessary  to capture the long-range dependencies of insideness. 
In Appendix~\ref{sec:dilated_kernel}, we show
 that the $1\times N$ kernel can be substituted by multiple layers of dilated convolutions.


\noindent {\bf Third and Fourth Layers: Parity.} To calculate 
the parity of each unit's value  in $\boldsymbol{H}^{(2)}$, we borrow the 
DNN 
introduced by~\cite{SOS17} (namely, Lemma 3 in the supplemental material of the paper).  
This network obtains the parity of any integer bounded by a constant $C$. The network has 
$3C/2$
hidden \emph{ReLUs}
and one output unit, which is $1$ if the input is even, $0$ otherwise (see Appendix~\ref{secSuppParity} for details).

We apply  this parity network to all units in $\boldsymbol{H}^{(2)}$ via convolutions, reproducing the network for each unit. Since a ray through a closed curve in an $N\times N$ image can not have more than $N$ crossings, $C$ is upper bounded by $N$. Thus, the third layer has 
$3N/2$
kernels, and both the third and output layer are convolutions with a $1\times 1$ kernel. 
At this point we have shown that the DNN explained above is feasible in practice, as the number of kernels is $O(N)$, and it requires no more than $4$ convolutional layers with \emph{ReLUs}. The network has a layer with a kernel of size $1\times N$, and next we show that this layer is equivalent to several layers of dilated convolutions of kernel size $3\times 3$.

\subsection{Dilated Convolutions to Implement the $1\times N$ kernel}
\label{sec:dilated_kernel}
We use $*_d$ to denote a dilated convolution, in which $d$ is the dilation factor. Let $\boldsymbol{H}\in \mathbb{R}^{N\times N}$ be the units of a layer and let  $\boldsymbol{K}\in \mathbb{R}^{k\times k}$ be a kernel of size $k\times k$. A dilated convolution is defined as follows:
$\left(\boldsymbol{H} *_d \boldsymbol{K}\right)_{i,j} = \sum_{-\lfloor k/2 \rfloor \leq v,w\leq  \lfloor k/2 \rfloor} H_{i+dv,j+dw} \cdot K_{v,w}$, 
in which $H_{i+dv,j+dw}$  is $0$ if $i+dv$ or  $j+dw$ are smaller than $0$ or larger than $N$,~\ie we abuse notation for the zero-padding.
Note that in the dilated convolution the kernel is applied in a sparse manner, every $d$ units, rather than in consecutive units. 
See~\citet{YuKoltun2016,chen2018deeplab} for more details on dilated convolutions.

Recall the kernel of size $1\times N$ is set to $\boldsymbol{ {\it 1}}_{1 \times N}$ so as to perform the sum of the units corresponding to the ray in the first layer,~\ie $\sum_{0 \leq v < N} H_{i,j+v}^{(1)}$. We can obtain this long-range sum with a series of dilated convolutions 
 using the following $3\times 3$ kernel:
\begin{align}
\boldsymbol{K} 
= 
\left[
\begin{array}{ccc}
0 & 0 & 0 \\
0 & 1 & 1 \\
0 & 0 & 0 
\end{array}
\right].
\end{align}
First, we apply this $\boldsymbol{K}$ to the $\boldsymbol{H}^{(1)}$ through $*_1$ in order to accumulate the first two entries in the ray, which yields:
$\left(\boldsymbol{H}^{(2)}\right)_{i,j} =  \left(\boldsymbol{H}^{(1)} *_1 \boldsymbol{K}\right)_{i,j} =  \sum_{0 \leq v \leq 1} H_{i,j+v}^{(1)}$.
As shown in Fig.~\ref{fig:Type1/2rays.eps}b, to accumulate the next entries of the ray, we can apply $\boldsymbol{K}$ with a dilated convolution of dilation factor $d=2$, which leads to $  \left(\boldsymbol{H}^{(3)}\right)_{i,j} = \sum_{0\leq v < 4} H_{i,j+v}^{(1)}$. To further accumulate more entries of the ray, we need larger dilation factors. It can be seen in  Fig.~\ref{fig:Type1/2rays.eps}b that these dilation factors are powers of $2$, which yield  the following expression:
\begin{align}
  \left(\boldsymbol{H}^{(l)}\right)_{i,j} =  \left(\boldsymbol{H}^{(l-1)} *_{2^{l-2}} \boldsymbol{K}\right)_{i,j} =  \sum_{0\leq v< 2^{l-2}} H_{i,j+v}^{(1)}.
\end{align}
Observe that when we reach layer $l=\log_2(N)+2$, the units accumulate the entire ray of length $N$,~\ie $\sum_{0\leq v < N} H_{i,j+v}^{(1)}$. Networks with dilation factors $d=2^l$ are common in practice,~\eg~\cite{YuKoltun2016} uses these exact dilation factors.

In summary, DNNs with dilated convolutions  can solve the insideness problem and are implementable in practice, since the number of layers and the number of kernels grow  logarithmically and linearly with the image size, respectively.      


\subsection{Parity Network by \cite{SOS17}}

\label{secSuppParity}

To calculate 
the parity of each unit's value  in $\boldsymbol{H}^{(2)}$, we borrow the 
DNN 
introduced by Shalev-Shwartz~\etal (namely, Lemma 3 in the supplemental material of~\cite{SOS17}).  
This network obtains the parity of any integer bounded by a constant $C$. The network has $\frac{3C}{2}$ hidden units with \emph{ReLUs}
and one output unit, which is $1$ if the input is even, $0$ otherwise.  Since such a network requires an upper bound on the number whose parity is being found, we define $C$ as the maximum number of times that a horizontal ray can cross $\mathcal{F}_{\boldsymbol{X}}$. This number can be regarded as an index to express the complexity of the shape.

There is a subtle difference between the network introduced by~\cite{SOS17} and the network we use in the paper. In~\cite{SOS17}, the input of the network is a string of bits, but in our case, the sum is done in the previous layer, through the dilated convolutions. Thus, we use the network in~\cite{SOS17} after the sum of bits is done,~\ie after the first dot product in the first layer in~\cite{SOS17}.

To calculate the parity, for each even number between $0$ and $C$ ($0$ included), $\{2i \; | \; 0\leq i\leq \lfloor C/2 \rfloor \}$, the network has three hidden units that threshold at $(2i - \frac{1}{2})$, $2i$ and $(2i + \frac{1}{2})$,~\ie $-0.5, 0, 0.5, 1.5, 2, 2.5, 3.5, 4, 4.5, \ldots$  The output layer linearly combines all the hidden units and weights each triplet of units by $2$, $-4$ and $2$. Observe that when the input is an odd number, the three units in the triplet are either all  below or all above the threshold. The triplets that are all below the threshold contribute $0$ to the output because the units are inactive, and the triplets that are all above the threshold also contribute $0$ because the linear combination is $2(2i - \frac{1}{2}) - 4(2i) + 2(2i + \frac{1}{2}) = 0$.  For even numbers, the triplet corresponding to that even number  has one unit below, equal and above the threshold. The unit that is above the threshold contributes $1$ to the output, yielding the parity function.

\section{Implementing the \emph{Coloring Routine} with a Convolutional LSTM}
\label{secSuppEXPBLCK}

Here we prove that a ConvLSTM 
can implement the coloring routine, namely, the  iteration of the expansion and the blocking operations.  
A ConvLSTM applied on an image $\boldsymbol{X}$ is defined as the following set of layers (see~\cite{xingjian2015convolutional} for  a comprehensive  introduction to  the ConvLSTM):
\begin{align}
  \boldsymbol{I}^{t} = \sigma\left(\boldsymbol{W}^{xi} * \boldsymbol{X} +\boldsymbol{W}^{hi} * \boldsymbol{H}^{t-1} + \boldsymbol{b}^{i}\right),\\
   \boldsymbol{F}^{t} = \sigma\left(\boldsymbol{W}^{xf} * \boldsymbol{X} +\boldsymbol{W}^{hf} * \boldsymbol{H}^{t-1} + \boldsymbol{b}^{f} \right), \label{equationLSTM:in}  \\
  \boldsymbol{\tilde{C}}^{t} = \tanh \left(\boldsymbol{W}^{xc} * \boldsymbol{X} +\boldsymbol{W}^{hc} * \boldsymbol{H}^{t-1} + \boldsymbol{b}^c \right),\\
  \boldsymbol{C}^{t} = \boldsymbol{F}^{t} \odot \boldsymbol{C}^{t-1} + \boldsymbol{I}^{t} \odot  \boldsymbol{\tilde{C}}^{t},\\
  \boldsymbol{O}^{t} = \sigma\left(\boldsymbol{W}^{xo} * \boldsymbol{X} +\boldsymbol{W}^{ho} * \boldsymbol{H}^{t-1} + \boldsymbol{b}^{o} \right),\label{equationLSTM:o}\\
  \boldsymbol{H}^{t} =   \boldsymbol{O}^{t} \odot \tanh\left(  \boldsymbol{C}^{t} \right),  \label{equationLSTM:H}
\end{align}
where $\boldsymbol{I}^{t}$,  $\boldsymbol{F}^{t}$, $\boldsymbol{C}^{t}$, $\boldsymbol{O}^{t}$ and  $\boldsymbol{H}^{t}\in \mathbb{R}^{N\times N}$ are 
the activation of the units of the input, forget, cell state, output and hidden layers at $t$,
respectively. 
Note that $\boldsymbol{C}^{t}$ has been decomposed with the help of the auxiliary equation 
defining 
$\boldsymbol{\tilde{C}}^{t}$. 
Note also that each of these layers use a different set of weights that are applied to $\boldsymbol{X}$ and to $\boldsymbol{H}^{t}$ 
denoted as $\boldsymbol{W}\in \mathbb{R}^{N\times N}$ with superindices that indicate the connections between layers,~\eg $\boldsymbol{W}^{xi}$ are the weights that connect $\boldsymbol{X}$ to $\boldsymbol{I}$. Similarly, the biases are denoted as $\boldsymbol{b}\in \mathbb{R}^{N\times N}$ with the superindices indicating the layers. 
The symbols $*$ and $\odot$ denote the (usual, not dilated) convolution and  the element-wise product, respectively. 
Finally, $\sigma$ and $\tanh$ are the sigmoid and the hyperbolic tangent, which are used as non-linearities.

We can see by analyzing~\eqref{equationLSTM:o} and~\eqref{equationLSTM:H} that the output layer, $\boldsymbol{O}^{t}$, back-projects to the hidden layer, $\boldsymbol{H}^{t}$. In the coloring algorithm, $\boldsymbol{E}^{t}$ and $\boldsymbol{B}^{t}$ are related in a similar manner. Thus, we define $\boldsymbol{O}^{t}=\boldsymbol{E}^{t}$ (expansion) and $\boldsymbol{H}^{t}=\frac{1}{2} \boldsymbol{B}^{t}$ (blocking), as depicted in Fig.~\ref{fig:fig/ColoringMethod.eps}b. The $\frac{1}{2}$ factor will become clear below, and it does not affect the correctness.  We initialize $\boldsymbol{H}^{0}=\frac{1}{2}\boldsymbol{B}^{0}$ (recall $\boldsymbol{B}^{0}$ is $1$ for all pixels in the border of the image and  $0$ for the rest). 
We now show how to implement the iteration of the expansion and the blocking operations with the ConvLSTM:



\noindent {\bf (i) Expansion, $\boldsymbol{O}^t$:} We set the output layer in~\eqref{equationLSTM:o} in the following way:
\begin{align}
   \boldsymbol{O}^{t} =   \sigma\left(2q\boldsymbol{{\it 1}}_{3\times 3} * \boldsymbol{H}^{t-1} - \frac{q}{2} \boldsymbol{{\it 1}}_{N\times N}  \right).
\end{align}
Note that this layer does not use the input, and sets the convolutional layer $\boldsymbol{W}^{ho}$ to use a $3\times 3$ kernel that is equal to $2q\boldsymbol{{\it 1}}_{3\times 3}$, in which  $q$ is a scalar constant, and the bias equal to $- \frac{q}{2} \boldsymbol{{\it 1}}_{N\times N}$. For very large values of $q$, this layer expands the outside region. This can be seen by noticing that for a unit in $\boldsymbol{H}^{t-1}$, if at least one neighbor has value $1/2$, then
$O_{i,j}^{t} = \lim_{q\rightarrow \infty}\sigma (q)=1$.
Also,  when all neighbouring elements of the unit are $0$,  
then no expansion occurs because $O_{i,j}^{t}  = \lim_{q\rightarrow \infty}\sigma (- \frac{q}{2})=0$. 




\noindent {\bf (ii) Blocking, $\boldsymbol{H}^{t}$:} To stop the outside region from expanding to the inside of the curve, $\boldsymbol{H}^{t}$ takes the expansion output $\boldsymbol{O}^{t}$ and sets the  pixels at the curve's location to $0$ (inside). This is the same as the element-wise product between  $\boldsymbol{O}^{t}$ and the element-wise ``Boolean not'' 
of $\boldsymbol{X}$, which is denoted as $\neg \boldsymbol{X}$. Thus, the blocking operation can be implemented as   $\boldsymbol{H}^{t} = \frac{1}{2}(\boldsymbol{O}^{t} \odot \neg \boldsymbol{X})$. Observe that if $\boldsymbol{C}^{t}=\neg \boldsymbol{X}$, this is equal to~\eqref{equationLSTM:H} of the LSTM, because $\tanh(0)=0$ and $\tanh(1)=1/2$,~\ie
 \begin{align}
   \boldsymbol{H}^{t} =   \boldsymbol{O}^{t} \odot \tanh\left(  \boldsymbol{C}^{t} \right) =   \frac{1}{2} \boldsymbol{O}^{t} \odot  \neg \boldsymbol{X}.
   \label{eqHiddenBlock}
    \end{align}

We can obtain $\boldsymbol{C}^{t}=\neg \boldsymbol{X}$, by imposing $\boldsymbol{I}^{t}=\neg\boldsymbol{X}$ and $\boldsymbol{C}^{t}=\boldsymbol{I}^{t}$, as shown in Fig.~\ref{fig:fig/ColoringMethod.eps}b. To do so, let $\boldsymbol{W}^{xi}=-q\boldsymbol{{\it 1}}_{1 \times 1}$, $\boldsymbol{W}^{hi}= \boldsymbol{ {\it 0}}_{N\times N}$,  and $\boldsymbol{b}^{i}= \frac{q}{2} \boldsymbol{{\it 1}}_{N\times N} $, and ~\eqref{equationLSTM:in} becomes the following expression:
\begin{align}
  \boldsymbol{I}^{t} = \lim_{q\rightarrow \infty} \sigma\left(-q\boldsymbol{{\it 1}}_{1\times 1} * \boldsymbol{X} +  \frac{q}{2} \boldsymbol{{\it 1}}_{N\times N} \right).
  \end{align}
Observe that when $q$ tends to infinity, we have $ I_{i,j}^{t} = \lim_{q\rightarrow \infty}\sigma(\frac{q}{2})=1$ when $X_{i,j}=0$ and $ I_{i,j}^{t}  = \lim_{q\rightarrow \infty}\sigma(-\frac{q}{2})=0$ when $X_{i,j}=1$, which means  $\boldsymbol{I}^{t}= \neg \boldsymbol{X} $. Next, to obtain $\boldsymbol{C}^{t}=\boldsymbol{I}^{t}$, 
we set $\boldsymbol{W}^{xf} = \boldsymbol{W}^{hf} = \boldsymbol{W}^{xc} = \boldsymbol{W}^{hc}= \boldsymbol{{\it 0}}_{N \times N}, \boldsymbol{b}^{f} = - q \boldsymbol{{\it 1}}_{N \times N}$ and $\boldsymbol{b}^{c} = q \boldsymbol{{\it 1}}_{N \times N}$. 
This leads to the desired result:
 \begin{align}
  \boldsymbol{F}^{t} =  \lim_{q\rightarrow \infty} \sigma\left(-q \boldsymbol{{\it 1}}_{N \times N} \right) = \boldsymbol{{\it 0}}_{N\times N},\\
   \boldsymbol{\tilde{C}}^{t} =  \lim_{q  \rightarrow \infty} \tanh \left(q \boldsymbol{{\it 1}}_{N \times N} \right) = \boldsymbol{{\it 1}}_{N\times N},\nonumber \\
  \boldsymbol{C}^{t} = \boldsymbol{{\it 0}}_{N\times N} \odot \boldsymbol{C}^{t-1} + \boldsymbol{I}^{t} \odot  \boldsymbol{{\it 1}}_{N\times N} =  \boldsymbol{I}^{t} = \neg \boldsymbol{X}. 
\end{align}

Thus, the coloring method can be implemented with a network as small as one ConvLSTM with one kernel. 
A network with more than one kernel and multiple stacked ConvLSTM  can also solve the insideness problem for any given curve. The kernels that are not needed to implement the coloring method can be just set to $0$ and the ConvLSTM that are not needed should implement the identity operation,~\ie the output layer is equal to the input.
To implement the identity operator, \eqref{equationLSTM:o} can be rewritten in the following way:
\begin{align}
    \boldsymbol{O}^{t} = \lim_{q\rightarrow \infty} \sigma\left(q\boldsymbol{{\it 1}}_{1\times 1} * \boldsymbol{X} -  \frac{q}{2} \boldsymbol{{\it 1}}_{N\times N} \right)
\end{align}
where $\boldsymbol{W}^{ho}=\boldsymbol{{\it 0 }}_{1\times 1}$ is to remove the connections with the hidden units, and $q$ is the constant that tends to infinity. Observe that if $X_{i,j}=1$, then $\boldsymbol{O}^{t} =\lim_{q\rightarrow \infty} \sigma(q/2)=1$. If $X_{i,j}=0$, then
$\boldsymbol{O}^{t} =\lim_{q\rightarrow \infty} \sigma(-q/2)=0$. Thus, the ConvLSTM implements the identity operation.

\section{Coloring Routine with a Sigmoidal Convolutional RNN}
\label{secRNN}

There are other  recurrent networks simpler than a ConvLSTM that can also implement the coloring algorithm. We introduce here a convolutional recurrent network that uses sigmoids as non-linearities. Since it is a convolutional network, for the sake of simplicity we just describe the operations done to obtain an output pixel in a step. The network has only one hidden layer, which also corresponds to the output of the network.
Let $\{h_k^t\}_{k\in \mathcal{N}_{i,j}}$ be the hidden state of  the output pixel indexed by $i,j$ and its 4-neighbourhood, at step $t$. Let $X_{i,j}$ be the only relevant input image pixel. A necessary condition is that the outputs of the sigmoid should asymptotically be close to $0$ or $1$, otherwise the coloring routine would fade after many steps. It is easy to check that $h^{t+1}_{i,j} = \sigma\left(q\left(\sum_{k\in\mathcal{N}_{ij}} h_{k}^t-5X_{i,j}-{1/2}\right)\right)$ implements the coloring routine, where $q$ is the factor that ensures saturation of the sigmoid.

\section{Dataset Generation}

\label{secSuppDataset}

\begin{figure}[t]
  \footnotesize
  \centering
  \begin{tabular}{c c}
    \multicolumn{2}{c}{\bf Examples of Jordan Curves of Each Dataset }\\
    \includegraphics[width=0.47\textwidth]{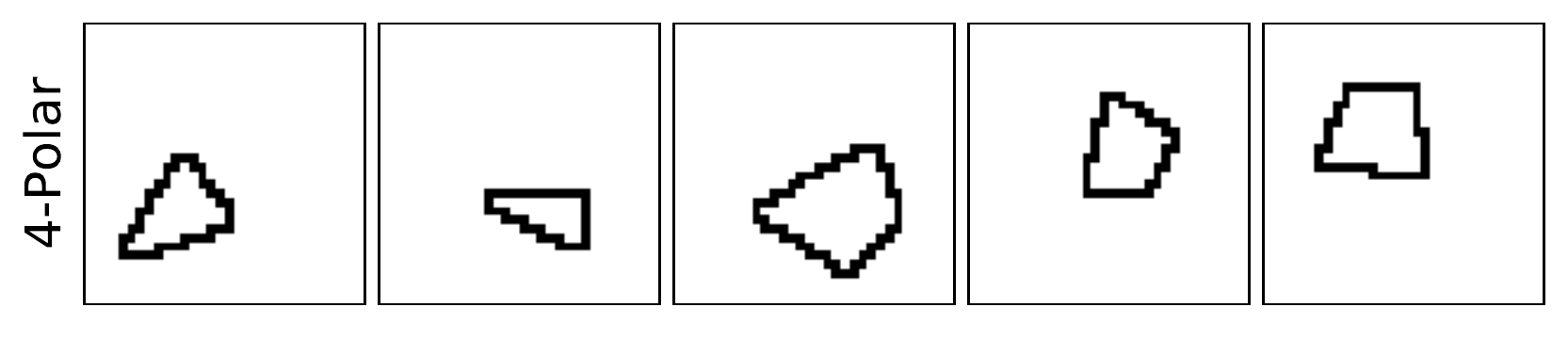}&
    \includegraphics[width=0.47\textwidth]{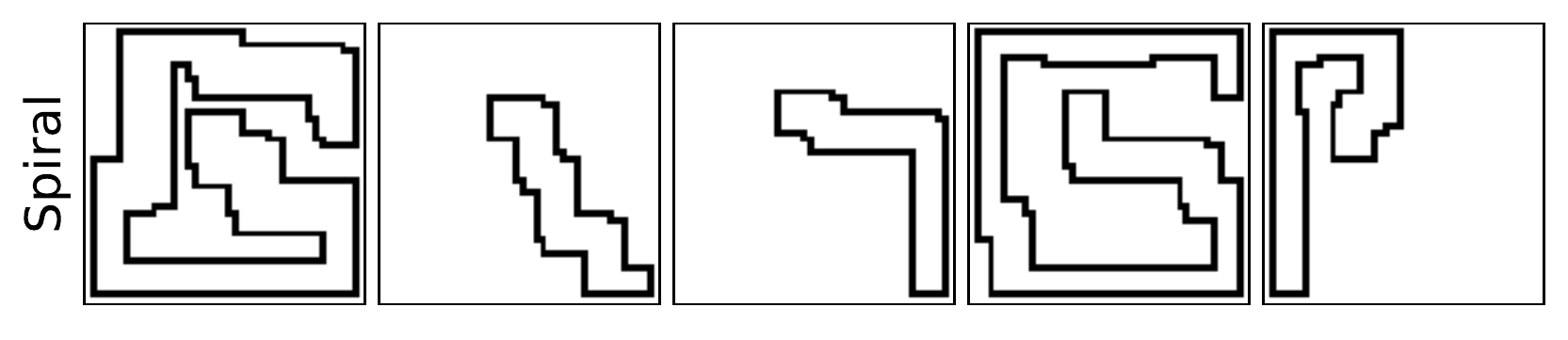}\\ 
    \includegraphics[width=0.47\textwidth]{./fig/dataset/10.pdf}&
    \includegraphics[width=0.47\textwidth]{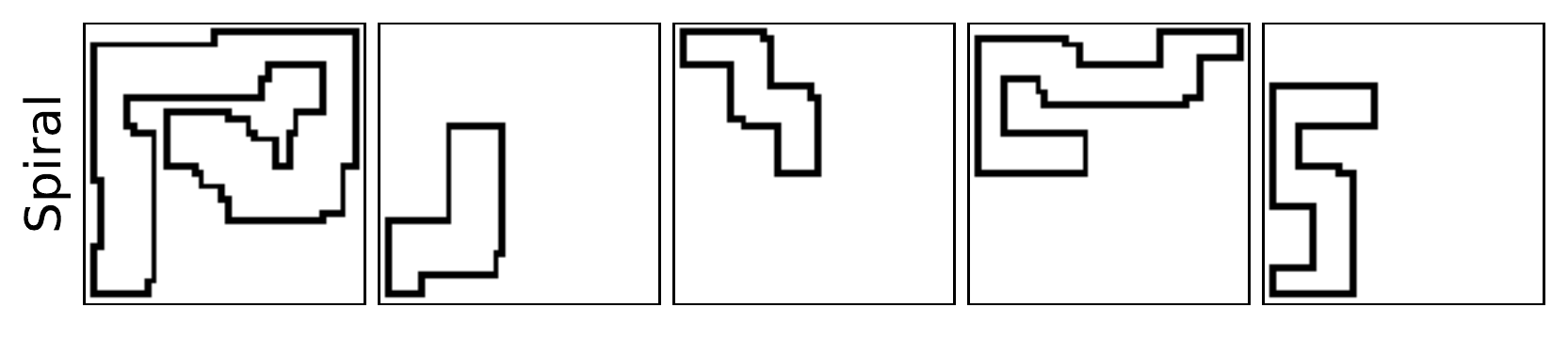} \\
    \includegraphics[width=0.47\textwidth]{./fig/dataset/12.pdf}&
    \includegraphics[width=0.47\textwidth]{./fig/dataset/5.pdf}\\ 
    \includegraphics[width=0.47\textwidth]{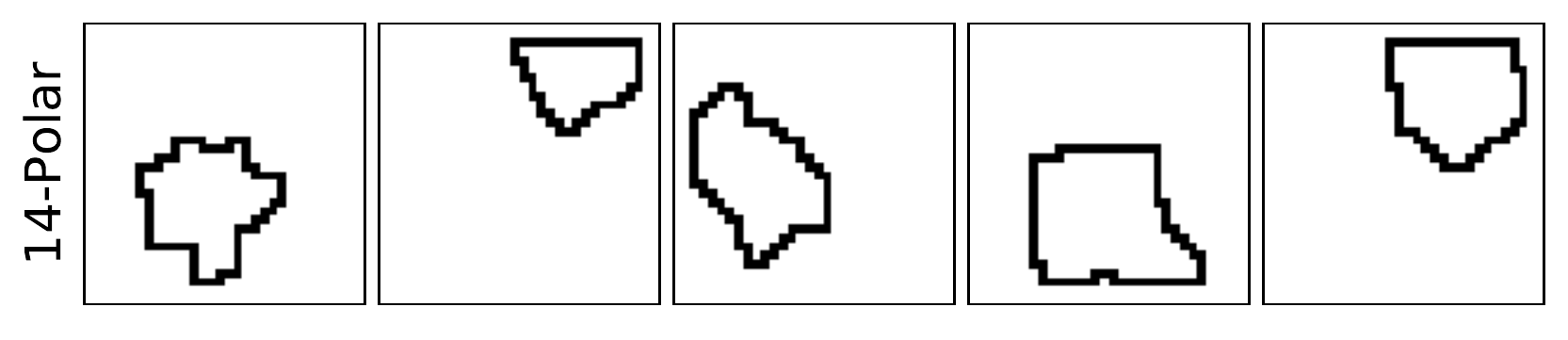}&
    \includegraphics[width=0.47\textwidth]{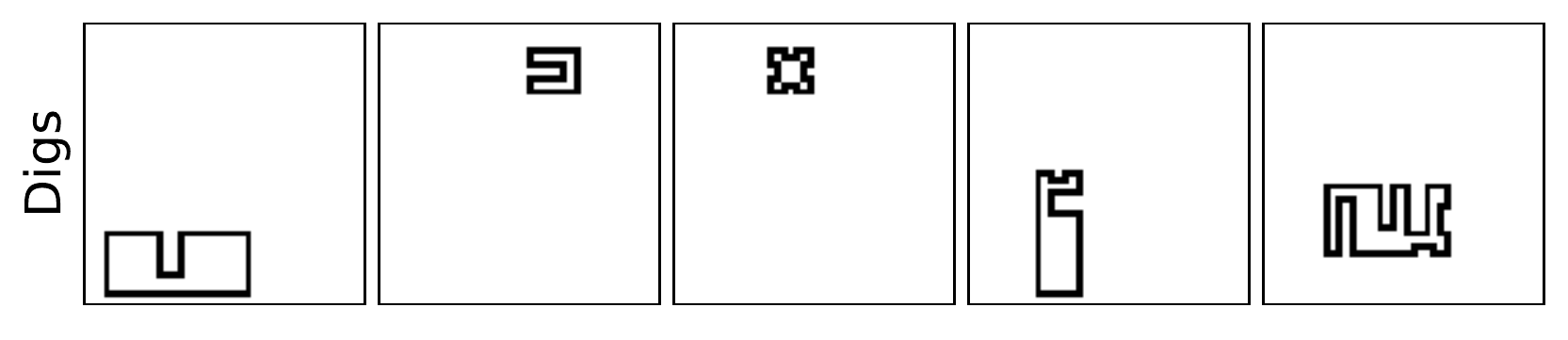}\\ 
    \includegraphics[width=0.47\textwidth]{./fig/dataset/0.pdf}&
    \includegraphics[width=0.47\textwidth]{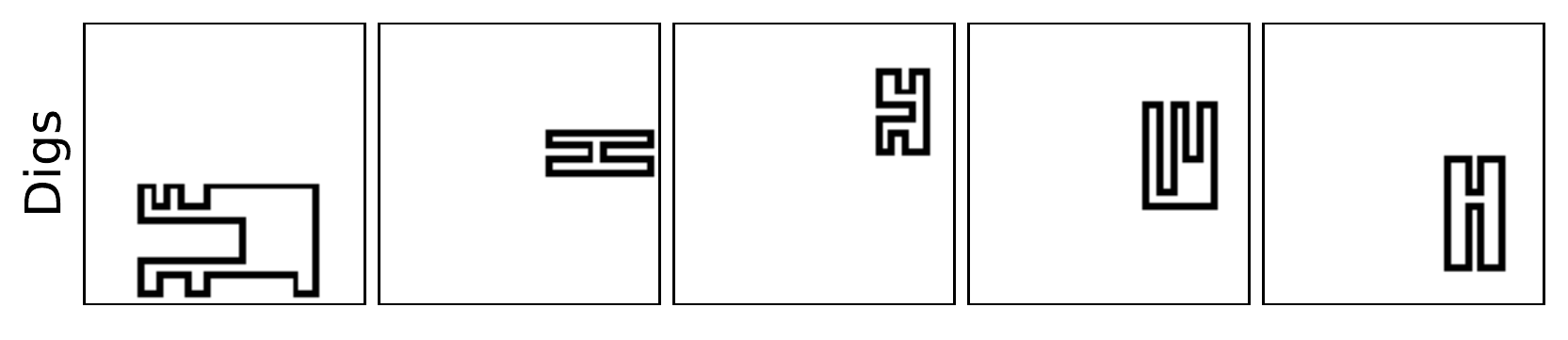}\\
    \includegraphics[width=0.47\textwidth]{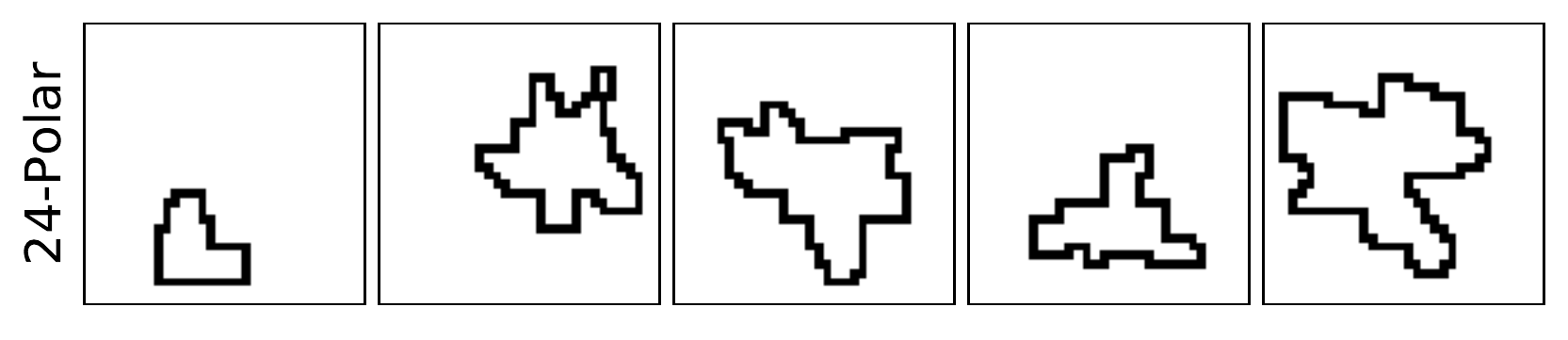}&
    \includegraphics[width=0.47\textwidth]{./fig/dataset/8.pdf}
  \end{tabular}
\caption{\emph{Datasets.} Images of the curves used to train and test the DNNs. Each row correspond to a different dataset.}
\label{fig:SuppDatasets}
\end{figure}

In Fig.~\ref{fig:SuppDatasets}, we show more examples of curves in the datasets. For testing and validation sets, we only use images that are dissimilar to all images from the training set. Two images are considered dissimilar if at least $25\%$ of the pixels of the curve are in different locations.  In the following we provide a  description of the algorithms to generate the curves:

\noindent - \emph{Polar Dataset} ($32\times 32$ pixels): We use polar coordinates to generate this dataset. We randomly select the center of the figure and a random number of vertices that are connected with straight lines. These lines are constrained to follow the definition of digital Jordan curve in Section~\ref{secJordan} in the main paper (and Appendix~\ref{secSuppJordan} in this supplementary material). The vertices are determined by their angles, which are randomly generated. The distance with respect to the center of the figure are also randomly generated to be between 3 to 14 pixels away from the center. 

We generate $5$ datasets with different maximum amount of vertices, namely, $4$, $9$, $14$, $19$ and $24$. We refer to each of these datasets as  \emph{Polar} with a prefix with the amount of vertices.

\renewcommand{\arraystretch}{0}

\noindent - \emph{Spiral Dataset} ($42\times 42$ pixels): The curves in these data set are generated from a random walk. First, a starting position is chosen uniformly at random from $[10,20]\times [10,20]$. Then, a segment of the spiral is built in the following way: a random direction (up, down, left, right) and a random length $r$ from $3$ to $10$ are chosen so that the walk is extended by turning $r$ pixels in the given direction. However, such extension can only happen if adding a random thickness $t\in \{1,2,3,4\}$ to both sides of this segment does not cause self intersections. These segments are added repeatedly until there is no space to add a new segment without violating the definition of a Jordan curve.

\noindent - \emph{Digs Dataset} ($42\times 42$ pixels): We generate a rectangle of random 
size 
and then, we create  ``digs'' of random thicknesses in the 
rectangle. The number of ``digs'' is a random number between $1$ to $10$.
The digs are created sequentially and they are of random depth (between $1$ pixel to the length of the rectangle minus $2$ pixels). For each new ``dig'', we made sure to not cross previous digs by adjusting the depth of the ``dig''. 

\section{Hyperparameters}
\label{secHyper}

The parameters are initialized using Xavier initialization \citep{Glorot10understandingthe}. The derived parameters we obtained in the theoretical demonstrations obtain $100\%$ accuracy but we do not use them in this analysis as they are not learned from examples.
The ground-truth consists on the insideness for each pixel in the image, as in~\eqref{ConditionOfAnswer}. For all experiments, we use the cross-entropy with softmax as the loss function averaged accross pixels. Thus, the networks have two outputs per pixel (note that this does not affect the result that the networks are sufficiently complex to solve insideness, as the second output can be set to a constant threshold of $0.5$). We found that the cross-entropy loss leads to better accuracy than other losses. Moreover, we found that using a weighted loss improves the accuracy of the networks. The weight, which we denote as $\alpha$, multiplies the loss relative to inside, and $(1-\alpha)$ multiplies the loss relative to outside. 
This $\alpha$ is a hyperparamter that we tune and can be equal to $0.1$, $0.2$ and $0.4$. We try batch sizes of $32$, $256$ and $2048$ when they fit in the GPUs' memory  ($12$GB), and we try learning rates from $1$ to $10^{-5}$ (dividing by $10$). We train the networks for all the hyperparameters for at least $50$ epochs, and until there is no more improvement of the validation set loss. 

  In the following we report all the tried hyperparameters for all architectures. In all cases, the convolutional layers use zero-padding.
  
\noindent - \emph{Dilated Convolution DNN (Dilated):} This network was introduced in Section~\ref{secDilation}. We use the same hyperparameters as in~\cite{YuKoltun2016}: $3\times 3$ kernels, a number of kernels equal to $2^l\times \{2, 4, 8\}$, where \textit{l} is the number of layers and ranges between $8$ to $11$, with $d=2^l$ (the first layer and the last two layers $d=1$). The number of kernels in the layer that calculates the parity can be $\{5, 10, 20, 40, 80\}$.

\noindent - \emph{Ray-intersection network (Ray-int.):} This is the architecture introduced in Section~\ref{secDilation}, which uses a receptive field of $1\times N$ instead of the dilated convolutions. The rest of the hyperparameters are as in \emph{Dilated}.

\noindent - \emph{Convolutional DNN (CNN):} To analyze the usefulness of the dilated convolutions, we use the \emph{Dilated} architecture with all dilation factors $d=1$. Also, we try adding more layers than in \emph{Dilated}, up to $25$.

\noindent - \emph{UNet:} This is a popular architecture with skip connections and de-convolutions. We use similar hyperparameters as in~\cite{RFB15}:  starting with $64$ kernels ($3\times 3$) at the first layer and doubling this number after each max-pooling; a total of $1$ to $3$ max-pooling layers in all the network, that are placed after sequences of $1$ or $2$ convolutional layers.

\noindent - \emph{Convolutional LSTM (1-LSTM):} This is the architecture with just one ConvLSTM, introduced in Section~\ref{secLSTM}. We use backpropagation through time by unrolling $60$, $30$ or $10$ time steps for training (we select the best performing one). For testing, we unroll until there is no change in the output labeling. We initialize the hidden and cell states to $0$ (inside) everywhere except the border of the image which is initialized to $1$ (outside).

\noindent - \emph{2-layers Convolutional LSTM (2-LSTM):} We stack one convolutional LSTM after another. The first LSTM has $64$ kernels, and 
the hidden and cell states
are initialized as in the 1-LSTM.

\noindent - \emph{2-layers Convolutional LSTM without initialization (2-LSTM w/o init.):} this is the same as the \emph{2-LSTM} architecture the hidden and cell states  are  
initialized
to $0$ (outside).

\section{Regularizing Feed-Forward Networks}
\label{secSuppDilated}



In Fig.~\ref{fig:General2}, we have observed that \emph{Dilated} trained on both \emph{24-Polar} and \emph{Spiral} datasets, obtains a test accuracy of less than $95$\% on these datasets while the accuracy in the training set is very close to $100\%$. We added weight decay in all the layers in order to regularize the network. We tried values between $10^{-5}$ to $1$, scaling by a factor of $10$. In all these experiments we have observed overfitting except for a weight decay of $1$, in which the training never converged.

Also, note that the 
\emph{CNN} 
does not have this overfitting problem. Yet, the number of layers needed is $25$, which is more than the double than for \emph{Dilated}, which is $9$ layers. We added more layers to \emph{Dilated}  but the accuracy did not improve.





\clearpage

\section{Additional Figures and Visualizations} \label{Additional_Figures_and_Visualizations}

\begin{figure*}[ht]
  \footnotesize
  \centering
  \begin{tabular}{c@{\hspace{-0.01cm}}c@{\hspace{-0.01cm}}c}
        \includegraphics[width=0.3\textwidth]{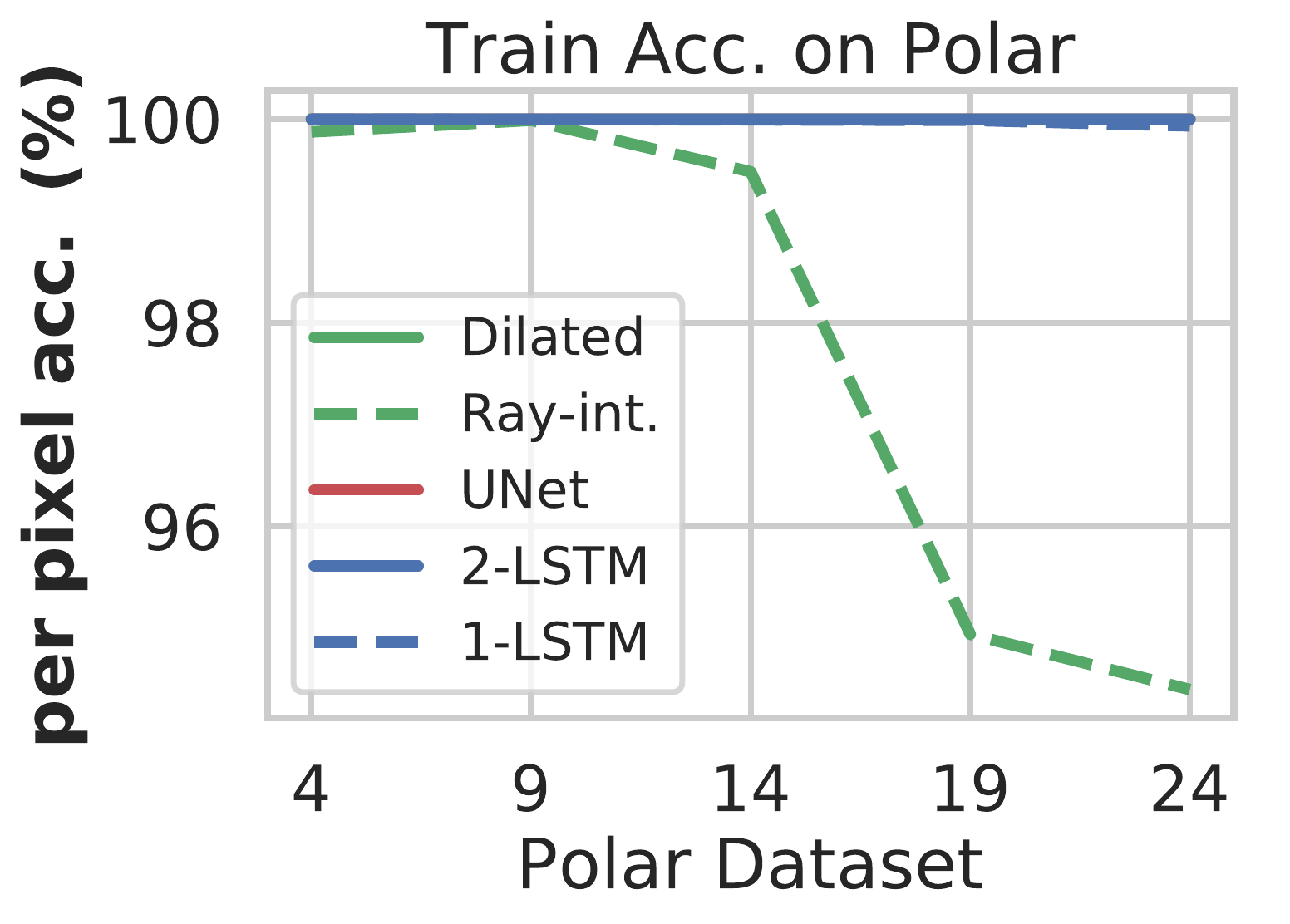}&
        \includegraphics[width=0.3\textwidth]{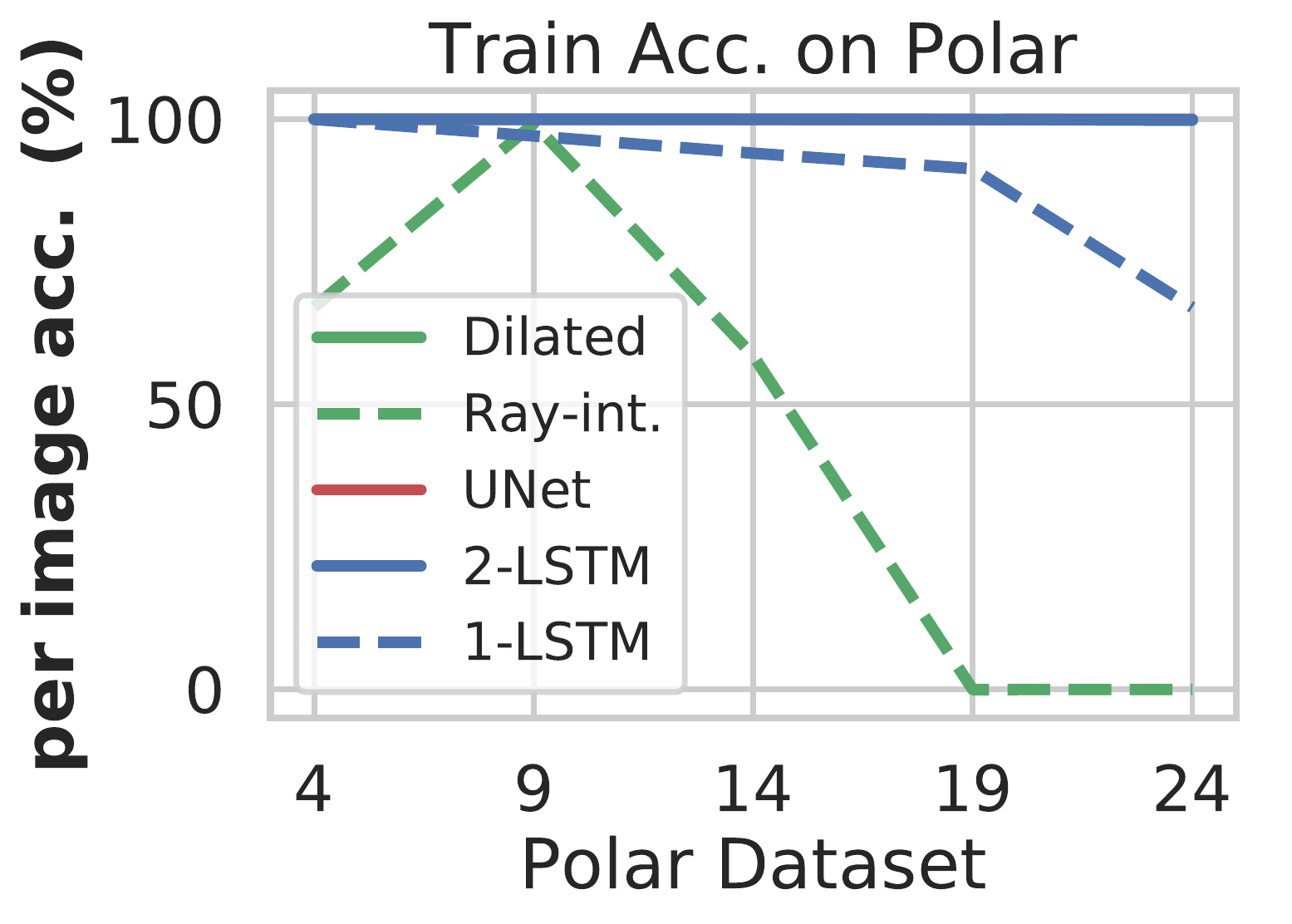} &\includegraphics[width=0.3\textwidth]{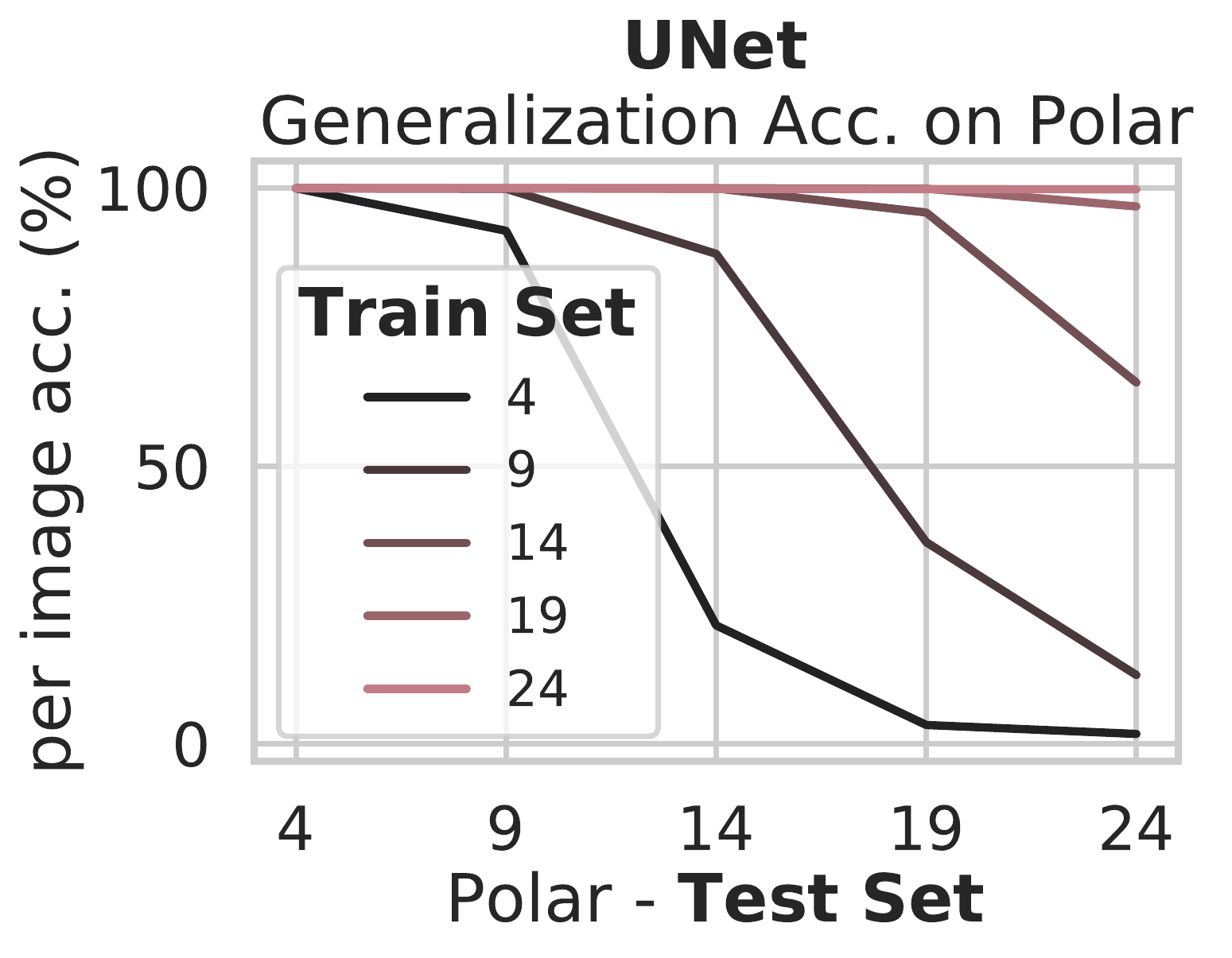}\\
        (a) & (b) & (c)
  \end{tabular}
\caption{\emph{Training Accuracy in the Polar Dataset.} Intra-dataset evaluation using (a) per pixel accuracy and (b) per image accuracy on the training set, which are very similar to the test accuracy reported in Figs.~\ref{fig:Datasets}b and~c. (c) Intra-dataset evaluation of \emph{Unet}.}
\label{resSuppQuant1}
\end{figure*}

\renewcommand{\arraystretch}{0}

\begin{figure*}[ht]
  \centering
  \footnotesize
  \begin{tabular}{c@{\hspace{-0.06cm}}c@{\hspace{-0.06cm}}c@{\hspace{-0.06cm}}c@{\hspace{-0.06cm}}c@{\hspace{-0.06cm}}c}
\multicolumn{6}{c}{\bf Layer 2}\\ [0.1cm]
    { Unit 0}& { Unit 2}& { Unit 12}& { Unit 24}& { Unit 26}& { Unit 28}\\
        \includegraphics[width=0.165\textwidth]{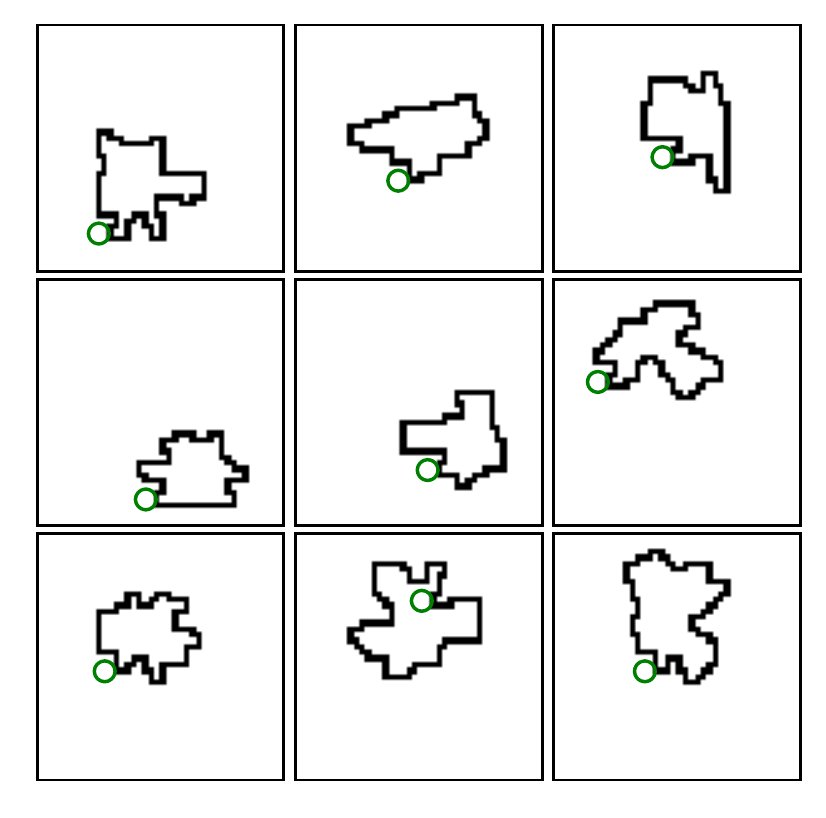}&
    \includegraphics[width=0.165\textwidth]{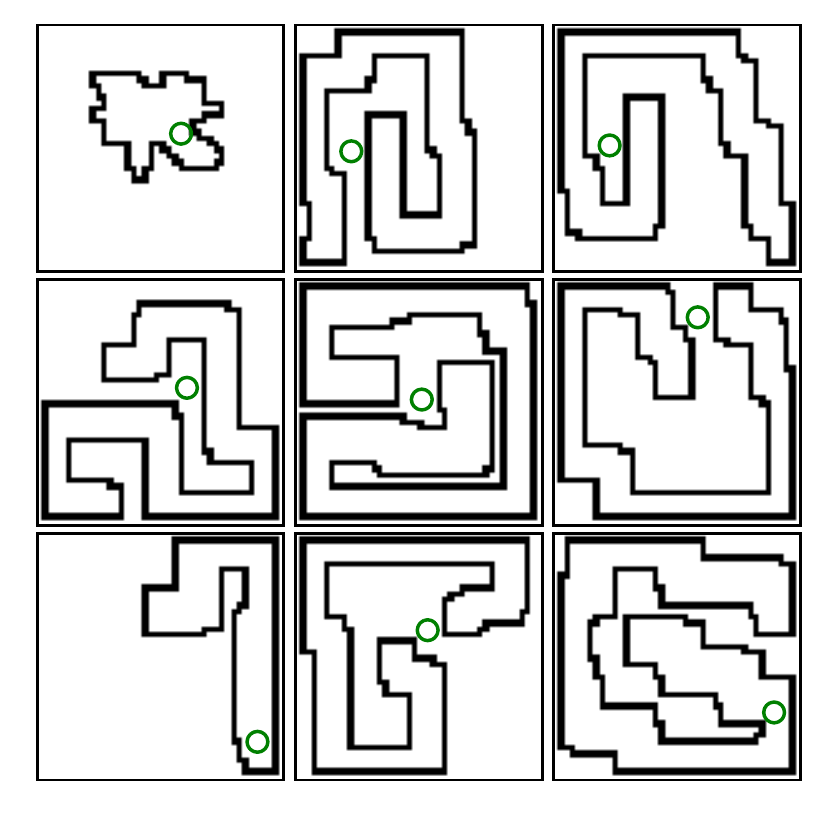}&
    \includegraphics[width=0.165\textwidth]{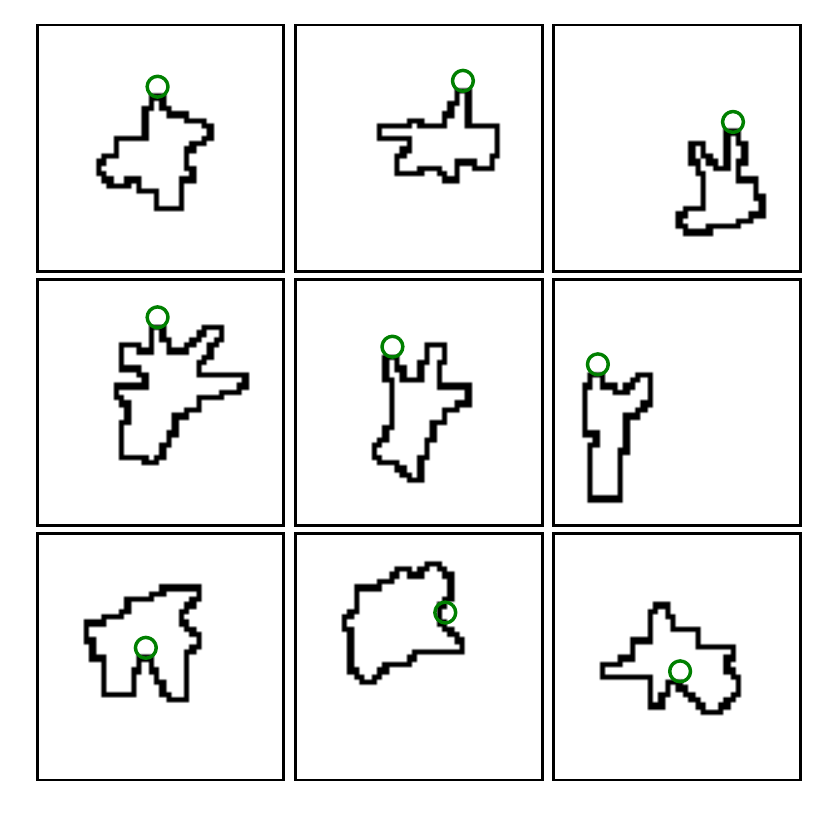}&
    \includegraphics[width=0.165\textwidth]{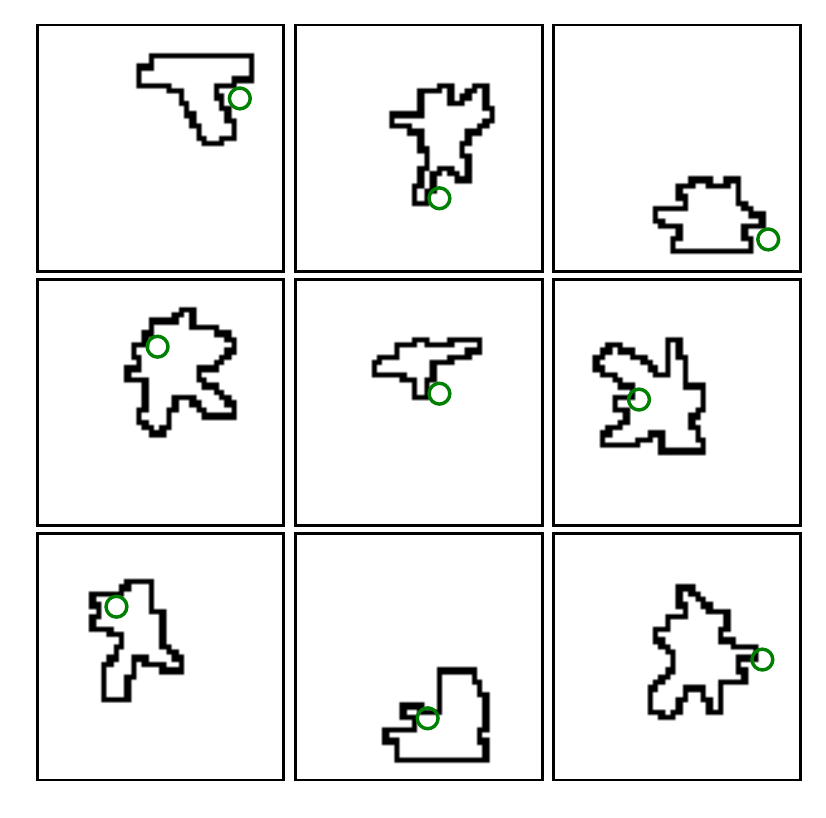}&
    \includegraphics[width=0.165\textwidth]{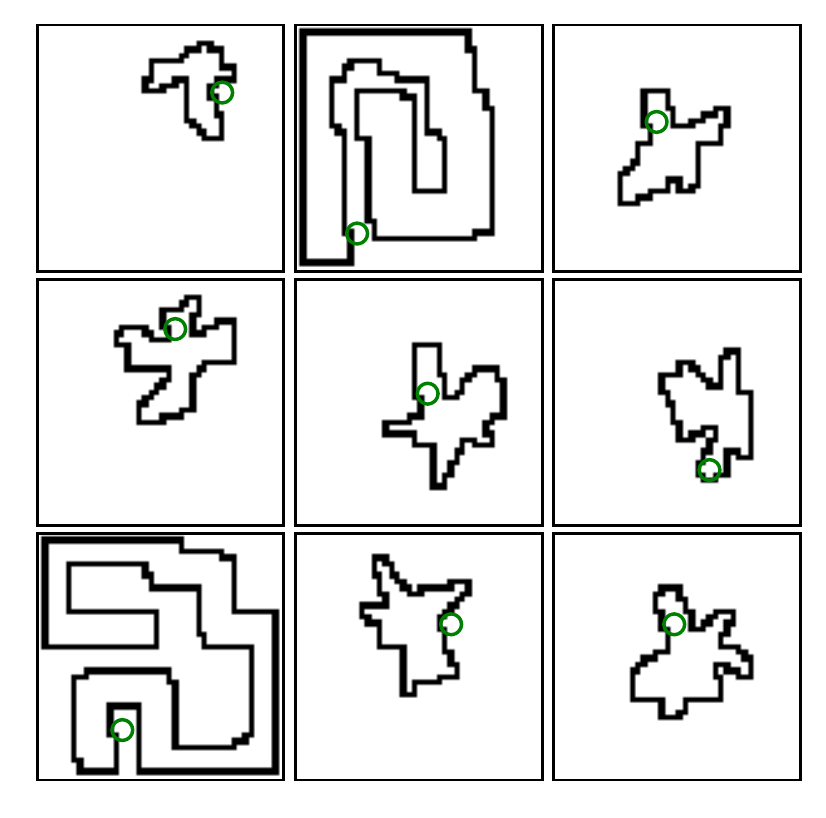}&
     \includegraphics[width=0.165\textwidth]{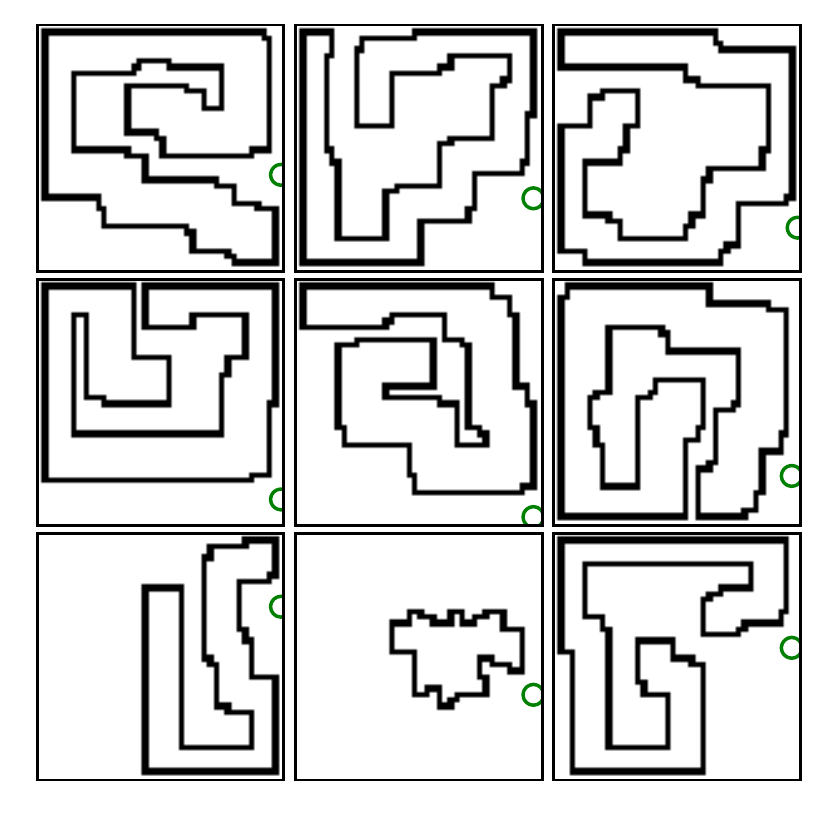}\\[0.1in]
     \multicolumn{6}{c}{\bf Layer 4}\\ [0.1cm]
    { Unit 0}& { Unit 3}& { Unit 4}& { Unit 7}& { Unit 10}& { Unit 14}\\
        \includegraphics[width=0.165\textwidth]{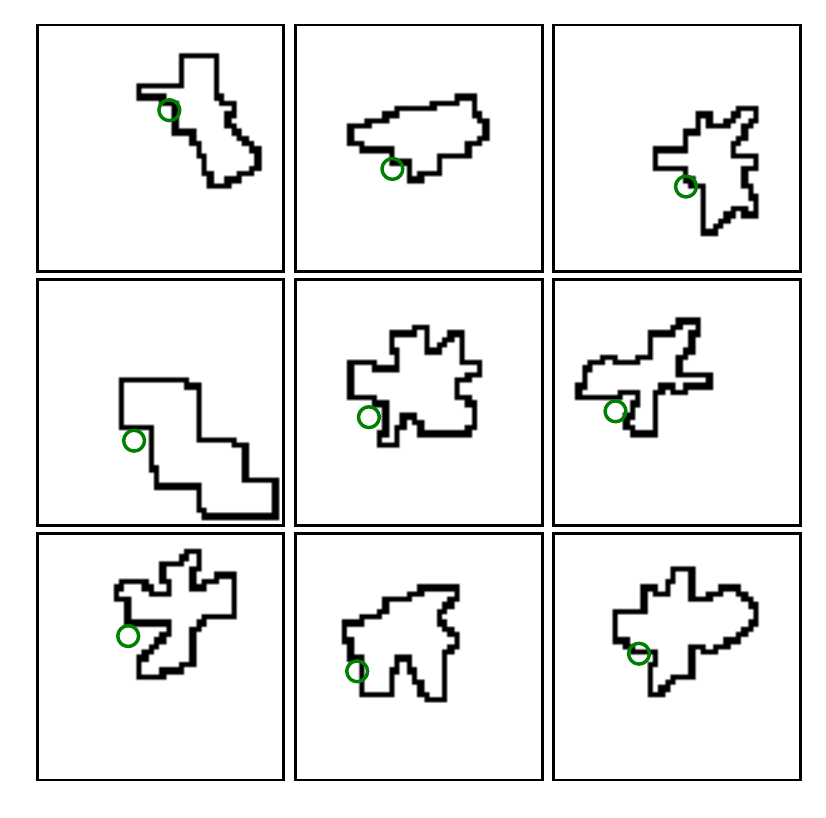}&
    \includegraphics[width=0.165\textwidth]{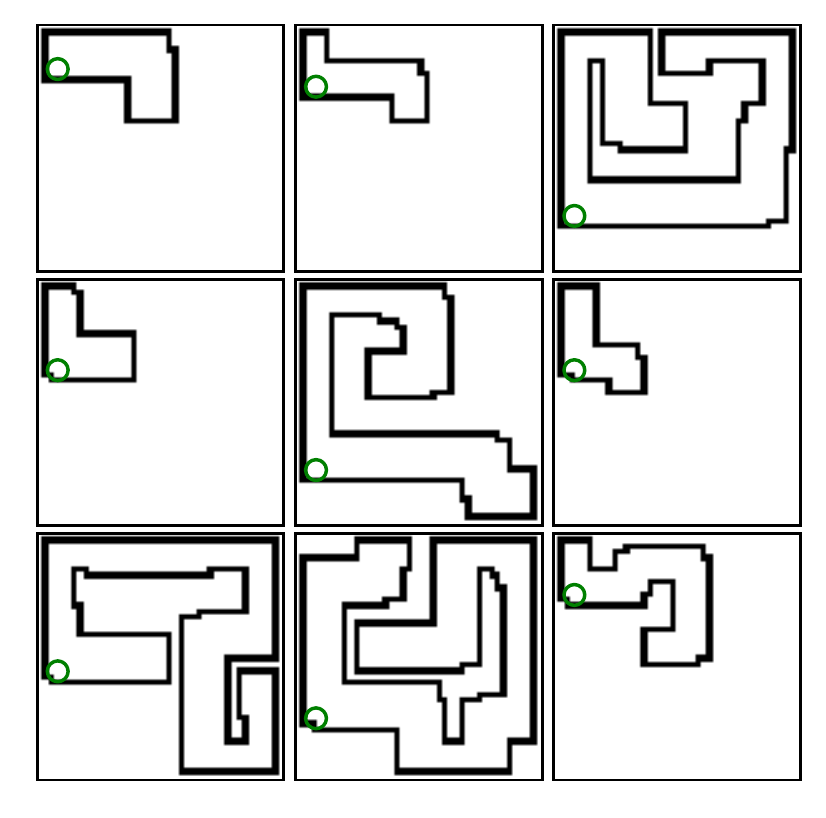}&
    \includegraphics[width=0.165\textwidth]{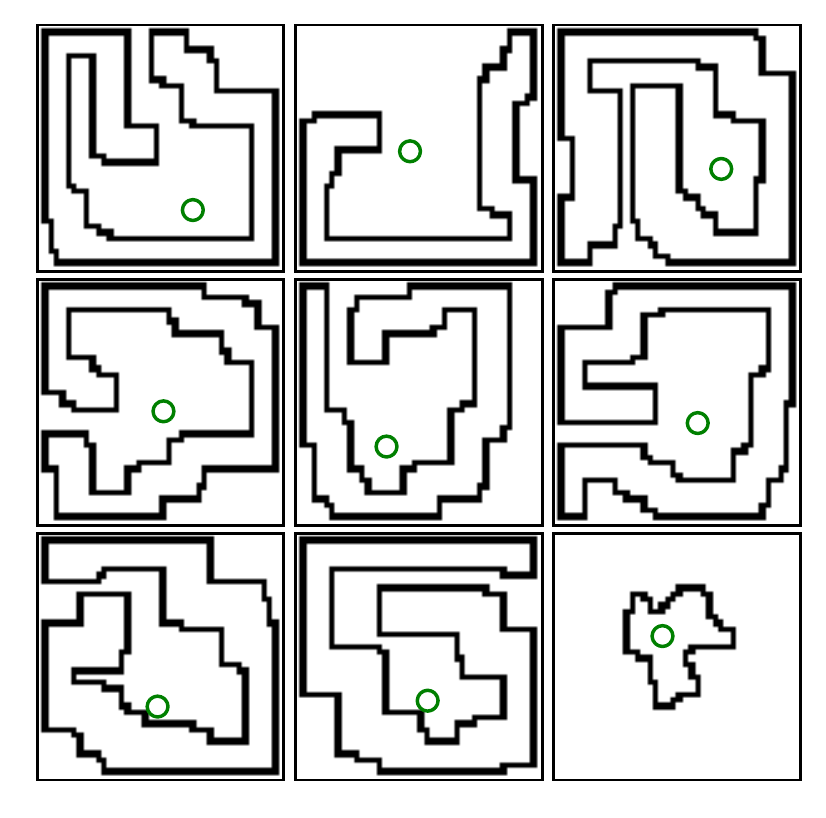}&
    \includegraphics[width=0.165\textwidth]{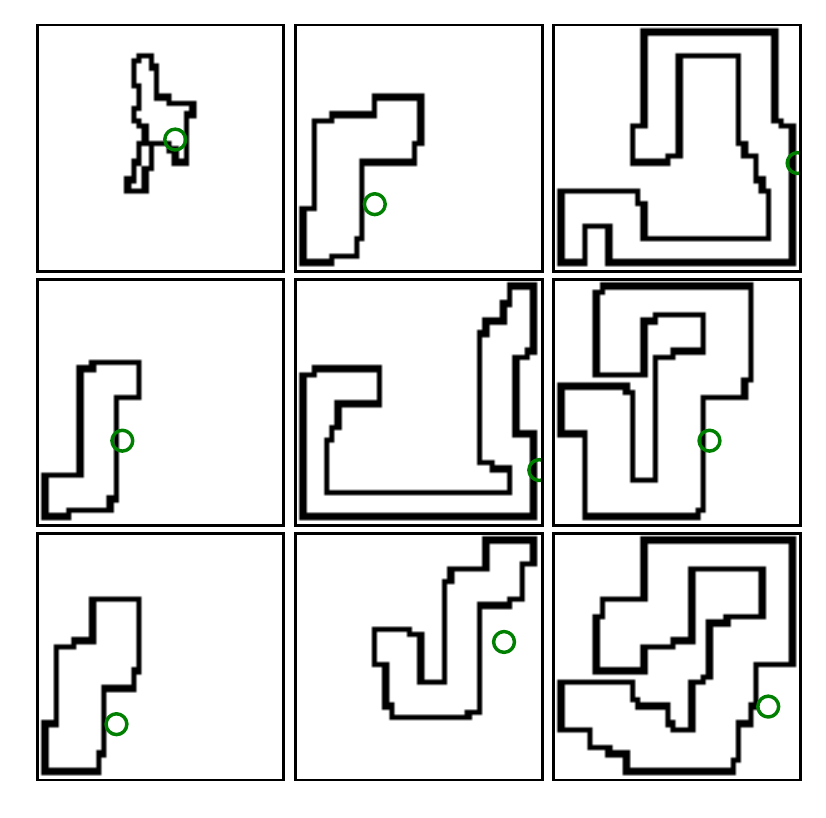}&
    \includegraphics[width=0.165\textwidth]{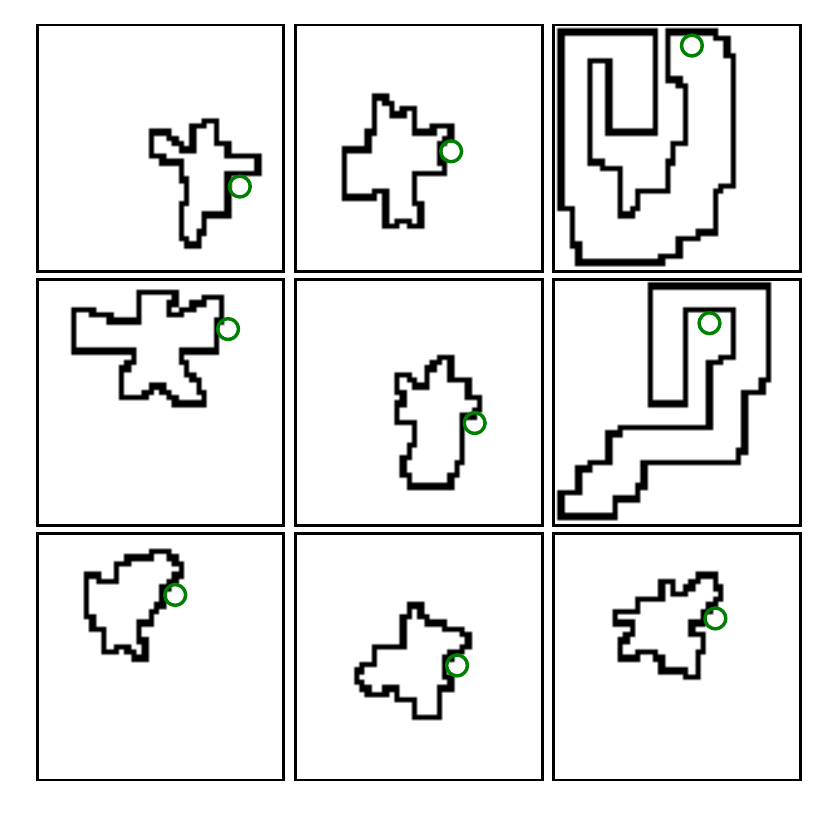}&
     \includegraphics[width=0.165\textwidth]{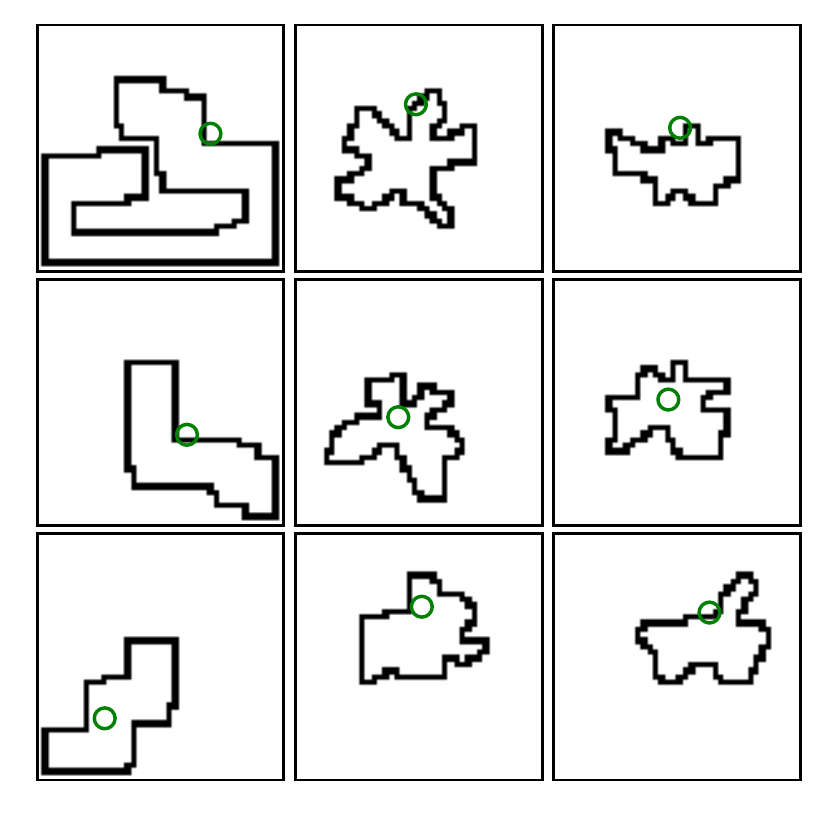}\\[0.1in]
     \multicolumn{6}{c}{\bf Layer 6}\\ [0.1cm]
    { Unit 2}& { Unit 4}& { Unit 8}& { Unit 16}& { Unit 17}& { Unit 21}\\
        \includegraphics[width=0.165\textwidth]{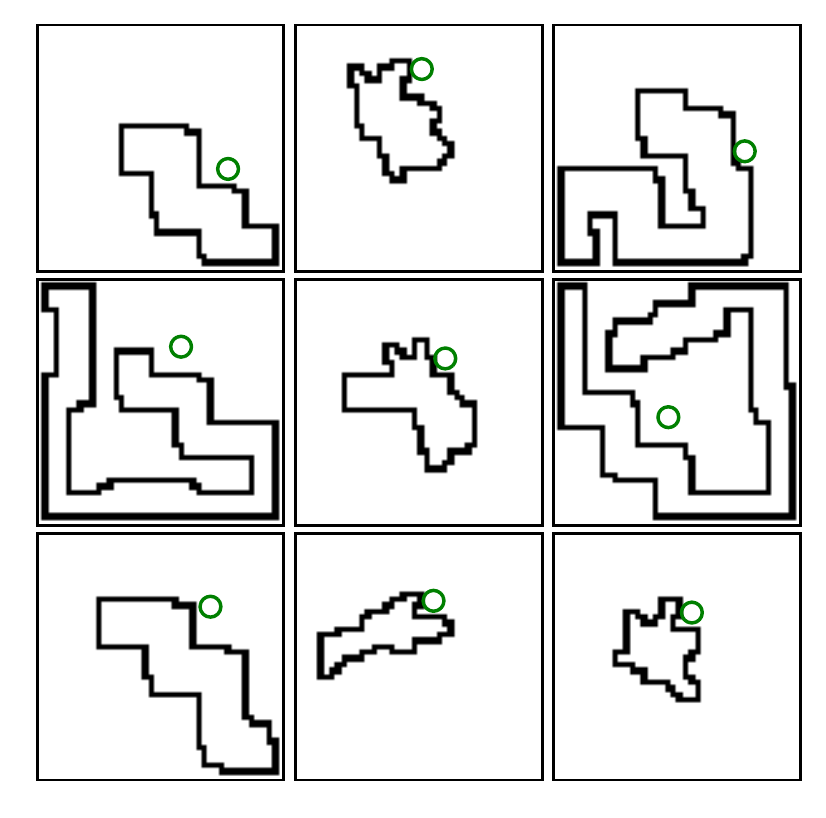}&
    \includegraphics[width=0.165\textwidth]{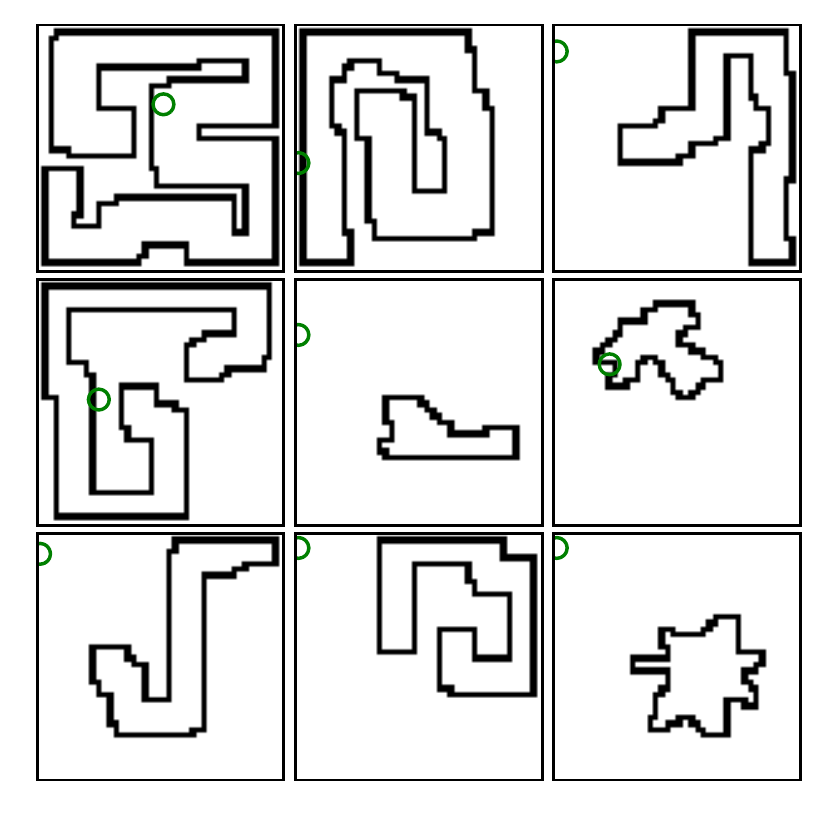}&
    \includegraphics[width=0.165\textwidth]{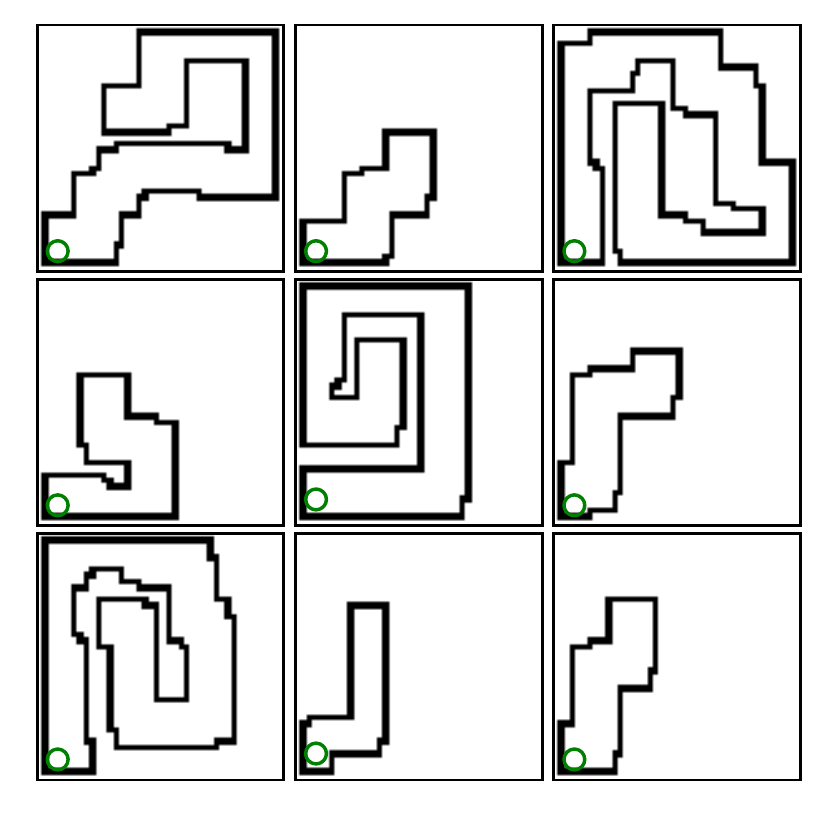}&
    \includegraphics[width=0.165\textwidth]{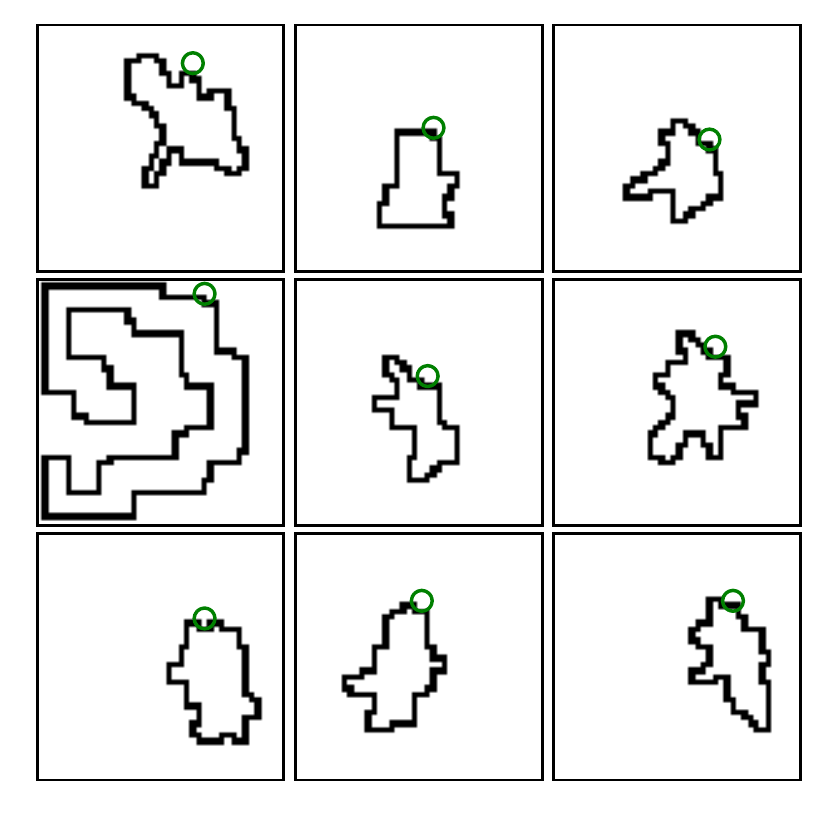}&
    \includegraphics[width=0.165\textwidth]{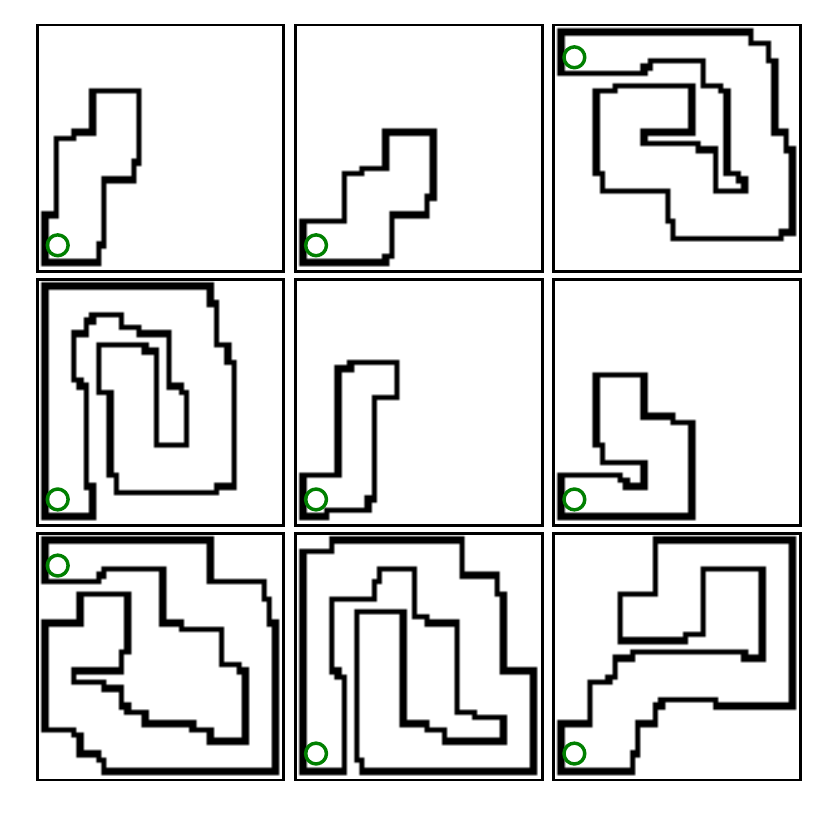}&
     \includegraphics[width=0.165\textwidth]{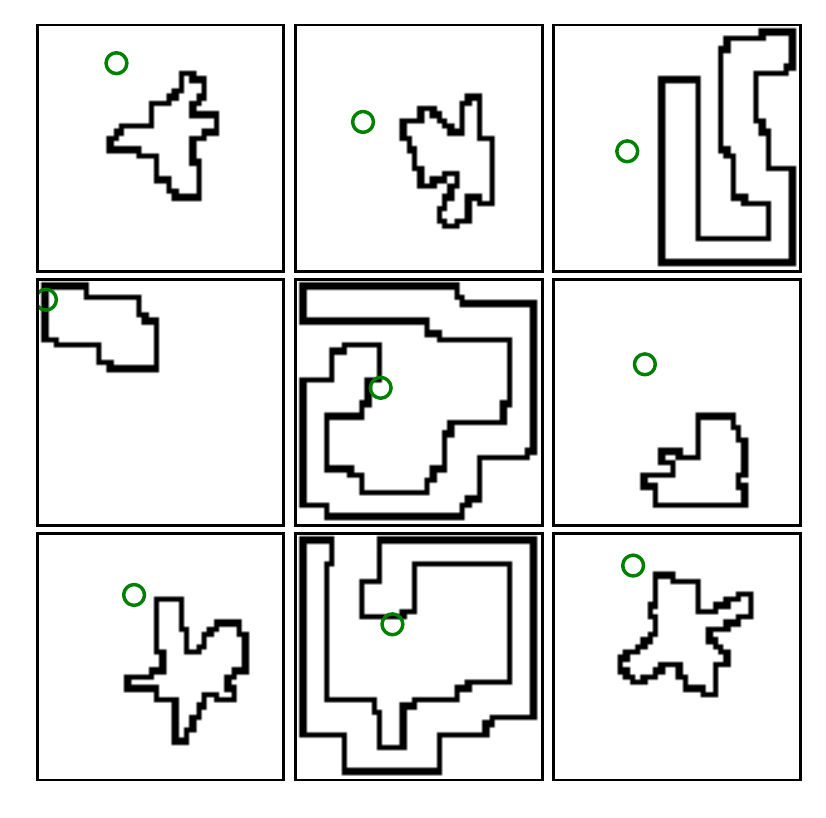}
  \end{tabular}
\caption{\emph{More examples of Visualization of the Units Learnt by~\emph{Dilation}.} The green dot indicates the location of the unit in the feature map.}
\label{visSuppDilation}
\end{figure*}

\renewcommand{\arraystretch}{0}

\begin{figure}[t]
  \centering
  \begin{tabular}{ll}
    \footnotesize
     \quad \quad {\bf t=0} \  \quad {\bf t=4} \ \ \ \ \ {\bf t=8} \quad  {\bf t=12} \quad {\bf t=16} \quad  {\bf t=20} &  
    \footnotesize
      \quad \quad {\bf t=0} \  \quad {\bf t=4} \ \ \ \ \ {\bf t=8} \quad  {\bf t=12} \quad {\bf t=16} \quad  {\bf t=20} \\
     \includegraphics[width=0.47\textwidth]{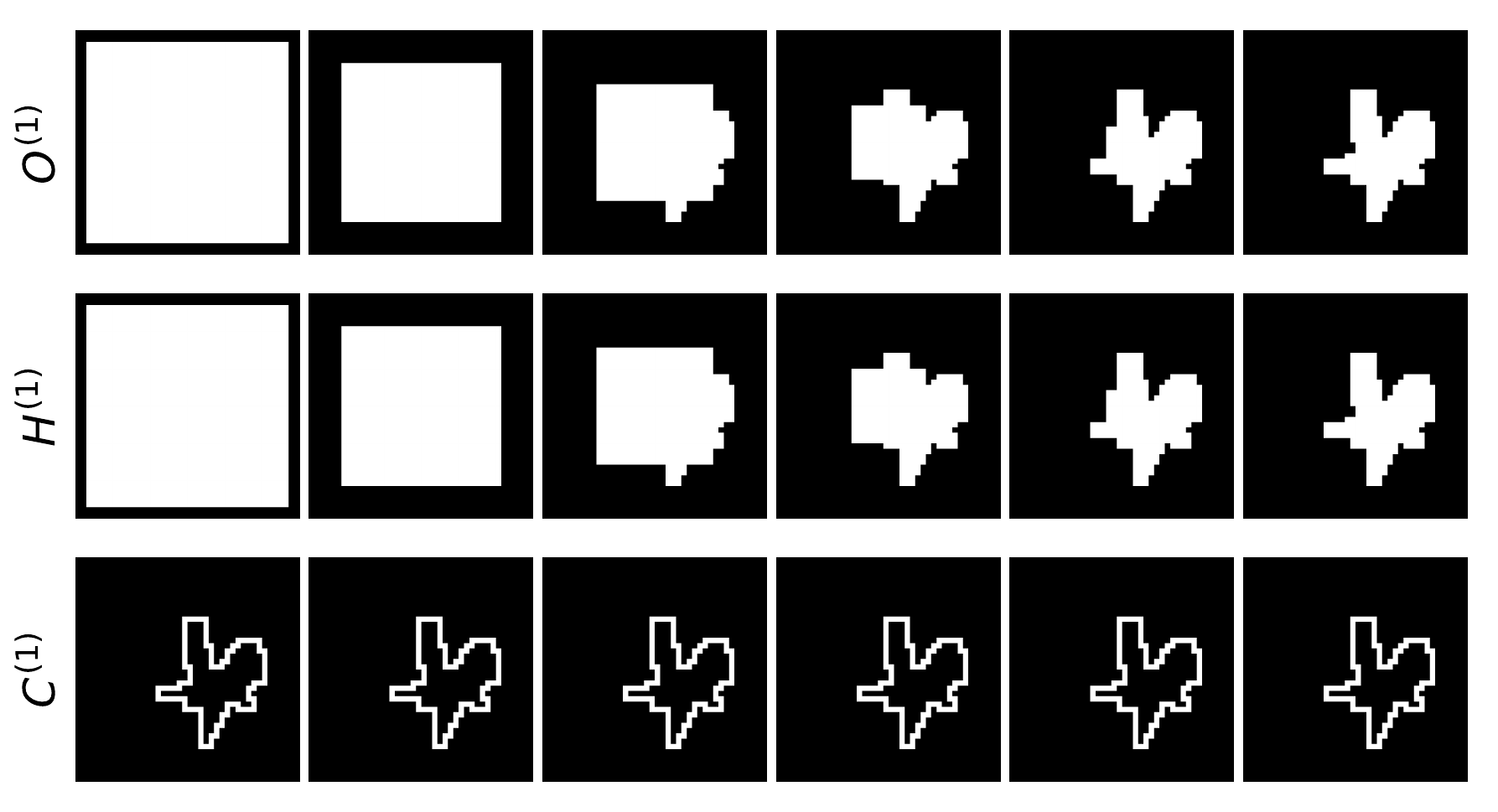} &
     \includegraphics[width=0.47\textwidth]{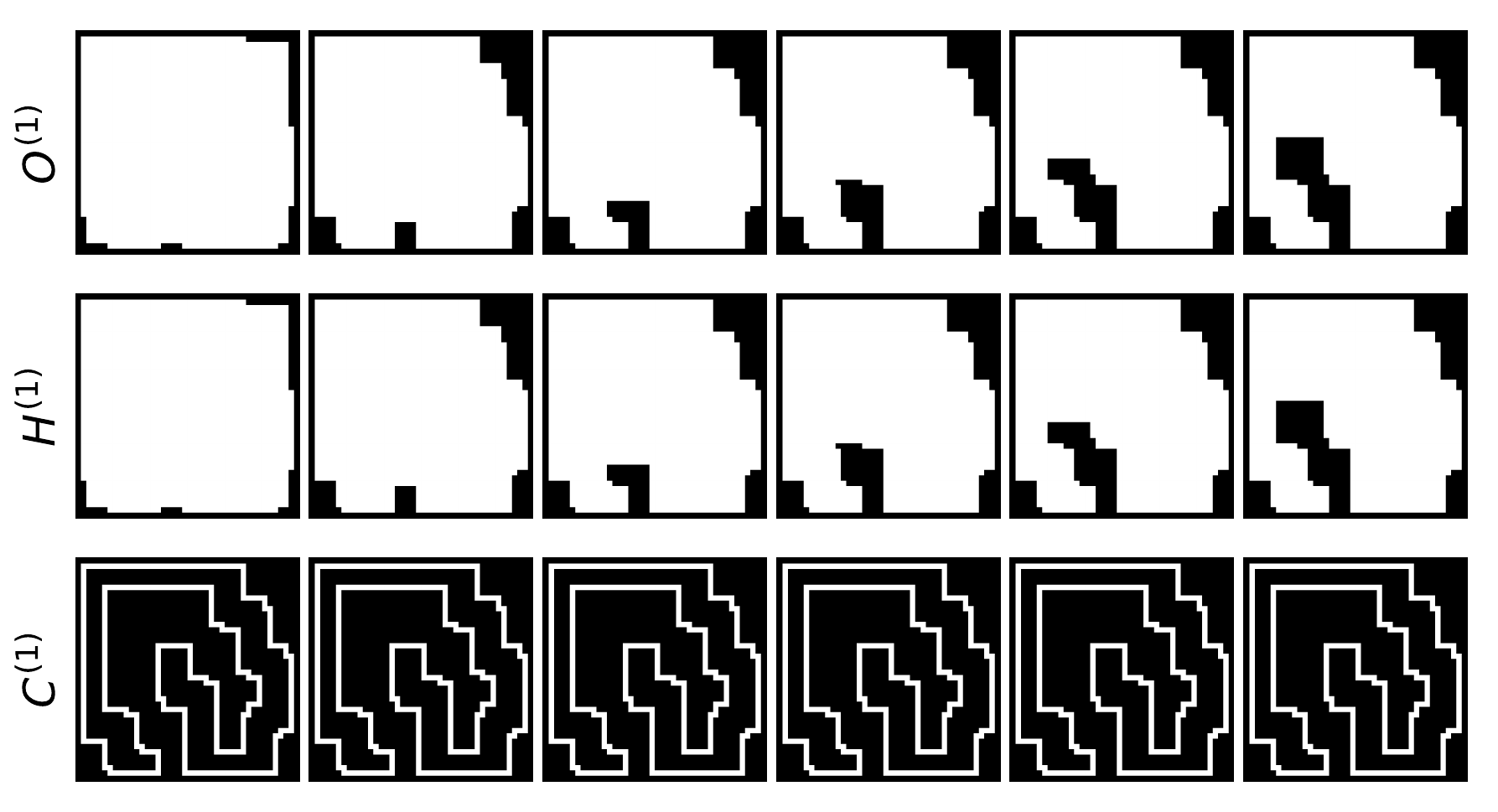} \\ 
  \end{tabular}
\caption{\emph{Visualization of Convolutional LSTM with the Mathematically Derived Parameters.}  We can see that only the border of the image (outside) is propagated, and not the curve, as in the learnt solution.}
\label{figMAthLSTM}
\end{figure}

\renewcommand{\arraystretch}{0}

\begin{figure}[t!]
\vspace*{-2cm}
  \centering
  \begin{tabular}{c}
    \scriptsize
         \ \ \ \ \ \ \quad {\bf t=0} 
     \ \ \ \ \ \quad {\bf t=5}
      \ \ \ \ \quad {\bf t=10}
      \ \ \ \quad {\bf t=15}
      \ \ \ \ \quad {\bf t=20}
      \ \ \ \ \quad {\bf t=25} 
      \ \ \ \ \quad {\bf t=30}
      \ \ \ \ \quad {\bf t=35}
      \ \ \ \ \quad {\bf t=40}
      \ \ \ \ \quad {\bf t=45}
      \ \ \ \quad {\bf t=50}
      \ \ \ \ \quad {\bf t=55}\\
     \includegraphics[width=0.99\textwidth]{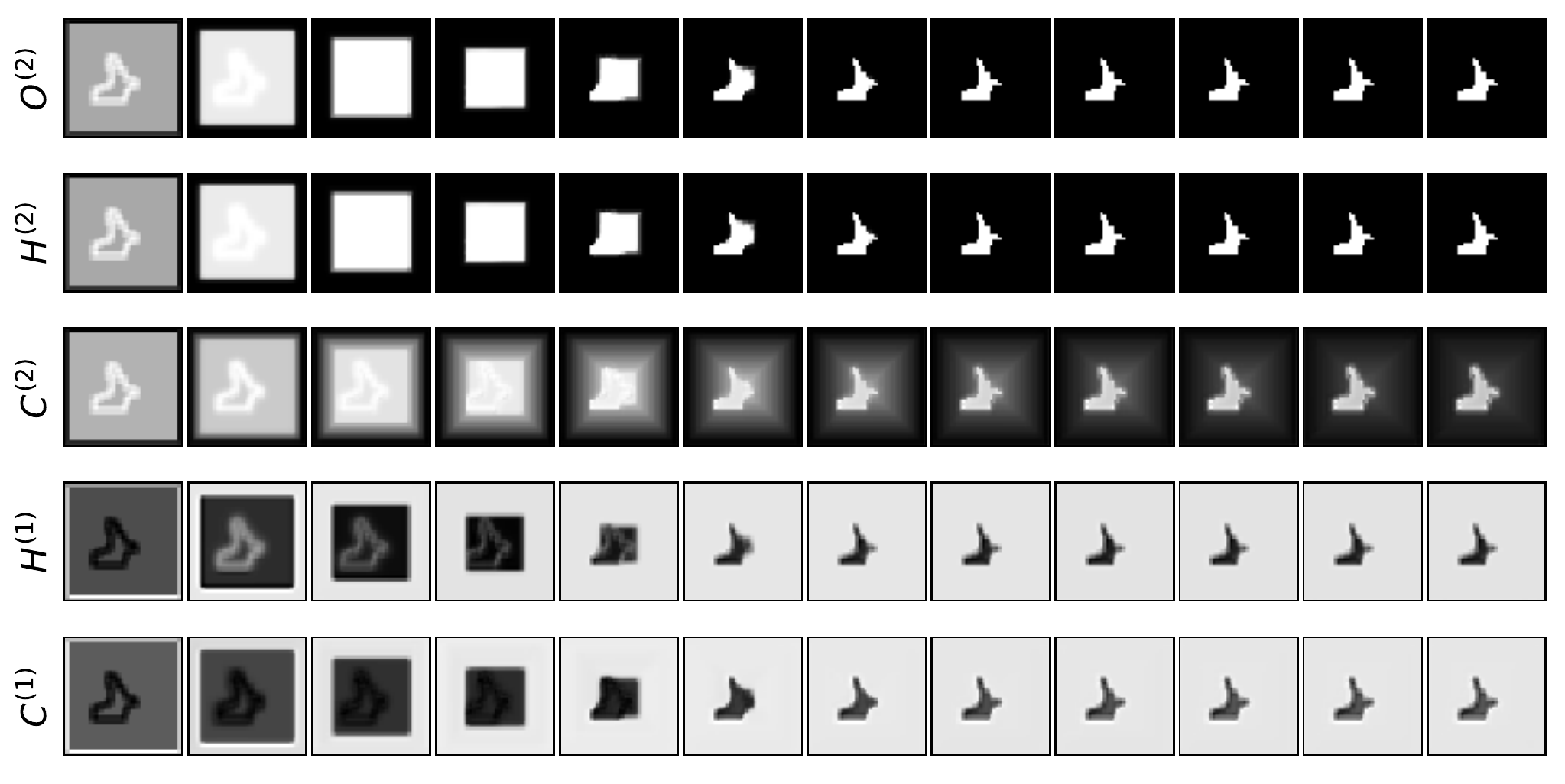} \\\\ 
     \scriptsize
     \ \ \ \ \ \ \quad {\bf t=0} 
     \ \ \ \ \ \quad {\bf t=5}
      \ \ \ \ \quad {\bf t=10}
      \ \ \ \quad {\bf t=15}
      \ \ \ \ \quad {\bf t=20}
      \ \ \ \ \quad {\bf t=25} 
      \ \ \ \ \quad {\bf t=30}
      \ \ \ \ \quad {\bf t=35}
      \ \ \ \ \quad {\bf t=40}
      \ \ \ \ \quad {\bf t=45}
      \ \ \ \quad {\bf t=50}
      \ \ \ \ \quad {\bf t=55}\\
     \includegraphics[width=0.99\textwidth]{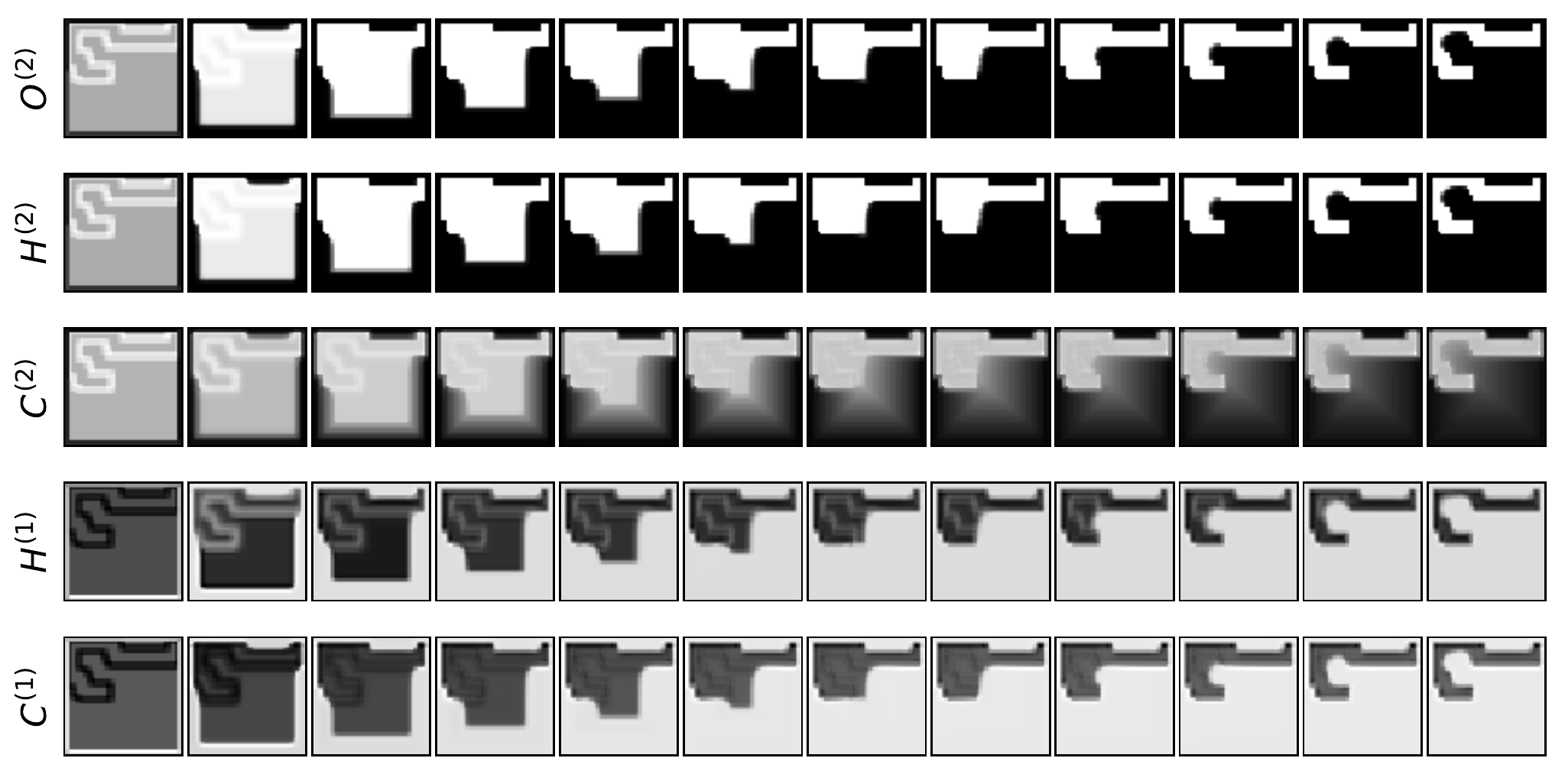}
      \\
  \end{tabular}
\caption{\emph{Activation Maps of the Learnt Representations by \emph{2-LSTM}, Trained on Small Images ($18\times 18$ pixels).}  The implementation of the coloring routine is different from the one theoretically derived in the paper, which is depicted in Fig.~\ref{figMAthLSTM}. The cell state has the same trend as the hidden state but in the derived solution the cell state is equal to $\neg \boldsymbol{X}$.  This demonstrates that there are different ways of implementing the coloring routine. Also, note that the first LSTM seems to already have solved the insideness problem. Recall that the 2-LSTM network has one more LSTM than necessary to implement the analytically-derived solution and hence, it is reasonable that the unnecessary LSTM may have no functional role. For a network with just one LSTM, we were unable to minimize the loss function. Thus, the second LSTM is helpful for the optimization as gradient descend in a network with more parameters may have more paths to minimize the loss. }
\label{visSuppLSTM_shaping}
\end{figure}

\clearpage

\section{Per-step Learning of the \emph{Coloring Routine}}
\label{sec:perstep}
\begin{figure}[t!]
\centering
    \includegraphics[width=.47\linewidth]{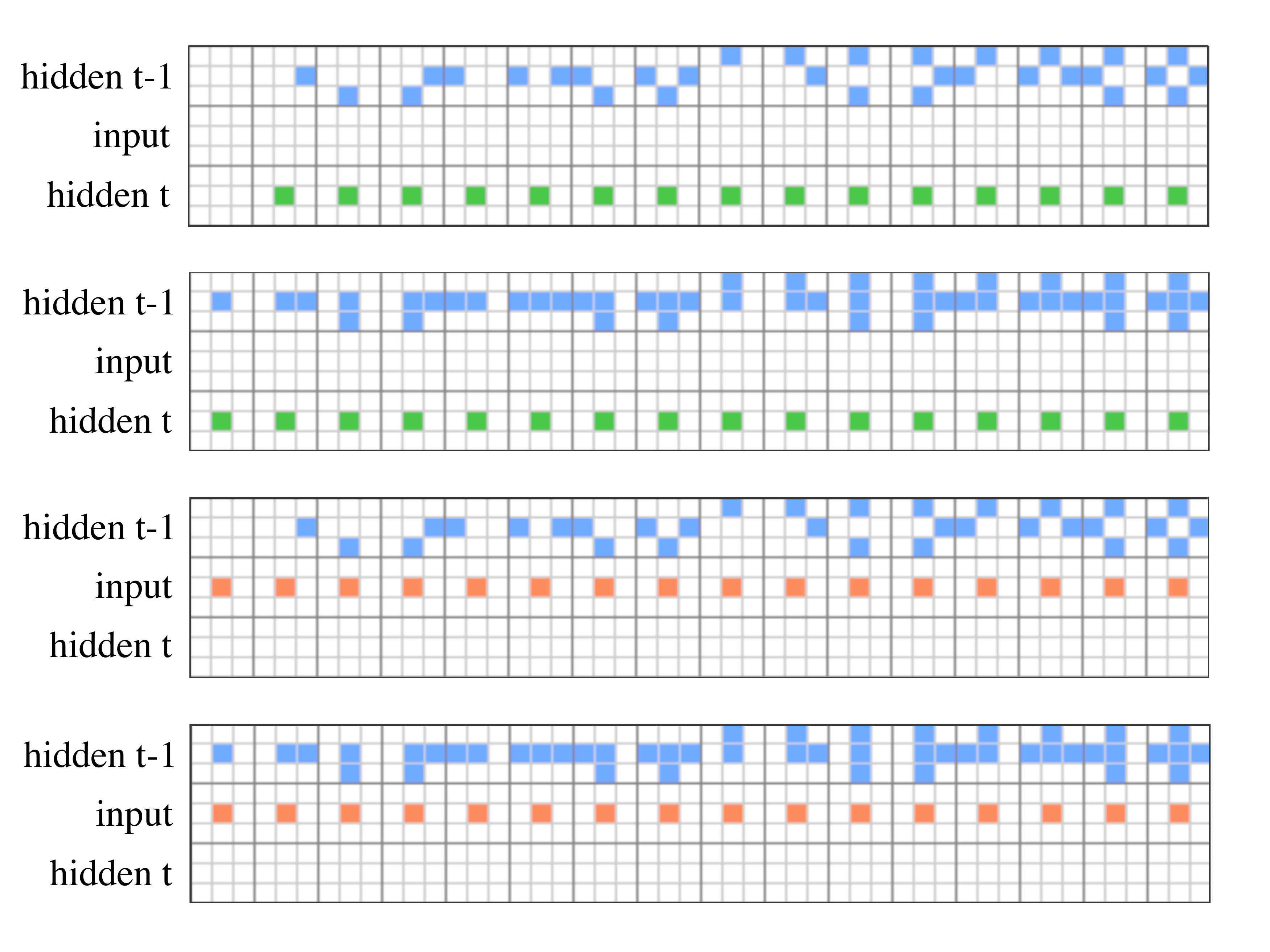}
\caption{\emph{Learning the Coloring Routine}.  $64$ possible inputs and outputs of the training set of the per-step training of the \emph{RNN} for the relevant inputs. }
\label{figRock}
\end{figure}

The \emph{Coloring Routine} can be learned by enforcing to each step the ground-truth produced by the routine, rather than waiting until the last step.  The inputs of a step are the image and the hidden state of the previous step. Recall that the \emph{Coloring Routine} determines that a pixel is outside if there is at least one neighbor assigned to outside that is not at the curve border.  All input cases (64) are depicted in Fig.~\ref{figRock}, leaving the irrelevant inputs for the \emph{Coloring Routine} at $0$.  During learning, such irrelevant pixels are assigned randomly a value of $0$ or $1$.

We  could  not  make  any  of the  previously  introduced LSTM networks  fit  a  step  of  the coloring  routine due  to  optimization problems. Yet, we found that the \emph{RNN} network was able to learn with the per-step training. The \emph{RNN} reached $0$ training error about $40\%$ of the times after randomly initializing the parameters.  After training the \emph{RNN} in one step, we unroll it and apply it to images of Jordan curves.  We can see that with less than $1000$ examples the \emph{RNN} is able to generalize to any of the datasets for more than $99\%$ of the images.  This demonstrates the great potential of decomposing the learning to facilitate the emergence of the routine.

\section{Networks for Natural Images}
\label{secSuppNatural}

We evaluate if training with the large-scale datasets in natural images solves the generalization issues that we reported before. 

We chose two state-of-the-art networks on Instance Segmentation, DEXTR~\citep{maninis2018deep} and DeepLabv3+~\citep{chen2018deeplabv3+},  to investigate their ability in solving the insideness problem. In the following, we show that these methods fail to determine the insideness for a vast majority of curves, even after fine-tuning in the~\emph{Both} dataset (DeepLabv3+ achieved  $36.58\%$ per image accuracy in \emph{Both} dataset and $2.18\%$ in~\emph{Digs}.

\begin{figure*}[t]
  \footnotesize
  \begin{tabular}{c@{\hspace{-0.01cm}}c@{\hspace{+0.2cm}}c@{\hspace{-0.01cm}}c@{\hspace{-0.01cm}}}
        \includegraphics[width=0.24\textwidth]{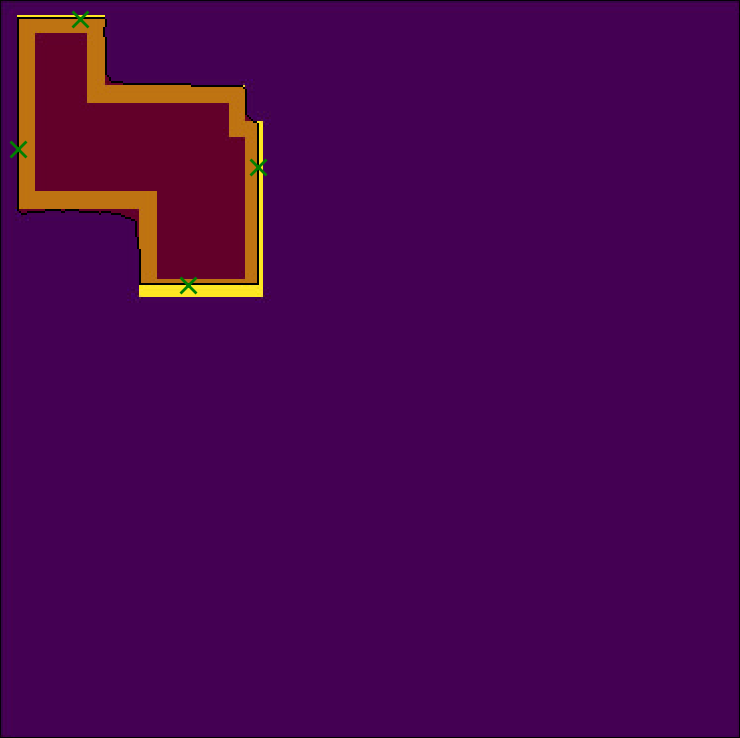}&
        \includegraphics[width=0.24\textwidth]{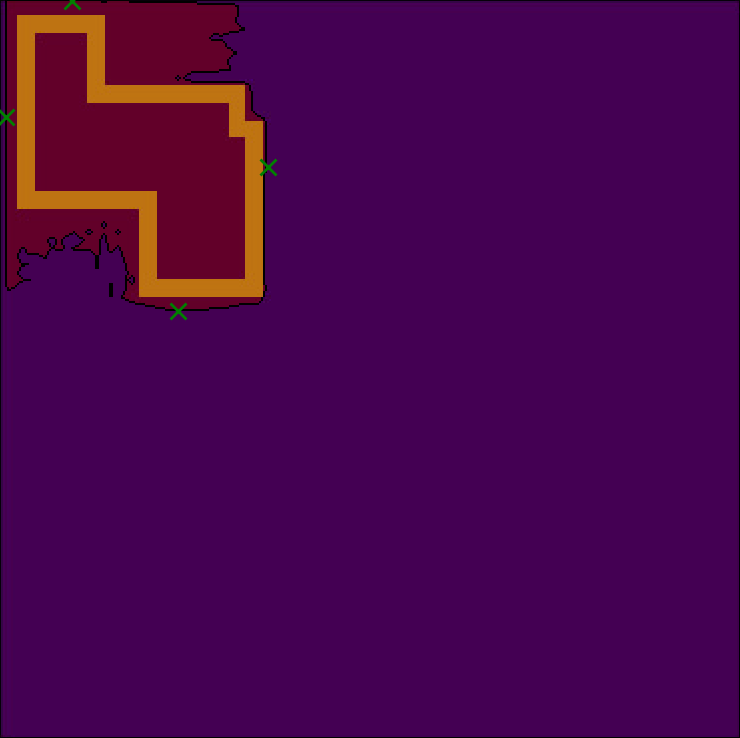} &\includegraphics[width=0.24\textwidth]{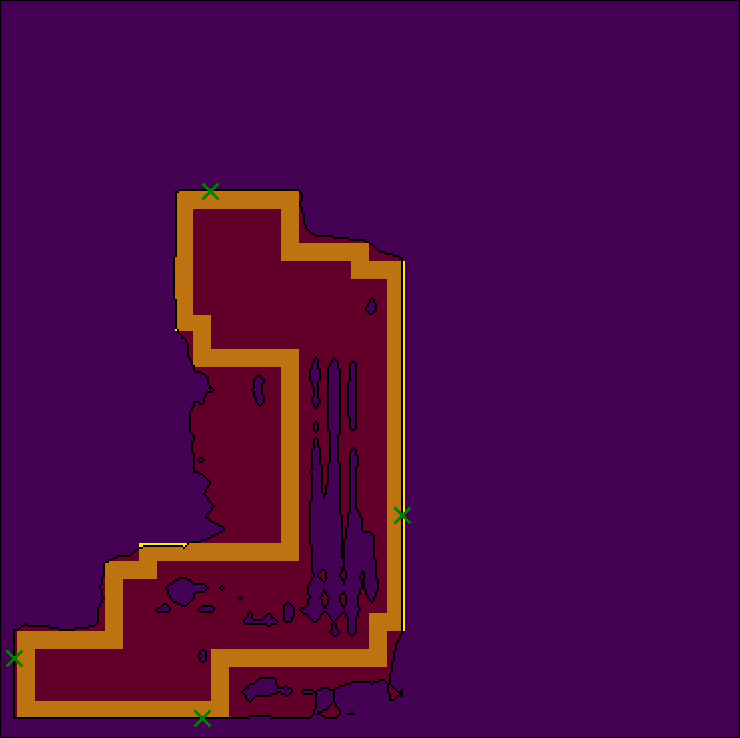}&
        \includegraphics[width=0.24\textwidth]{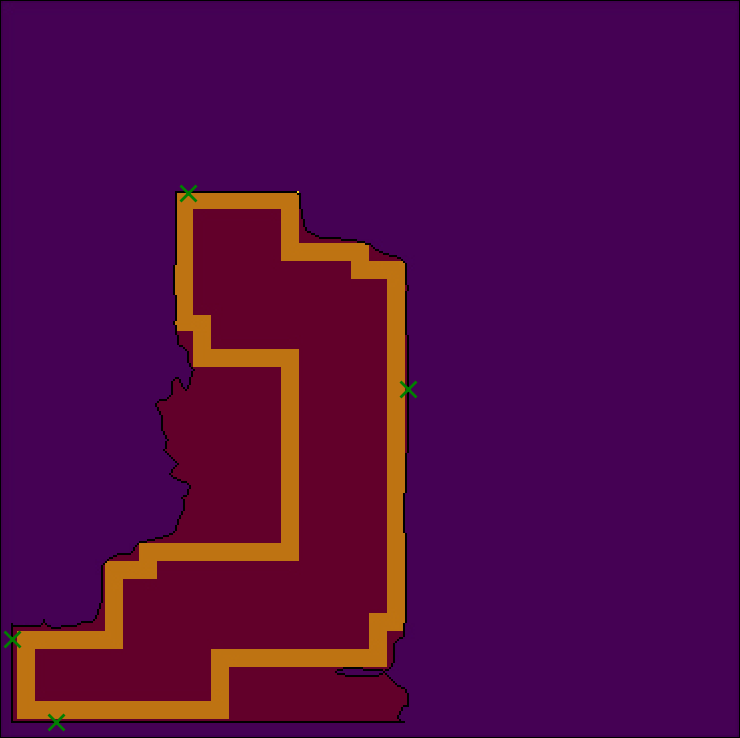} \\
        \includegraphics[width=0.24\textwidth]{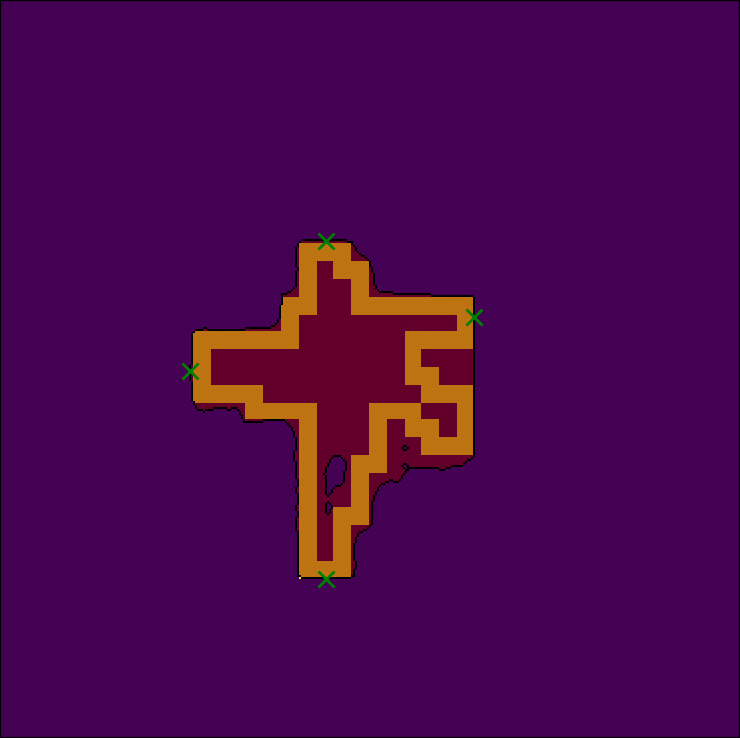}&
        \includegraphics[width=0.24\textwidth]{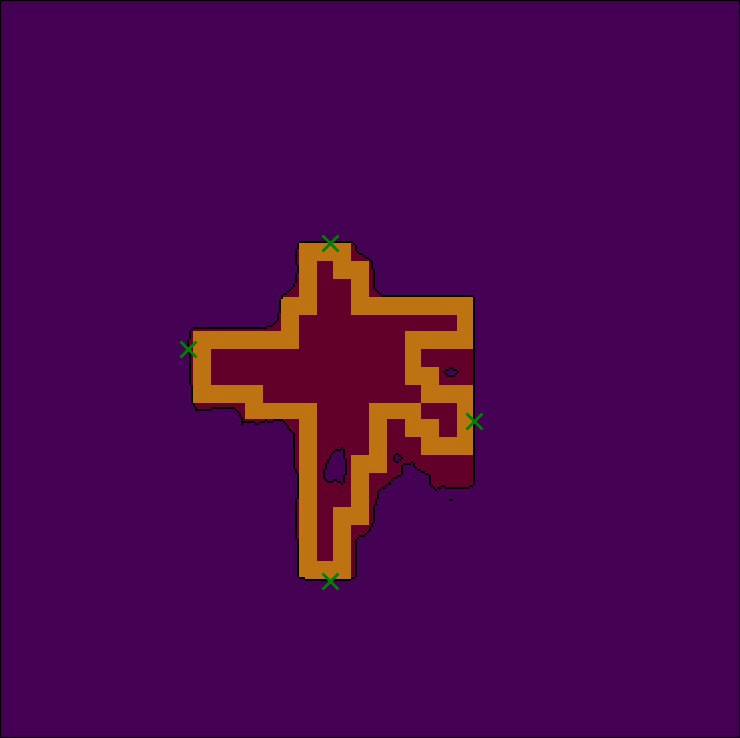} &\includegraphics[width=0.24\textwidth]{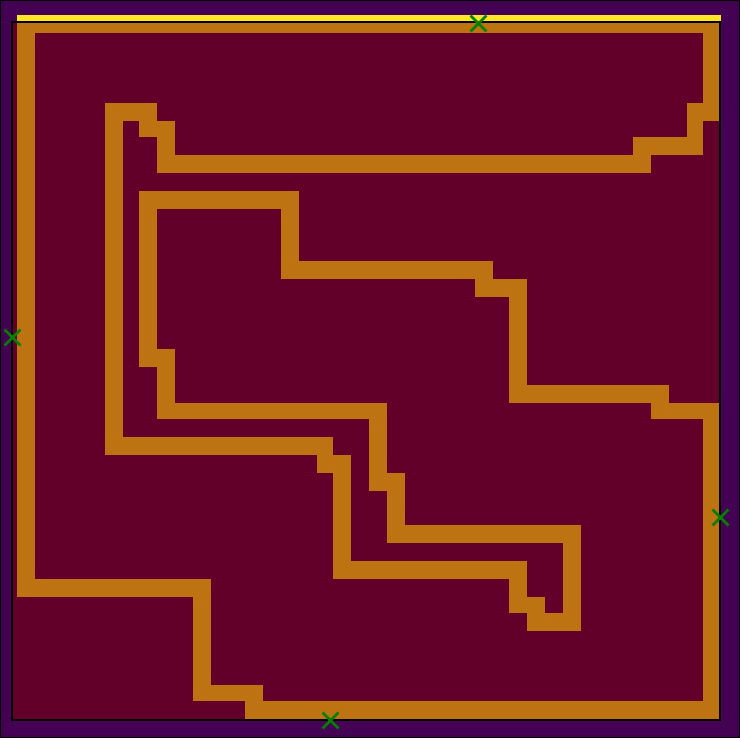}&
        \includegraphics[width=0.24\textwidth]{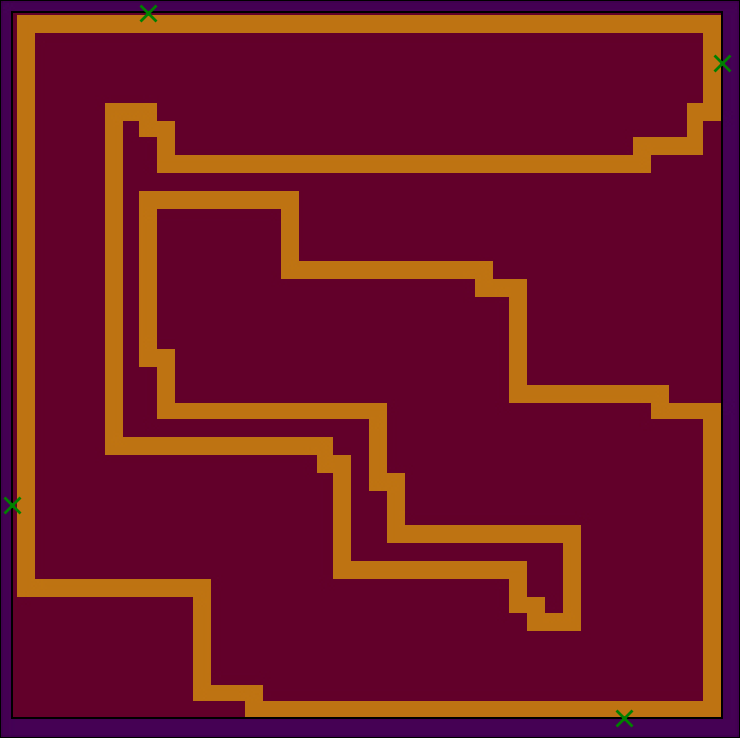} \\
         \includegraphics[width=0.24\textwidth]{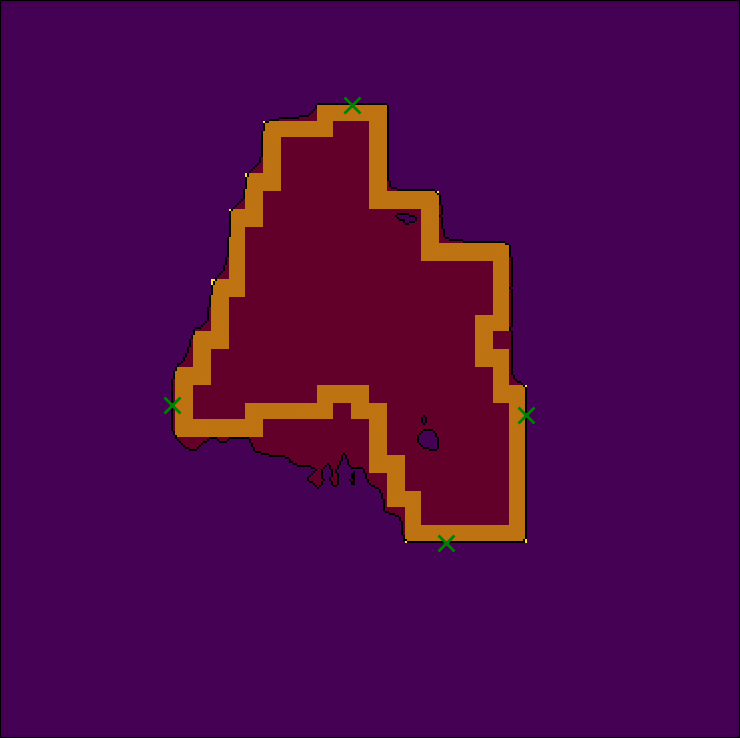}&
        \includegraphics[width=0.24\textwidth]{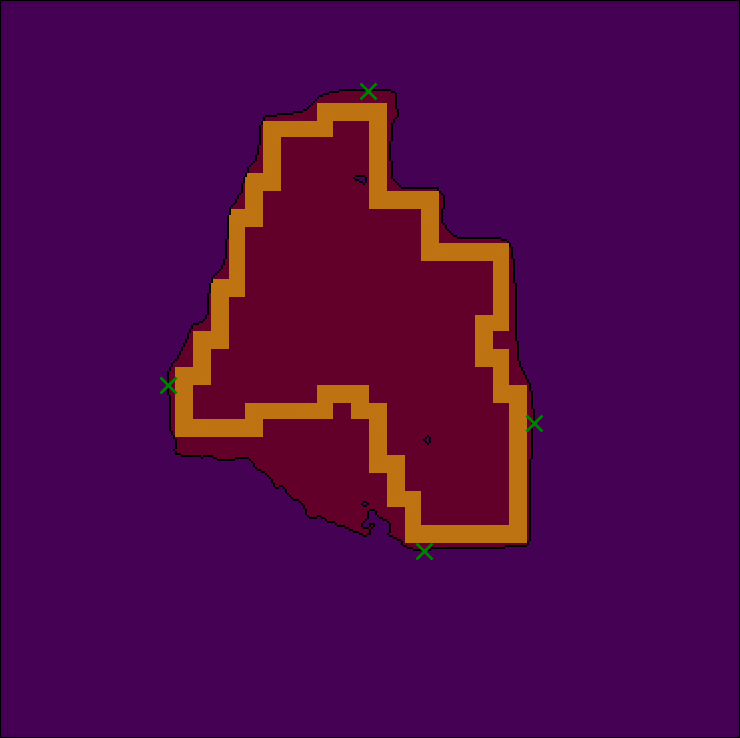} &\includegraphics[width=0.24\textwidth]{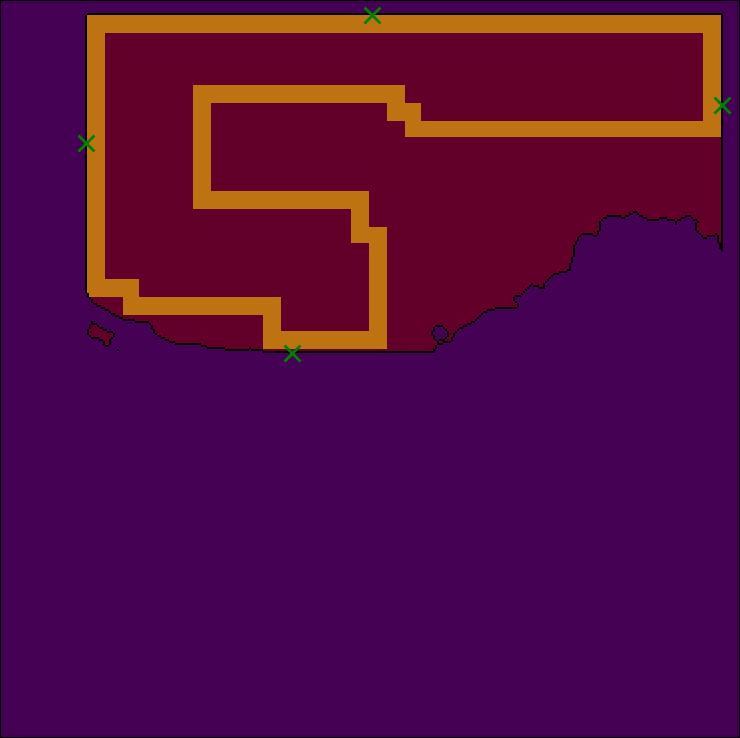}&
        \includegraphics[width=0.24\textwidth]{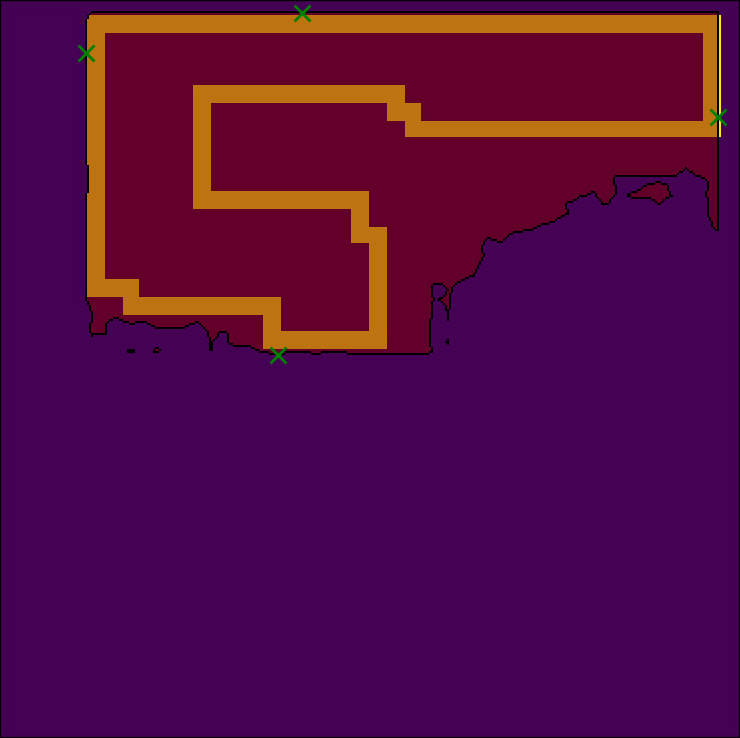} \\
         \includegraphics[width=0.24\textwidth]{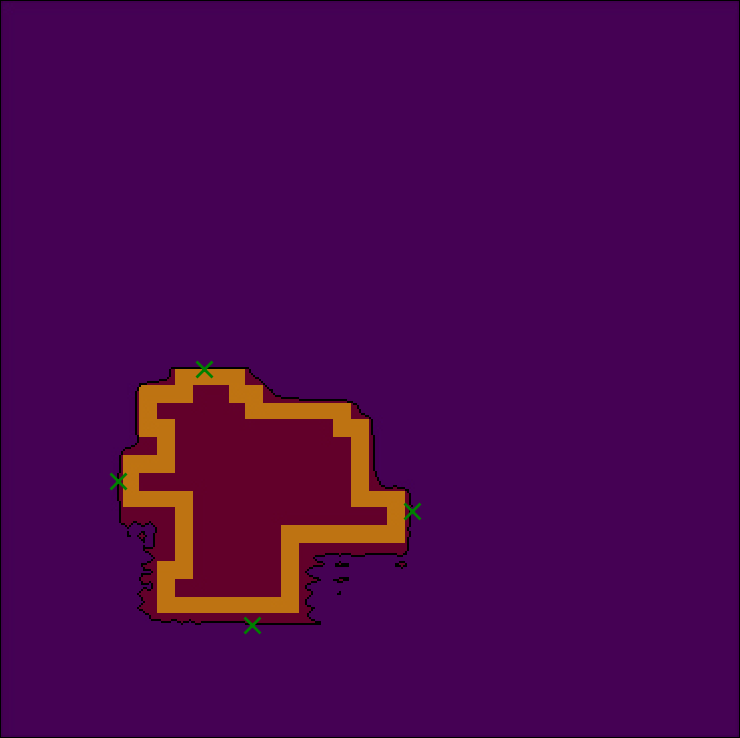}&
        \includegraphics[width=0.24\textwidth]{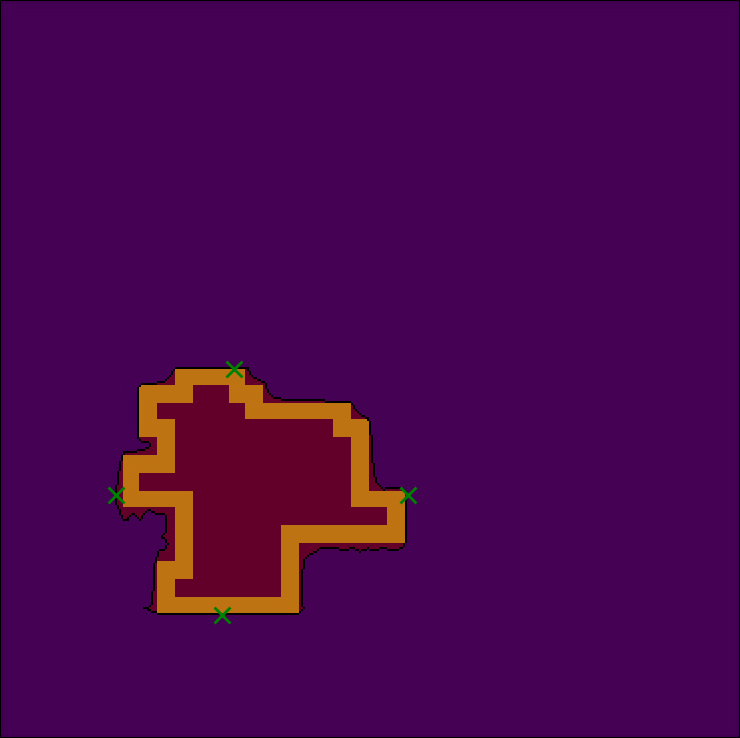} & \includegraphics[width=0.24\textwidth]{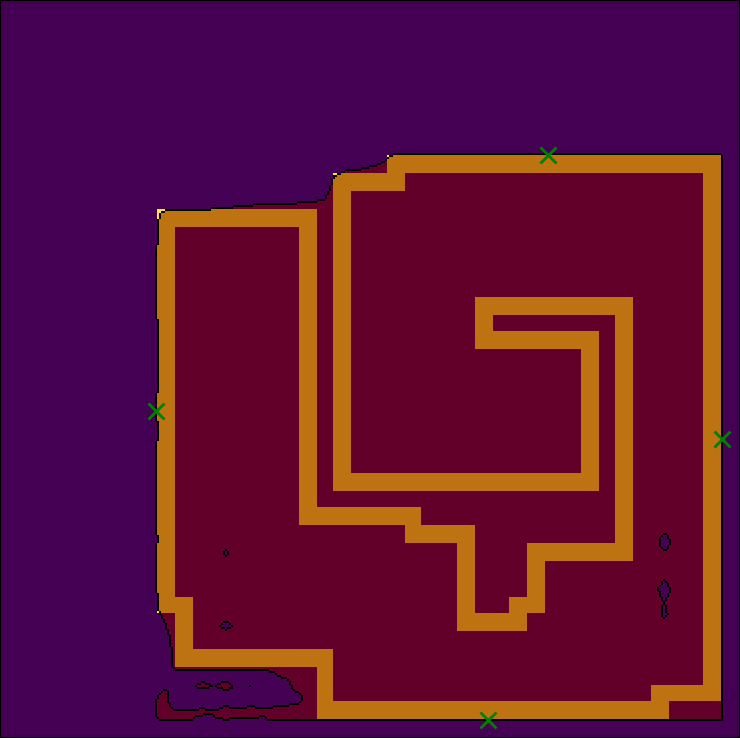}&
        \includegraphics[width=0.24\textwidth]{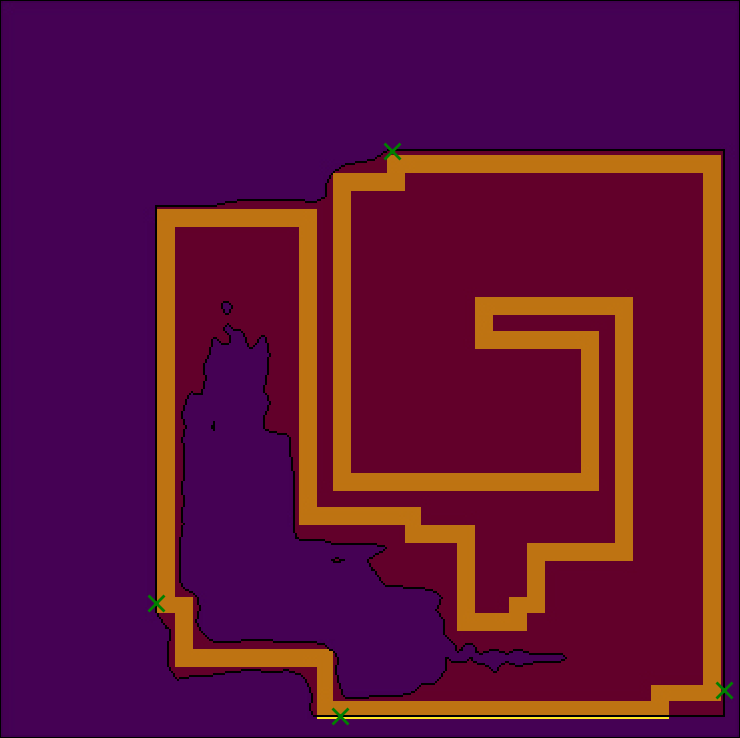} \\
         \includegraphics[width=0.24\textwidth]{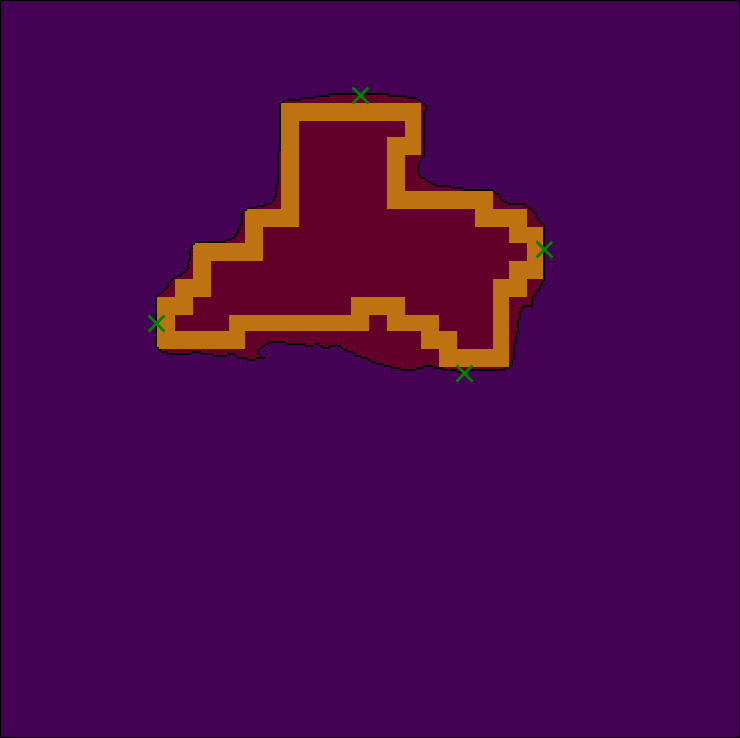}&
        \includegraphics[width=0.24\textwidth]{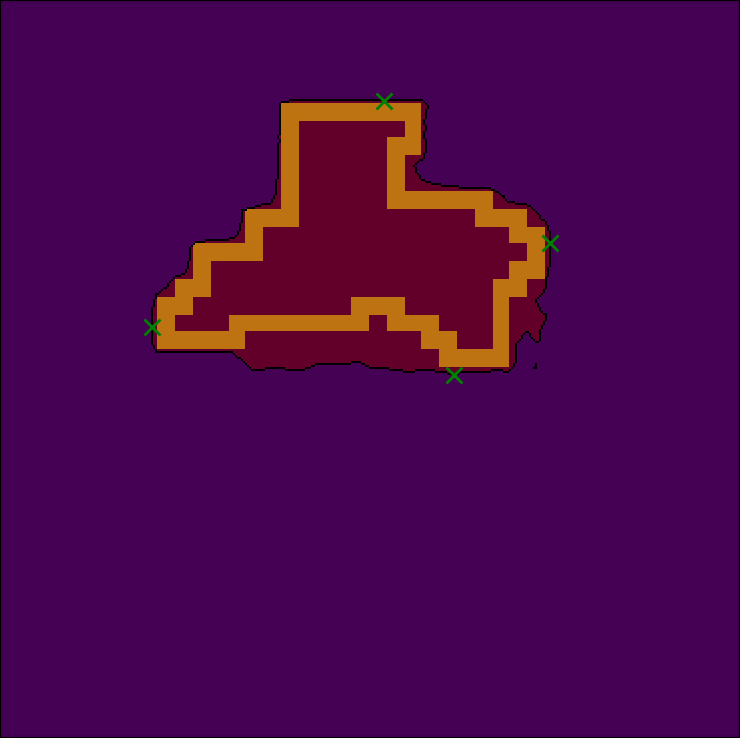} &\includegraphics[width=0.24\textwidth]{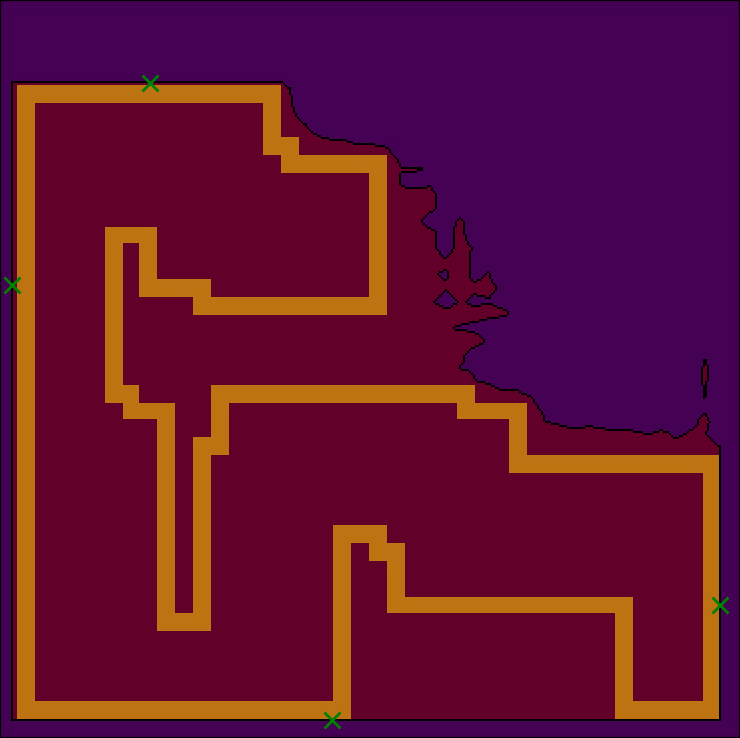}&
        \includegraphics[width=0.24\textwidth]{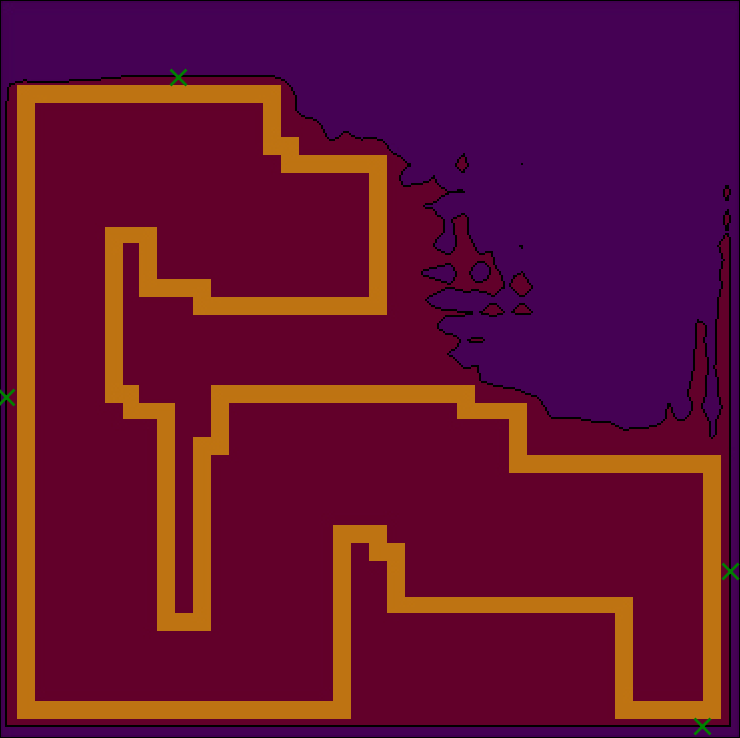} \\
  \end{tabular}
\caption{\emph{Qualitative Results with DEXTR on the Polar Dataset.}   We use the publicly available pre-trained DEXTR model~\citep{maninis2018deep}. DEXTR uses 4 points marked by the user (indicated with crosses). We report the best found points, two examples of them per image.}
\label{figDEXTR}
\end{figure*}

\textbf{DEXTR.} Deep Exteme Cut (DEXTR) is a neural network used for interactive instance segmentation. We use the pre-trained model on PASCAL VOC 2012 \citep{pascal-voc-2012} and show some of the qualitative results in Fig. \ref{figDEXTR}.

\textbf{DeepLabv3+.}
This architecture extends DeepLabv3~\citep{chen2018deeplab} by utilizing it as an encoder network and adding a decoder network to refine segmentation boundaries. The encoder employs dilated convolution and Atrous Spatial Pyramid Pooling module for feature extraction. We use DeepLabv3+ with Xception backbone pretrained on PASCAL VOC 2012, and fine-tune its last layer with Polar and Spiral datasets for training.
The ratio of input image spatial resolution to encoder output image is referred to as output stride and varies according to dilation rates. 
We use output strides of 8 and 16 as suggested in the paper; loss weight ($\alpha$) of 0.1, 0.2 and 0.4; and initial learning rates from 0.1 to $10^{-5}$ (dividing by $10$). We train the network on Polar and Spiral datasets until there is no improvement of the accuracy at the validations set, and we then reduce the learning rate by a ratio of $10$ and stop at the next plateau of the validation set accuracy. 

Finally, we also tried training this network from scratch and the per image accuracy never improved from $0\%$.

\begin{figure*}[t]
\setlength{\fboxsep}{0pt}
\setlength{\fboxrule}{0.05mm}

  \footnotesize
  \begin{tabular}{c@{\hspace{-0.01cm}}c@{\hspace{+0.1cm}}c@{\hspace{-0.01cm}}c@{\hspace{+0.1cm}}c@{\hspace{-0.01cm}}c@{\hspace{-0.01cm}}}
        \multicolumn{2}{c}{\bf Polar} & \multicolumn{2}{c}{\bf Spiral} & \multicolumn{2}{c}{\bf Digs} \\
        
        \fbox{\includegraphics[width=0.16\textwidth]{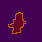}}&
        \fbox{\includegraphics[width=0.16\textwidth]{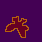}}&
        \fbox{\includegraphics[width=0.16\textwidth]{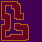}}&
        \fbox{\includegraphics[width=0.16\textwidth]{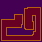}}&
        \fbox{\includegraphics[width=0.16\textwidth]{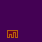}}&
        \fbox{\includegraphics[width=0.16\textwidth]{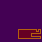}}\\

        \fbox{\includegraphics[width=0.16\textwidth]{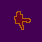}}&
        \fbox{\includegraphics[width=0.16\textwidth]{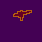}}&
        \fbox{\includegraphics[width=0.16\textwidth]{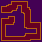}}&
        \fbox{\includegraphics[width=0.16\textwidth]{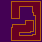}}& 
        \fbox{\includegraphics[width=0.16\textwidth]{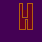}}&
        \fbox{\includegraphics[width=0.16\textwidth]{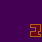}}\\
        
        \fbox{\includegraphics[width=0.16\textwidth]{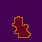}}&
        \fbox{\includegraphics[width=0.16\textwidth]{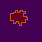}}&
        \fbox{\includegraphics[width=0.16\textwidth]{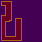}}&
        \fbox{\includegraphics[width=0.16\textwidth]{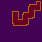}}&
        \fbox{\includegraphics[width=0.16\textwidth]{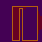}}&
        \fbox{\includegraphics[width=0.16\textwidth]{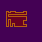}}\\
        
        \fbox{\includegraphics[width=0.16\textwidth]{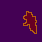}}&
        \fbox{\includegraphics[width=0.16\textwidth]{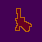}}&
        \fbox{\includegraphics[width=0.16\textwidth]{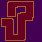}}&
        \fbox{\includegraphics[width=0.16\textwidth]{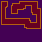}}&
        \fbox{\includegraphics[width=0.16\textwidth]{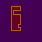}}&
        \fbox{\includegraphics[width=0.16\textwidth]{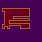}} \\
        
        \fbox{\includegraphics[width=0.16\textwidth]{./fig/deeplabv3/polar/000014_blend.png}}&
        \fbox{\includegraphics[width=0.16\textwidth]{./fig/deeplabv3/polar/000015_blend.png}}&
        \fbox{\includegraphics[width=0.16\textwidth]{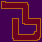}}&
        \fbox{\includegraphics[width=0.16\textwidth]{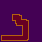}}&
        \fbox{\includegraphics[width=0.16\textwidth]{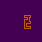}}&
        \fbox{\includegraphics[width=0.16\textwidth]{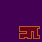}} \\
        
        \fbox{\includegraphics[width=0.16\textwidth]{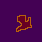}}&
        \fbox{\includegraphics[width=0.16\textwidth]{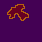}}&
        \fbox{\includegraphics[width=0.16\textwidth]{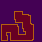}}&
        \fbox{\includegraphics[width=0.16\textwidth]{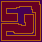}}&
        \fbox{\includegraphics[width=0.16\textwidth]{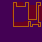}}&
        \fbox{\includegraphics[width=0.16\textwidth]{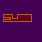}} \\
        
  \end{tabular}
\caption{\emph{Results of DeepLabv3+ on Polar, Spiral and Digs Datasets.} The network is fine-tuned on Polar and Spiral. The results show that the network predicts well most of the pixels except in the borders. For the cross-dataset evaluations in the Digs dataset, the network is not able to generalize. }
\label{figdeeplabv3+}
\end{figure*}

\clearpage
\printbibliography

@article{jeurissen2016serial,
  title={Serial grouping of {2D}-image regions with object-based attention in humans},
  author={Jeurissen, Danique and Self, Matthew W and Roelfsema, Pieter R},
  journal={Elife},
  volume={5},
  pages={e14320},
  year={2016},
  publisher={eLife Sciences Publications Limited}
}

@inproceedings{zeiler2014visualizing,
  title={Visualizing and understanding convolutional networks},
  author={Zeiler, Matthew D and Fergus, Rob},
  booktitle={Proceedings of the 13th {European Conference on Computer Vision (ECCV 2014)}},
  year={2014}
}

@article{cybenko1989approximation,
  title={Approximation by superpositions of a sigmoidal function},
  author={Cybenko, George},
  journal={Mathematics of Control, Signals and Systems},
  volume={2},
  number={4},
  pages={303--314},
  year={1989},
  publisher={Springer}
}

@article{soudry2018implicit,
  title={The implicit bias of gradient descent on separable data},
  author={Soudry, Daniel and Hoffer, Elad and Nacson, Mor Shpigel and Gunasekar, Suriya and Srebro, Nathan},
  journal={Journal of Machine Learning Research (JMLR)},
  volume={19},
  number={1},
  pages={2822--2878},
  year={2018}
}

@inproceedings{chen2018deeplabv3+,
  author    = {Liang{-}Chieh Chen and
               Yukun Zhu and
               George Papandreou and
               Florian Schroff and
               Hartwig Adam},
  title     = {Encoder-Decoder with Atrous Separable Convolution for Semantic Image
               Segmentation},
  booktitle={Proceedings of the 15th {European Conference on Computer Vision (ECCV 2018)}},
  pages={801--818},
  year      = {2018}
}

@article{chen2018deeplab,
  title={{DeepLab}: Semantic image segmentation with deep convolutional nets, atrous convolution, and fully connected {CRF}s},
  author={Chen, Liang-Chieh and Papandreou, George and Kokkinos, Iasonas and Murphy, Kevin and Yuille, Alan L},
  journal={IEEE Transactions on Pattern Analysis and Machine Intelligence (TPAMI)},
  volume={40},
  number={4},
  pages={834-848},
  year={2018}
}

@inproceedings{YuKoltun2016,
	author    = {Fisher Yu and Vladlen Koltun},
	title     = {Multi-Scale Context Aggregation by Dilated Convolutions},
  booktitle={Proceedings of the 4th {International Conference on Learning Representations (ICLR 2016)}},
	year      = {2016},
}

@inproceedings{RFB15,
  title={{U-Net}: Convolutional networks for biomedical image segmentation},
  author={Ronneberger, Olaf and Fischer, Philipp and Brox, Thomas},
  pages={234--241},
  booktitle={Proceedings of the 18th { International Conference on Medical Image Computing and Computer Assisted Intervention (MICCAI 2015)}},
  year={2015}, 
}

@article{BKC17,
  title={{SegNet}: A deep convolutional encoder-decoder architecture for image segmentation},
  author={Badrinarayanan, Vijay and Kendall, Alex and Cipolla, Roberto},
  journal={IEEE Transactions on Pattern Analysis and Machine Intelligence (TPAMI)},
  volume={39},
  number={12},
  pages={2481--2495},
  year={2017},
}

@inproceedings{long2015fully,
  title={Fully convolutional networks for semantic segmentation},
  author={Long, Jonathan and Shelhamer, Evan and Darrell, Trevor},
  booktitle={Proceedings of the 28th {IEEE/CVF Conference on Computer Vision and Pattern Recognition (CVPR 2015)}},
  pages={3431--3440},
  year={2015}
}

@article{Ullman84,
  title={Visual routines},
  author={Ullman, Shimon},
  journal={Cognition},
  volume={18},
  pages={97--159},  
  year={1984}
}

@book{Ullman96,
  title={High-Level Vision: Object Recognition and Visual Cognition},
  author={Ullman, Shimon},
  edition={1st},
  year={1996},
  publisher={MIT Press}
}

@book{MP69,
  title={Perceptrons: An Introduction to Computational Geometry},
  author={Minsky, Marvin L. and Papert, Seymour A.},
  edition={1st},
  year={1969},
  publisher={MIT Press}
}

@inproceedings{wu2018cognitive,
  title={Cognitive Deficit of Deep Learning in Numerosity},
  author={Wu, Xiaolin and Zhang, Xi and Shu, Xiao},
  booktitle={Proceedings of the 33rd {AAAI Conference on Artificial Intelligence (AAAI-19)}},
  year={2019}
}

@inproceedings{SOS17,
  title={Failures of gradient-based deep learning},
  author={Shalev-Shwartz, Shai and Shamir, Ohad and Shammah, Shaked},
  booktitle={Proceedings of the 34th {International Conference on Machine Learning (ICML 2017)}},
  pages={3067--3075},
  year={2017}
}

@incollection{Haines94,
 author = {Haines, Eric},
 title  = {Point in Polygon Strategies},
 booktitle = {Graphics Gems {IV}},
 editor = {Paul Heckbert},
 year = {1994},
 pages = {24--46},
 publisher = {Academic Press}
}

@article{HS97,
  title={Long short-term memory},
  author={Hochreiter, Sepp and Schmidhuber, J{\"u}rgen},
  journal={Neural Computation},
  volume={9},
  number={8},
  pages={1735--1780},
  year={1997},
  publisher={MIT Press}
}

@inproceedings{
KimDisentangling,
title={Disentangling neural mechanisms for perceptual grouping},
author={Junkyung Kim and Drew Linsley and Kalpit Thakkar and Thomas Serre},
booktitle={Proceedings of the 8th {International Conference on Learning Representations (ICLR 2020)}},
year={2020}}

@inproceedings{pascanu2013difficulty,
  title={On the difficulty of training recurrent neural networks},
  author={Pascanu, Razvan and Mikolov, Tomas and Bengio, Yoshua},
  booktitle={Proceedings of the 30th {International Conference on Machine Learning (ICML 2013)}},
  pages={1310--1318},
  year={2013}
}

@inproceedings{gruslys2016memory,
  title={Memory-efficient backpropagation through time},
  author={Gruslys, Audrunas and Munos, R{\'e}mi and Danihelka, Ivo and Lanctot, Marc and Graves, Alex},
  booktitle={{Advances in Neural Information Processing Systems 29 (NIPS 2016)}},
  pages={4125--4133},
  year={2016}
}

@article{bengio1994learning,
  title={Learning long-term dependencies with gradient descent is difficult},
  author={Bengio, Yoshua and Simard, Patrice and Frasconi, Paolo},
  journal={IEEE Transactions on Neural Networks},
  volume={5},
  number={2},
  pages={157-166},
  year={1994}
}

@incollection{Kong01,
author="Kong, T. Yung",
editor="Davis, Larry S.",
title="Digital Topology",
booktitle="Foundations of Image Understanding",
year="2001",
pages="73--93",
publisher="Springer",
}

@inproceedings{liu2018intriguing,
  title={An intriguing failing of convolutional neural networks and the {CoordConv} solution},
  author={Liu, Rosanne and Lehman, Joel and Molino, Piero and Such, Felipe Petroski and Frank, Eric and Sergeev, Alex and Yosinski, Jason},
  booktitle={{Advances in Neural Information Processing Systems 31 (NeurIPS 2018)}},
  pages={9605--9616},
  year={2018}
}

@inproceedings{li2018referring,
  title={Referring image segmentation via recurrent refinement networks},
  author={Li, Ruiyu and Li, Kaican and Kuo, Yi-Chun and Shu, Michelle and Qi, Xiaojuan and Shen, Xiaoyong and Jia, Jiaya},
  booktitle={Proceedings of the 31st {IEEE/CVF Conference on Computer Vision and Pattern Recognition (CVPR 2018)}},
  pages={5745--5753},
  year={2018}
}

@article{Rosenfeld70,
 author = {Rosenfeld, Azriel},
 title = {Connectivity in Digital Pictures},
 journal = {Journal of the ACM (JACM)},
 publisher = {ACM},
 address = {New York, NY, USA},
 volume = {17},
 number = {1},
 pages = {146–160},
 year = {1970},
}

@book{harary1969graph,
  title={Graph Theory},
  author={Harary, Frank},
  publisher={Addison-Wesley},
  year={1969}
}

@inproceedings{visin2016reseg,
  title={{ReSeg}: A recurrent neural network-based model for semantic segmentation},
  author={Visin, Francesco and Ciccone, Marco and Romero, Adriana and Kastner, Kyle and Cho, Kyunghyun and Bengio, Yoshua and Matteucci, Matteo and Courville, Aaron},
  booktitle={Proceedings of the 29th {IEEE/CVF Conference on Computer Vision and Pattern Recognition Workshops (CVPRW 2016)}},
  year={2016}
}

@article{LATEEF2019321,
author = "Fahad Lateef and Yassine Ruichek",
title = "Survey on semantic segmentation using deep learning techniques",
journal = "Neurocomputing",
volume = "338",
pages = "321 - 348",
year = "2019",
}

@misc{A140517,
  title = {{A140517}: {N}umber of cycles in an n X n grid, {In} \textit{The On-Line Encyclopedia of Integer Sequences}},
  note={\url{https://oeis.org/A140517}},
}

@misc{A140517-Table,
  author = {Karavaev, Artem M. and Iwashita, Hiroaki},
  title = "Table of n, a(n) for n = 0..26, {In} \textit{The On-Line Encyclopedia of Integer Sequences}",
  note = {\url{https://oeis.org/A140517/b140517.txt}},
}

@inproceedings{Iwashita+13b,
  title={Fast computation of the number of paths in a grid graph},
  author={Iwashita, Hiroaki and  Nakazawa, Yoshio and Kawahara, Jun and Uno, Takeaki and Minato, Shinichi},
  booktitle={Proceedings of the {16th Japan Conference on Discrete and Computational Geometry and Graphs (JCDCG2 2013)}},
  year={2013},
  month= sep,
  address= {Tokyo},
}

@inproceedings{xingjian2015convolutional,
  title={Convolutional {LSTM} network: A machine learning approach for precipitation nowcasting},
  author={Xingjian, SHI and Chen, Zhourong and Wang, Hao and Yeung, Dit-Yan and Wong, Wai-Kin and Woo, Wang-chun},
  booktitle={{Advances in Neural Information Processing Systems 28 (NIPS 2015)}},
  year={2015}
}

@inproceedings{linsley2018learning,
  title={Learning long-range spatial dependencies with horizontal gated recurrent units},
  author={Linsley, Drew and Kim, Junkyung and Veerabadran, Vijay and Windolf, Charles and Serre, Thomas},
  booktitle={{Advances in Neural Information Processing Systems 31 (NeurIPS 2018)}},
  pages={152--164},
  year={2018}
}

@inproceedings{song2018seednet,
  title={{SeedNet}: Automatic seed generation with deep reinforcement learning for robust interactive segmentation},
  author={Song, Gwangmo and Myeong, Heesoo and Mu Lee, Kyoung},
  booktitle={Proceedings of the 31st {IEEE/CVF Conference on Computer Vision and Pattern Recognition (CVPR 2018)}},
  year={2018}
}

@inproceedings{chen2018masklab,
  title={{MaskLab}: Instance segmentation by refining object detection with semantic and direction features},
  author={Chen, Liang-Chieh and Hermans, Alexander and Papandreou, George and Schroff, Florian and Wang, Peng and Adam, Hartwig},
  booktitle={Proceedings of the  31st {IEEE/CVF Conference on Computer Vision and Pattern Recognition (CVPR 2018)}},
  pages={4013--4022},
  year={2018}
}

@inproceedings{maninis2018deep,
  title={Deep extreme cut: From extreme points to object segmentation},
  author={Maninis, Kevis-Kokitsi and Caelles, Sergi and Pont-Tuset, Jordi and Van Gool, Luc},
  booktitle={Proceedings of the 31st {IEEE/CVF Conference on Computer Vision and Pattern Recognition (CVPR 2018)}},
  pages={616--625},
  year={2018}
}

@inproceedings{li2016iterative,
  title={Iterative instance segmentation},
  author={Li, Ke and Hariharan, Bharath and Malik, Jitendra},
  booktitle={Proceedings of the 29th {IEEE/CVF Conference on Computer Vision and Pattern Recognition (CVPR 2016)}},
  pages={3659--3667},
  year={2016}
}

@inproceedings{li2017fully,
  title={Fully convolutional instance-aware semantic segmentation},
  author={Li, Yi and Qi, Haozhi and Dai, Jifeng and Ji, Xiangyang and Wei, Yichen},
  booktitle={Proceedings of the 30th {IEEE/CVF Conference on Computer Vision and Pattern Recognition (CVPR 2017)}},
  pages={2359--2367},
  year={2017}
}

@inproceedings{hu2018learning,
  title={Learning to segment every thing},
  author={Hu, Ronghang and Doll{\'a}r, Piotr and He, Kaiming and Darrell, Trevor and Girshick, Ross},
  booktitle={Proceedings of the  31st {IEEE/CVF Conference on Computer Vision and Pattern Recognition (CVPR 2018)}},
  pages={4233--4241},
  year={2018}
}

@inproceedings{liu2018path,
  title={Path aggregation network for instance segmentation},
  author={Liu, Shu and Qi, Lu and Qin, Haifang and Shi, Jianping and Jia, Jiaya},
  booktitle={Proceedings of the 31st {IEEE/CVF Conference on Computer Vision and Pattern Recognition (CVPR 2018)}},
  pages={8759--8768},
  year={2018}
}

@inproceedings{he2017mask,
  title={Mask {R-CNN}},
  author={He, Kaiming and Gkioxari, Georgia and Doll{\'a}r, Piotr and Girshick, Ross},
  booktitle={Proceedings of the 16th {IEEE International Conference on Computer Vision (ICCV 2017)}},
  pages={2961--2969},
  year={2017}
}

@misc{pascal-voc-2012,
	author = {Everingham, Mark  and van Gool, Luc   and Williams, Chris  and Winn, John   and Zisserman, Andrew},
	title = {The {PASCAL} {V}isual {O}bject {C}lasses {C}hallenge 2012 {(VOC2012)} {R}esults},
	note = {\url{http://host.robots.ox.ac.uk/pascal/VOC/voc2012/}}
}

@INPROCEEDINGS{Glorot10understandingthe,
    author = {Xavier Glorot and Yoshua Bengio},
    title = {Understanding the difficulty of training deep feedforward neural networks},
    booktitle = {Proceedings of the 13th {International Conference on Artificial Intelligence and Statistics (AISTATS 2010)} },
    pages={249--256},
    year = {2010}
}

@article{kim2018not,
  title={{Not-So-CLEVR}: learning same--different relations strains feedforward neural networks},
  author={Kim, Junkyung and Ricci, Matthew and Serre, Thomas},
  journal={Interface Focus},
  volume={8},
  number={4},
  pages={20180011},
  year={2018}
}

@inproceedings{zhu2017unpaired,
  title={Unpaired image-to-image translation using cycle-consistent adversarial networks},
  author={Zhu, Jun-Yan and Park, Taesung and Isola, Phillip and Efros, Alexei A},
  booktitle={Proceedings of the 30th {IEEE/CVF Conference on Computer Vision and Pattern Recognition (CVPR 2017)}},
  year={2017}
}

@article{alom2018recurrent,
  title={Recurrent residual convolutional neural network based on {U-Net (R2U-Net)} for medical image segmentation},
  author={Alom, Md Zahangir and Hasan, Mahmudul and Yakopcic, Chris and Taha, Tarek M and Asari, Vijayan K},
  journal={arXiv preprint arXiv:1802.06955},
  year={2018}
}

\end{document}